\documentclass[11pt]{article}
\usepackage[letterpaper, margin=1in]{geometry}

\newcommand\blfootnote[1]{%
  \begingroup
  \renewcommand\thefootnote{}\footnote{#1}%
  \addtocounter{footnote}{-1}%
  \endgroup
}
\newcommand\authsep{\hspace{1em plus 0.5em minus 0.2em}\penalty0\hspace{0pt}}
\usepackage[hang,flushmargin]{footmisc}
\usepackage[hidelinks]{hyperref}
\usepackage{natbib} 

\usepackage[utf8]{inputenc} % allow utf-8 input
\usepackage[T1]{fontenc}    % use 8-bit T1 fonts
\usepackage{hyperref}       % hyperlinks
\usepackage{url}            % simple URL typesetting
\usepackage{booktabs}       % professional-quality tables
\usepackage{amsfonts}       % blackboard math symbols
\usepackage{nicefrac}       % compact symbols for 1/2, etc.
\usepackage{microtype}      % microtypography
\usepackage{xcolor}         % colors

\usepackage{float}
\usepackage{colortbl}
\usepackage{graphicx}
\usepackage{subcaption}
\usepackage{caption}
\usepackage{amsmath}
\usepackage{amssymb}
\usepackage{mathtools}
\usepackage{amsthm}
\usepackage{multirow}

\usepackage{pifont}
\newcommand{\cmark}{{\ding{51}}}
\newcommand{\xmark}{{\ding{55}}}

\definecolor{basiccolor}{gray}{0.95}
\definecolor{temporalcolor}{RGB}{218,232,252}
\definecolor{multicolor}{RGB}{198,224,180}
\definecolor{anomalycolor}{RGB}{248,206,204}

\usepackage[capitalize,noabbrev]{cleveref}

\theoremstyle{plain}

\theoremstyle{definition}

\theoremstyle{remark}

\usepackage[textsize=tiny]{todonotes}

\title{TS-Haystack: A Multi-Task Retrieval Benchmark\\for Long-Context Time-Series Reasoning}

\author{%
\begin{minipage}{\linewidth}\centering
\small
\textbf{Nicolas Zumarraga}$^{1\star}$\authsep
\textbf{Thomas Kaar}$^{1,2}$\authsep
\textbf{Ning Wang}$^{1}$\authsep
\textbf{William Tennien}$^{2}$\authsep
\textbf{Alpay Hasanli}$^{1}$\authsep
\textbf{Max Rosenblattl}$^{2}$\authsep
\textbf{Fan Wu}$^{1}$\authsep
\textbf{Kevin Riehl}$^{1,3}$\authsep
\textbf{Maxwell A.\ Xu}$^{4,5}$\authsep
\textbf{Markus Kreft}$^{1}$\authsep
\textbf{Kevin O'Sullivan}$^{1}$\authsep
\textbf{Elgar Fleisch}$^{1,6,7}$\authsep
\textbf{Paul Schmiedmayer}$^{2}$\authsep
\textbf{Robert Jakob}$^{1\dagger}$\authsep
\textbf{Patrick Langer}$^{1,2,6\dagger}$%
\end{minipage}\vspace{-1.5em}%
}
\date{}

\begin{document}

\maketitle

\blfootnote{%
$^\star$Correspondence: \texttt{nzumarraga@ethz.ch}. \quad
$^\dagger$Equal contribution as senior authors. \\[2pt]
$^{1}$Agentic Systems Lab, ETH Zurich. \quad
$^{2}$Stanford University. \quad
$^{3}$Traffic Engineering Group, Institute for Transport Planning and Systems, ETH Zurich. \quad
$^{4}$University of Illinois Urbana-Champaign. \quad
$^{5}$Google. \quad
$^{6}$Centre for Digital Health Interventions, ETH Zurich. \quad
$^{7}$Centre for Digital Health Interventions, University of St.~Gallen.
}

\vspace{-1em}
\begin{abstract}
Time Series Language Models (TSLMs) promise reasoning over real-world temporal data, but their ability to retrieve and reason over long time-series remains largely untested.
We introduce \texttt{TS-Haystack}, a multi-domain benchmark with ten event-grounded question-answering tasks over contexts from 100 seconds to 24 hours, spanning direct retrieval, temporal reasoning, multi-step reasoning, and contextual anomaly detection.
Existing TSLMs exhibit severe long-context degradation: accuracy declines with context length, direct-tokenization models run out of memory beyond 100 seconds on high-rate signals, and time-interval-grounded tasks collapse toward near-zero accuracy when increasing the time-series lengths, aligning with existing literature on text and multi-modal long context retrieval.
% We further introduce \texttt{ARTS}, an agentic retrieval baseline that augments GPT-5.4 with in-context time-series classifier tools.
% \texttt{ARTS} matches or outperforms SoTA TSLMs on 9 of 10 tasks, highlighting agentic retrieval as a promising direction for long-context TSLM research.
An agentic retrieval framework using specialized time-series classifier tools matches or outperforms SoTA TSLMs on 9 of 10 tasks, highlighting agentic retrieval as a promising approach for long-context TSLMs. Our code and datasets are publicly available at \texttt{https://github.com/AI-X-Labs/TS-Haystack}.
\end{abstract}

%%%%%%%%%%%%%%%%%%%%%%%%%%%%%%%%%%%%%%%%%%%%%%%%%%%%%%%%%%%%%%%%%%%%%%%%%%%%%%%
% INTRODUCTION
%%%%%%%%%%%%%%%%%%%%%%%%%%%%%%%%%%%%%%%%%%%%%%%%%%%%%%%%%%%%%%%%%%%%%%%%%%%%%%%

\section{Introduction}

% Paragraph 01: timeseries processing in general, why it matters, link from different methods to TSLMs
Time-series analysis enables the extraction of temporal dependencies and meaningful patterns in domains such as finance, healthcare, climate modeling, and industrial IoT~\citep{benidis2022deep}. 
%
% This processing addresses a variety of tasks including classification, forecasting, anomaly detection, retrieval, imputation, and representation learning~\citep{ismail2019deep}. 
%
Methodologically, the field has evolved from classical statistical models to deep learning architectures, which allow capturing of non-linear patterns in high-dimensions~\citep{torres2021deep}. 
Building upon this progression, time-series foundation models (TSFMs) and time-series language models (TSLMs) have emerged as a promising frontier for general time-series forecasting and reasoning. Similar to large language models (LLMs) for text and vision language models (VLMs) for images, TSLMs aim to achieve robust generalization across diverse datasets and temporal tasks~\citep{zhang2024large, abdullahi2025time, ansari2025chronos2}.

% Paragraph 02: limitations when it comes to temporal retrieval and reasoning in long continuous time-series
Various real-world applications and the increasing availability and frequency of data require the processing of long time-series horizons. Continuous physiological ICU recordings~\citep{yeche2021hirid}, full-night polysomnography~\citep{ghassemi2018you}, and wearable activity monitoring routinely~\citep{chan2024capture24} span eight to twenty-four hours, industrial sensor streams run uninterrupted for days or weeks~\citep{aircraftfailues}, and climate sensor signals may only reveal a salient narrative over the span of many years~\citep{era5climate}. Such real-world applications require \emph{retrieving}, \emph{aggregating} and \emph{reasoning} over events across large amounts of contextualized data. 
TSLM architectures that use full self-attention for time series enable more fine-grained temporal representations, but they incur high computational costs at longer sequence lengths due to the quadratic cost of attention with respect to time-series length~\citep{nie2023time, vaswani2017attention}. Conversely, fixed latent-space architectures improve efficiency through compression, but their ability to preserve temporal granularity at long context remains an open question.

% Paragraph 03: current benchmark datasets are limited to short range; no dataset for long-term data
While TSLMs’ reasoning capabilities have been extensively tested on short context window classification~\citep{langer2025opentslm, wang2025itformer, Xie_2025}, determining whether these models can localize, count, or compare events at longer temporal horizons remains an open question. 
Several established benchmarks have been proposed to better understand the behavior of LLMs on long-context retrieval tasks for text~\citep{kuratov2024babilongtestinglimitsllms, kamradt2023needle}, vision~\citep{wang2025multimodal}, and audio~\citep{he2025audiomarathon}, unveiling performance degradation at increased context lengths, and leading to architectural breakthroughs in long context retrieval~\citep{zhang2026recursivelanguagemodels, behrouz2025titans, gu2023mamba}.
While existing benchmarks for TSLMs have explored statistical reasoning~\citep{cai2024timeseriesexam}, classification~\citep{UCRArchive2018, bagnall2018ueamultivariatetimeseries}, and chain-of-thought augmented classification~\citep{langer2025opentslm}, a standardized benchmark for long-context reasoning and retrieval is missing for time-series.

\begin{figure}[t]
\begin{center}
\vspace{-10mm}
\includegraphics[width=\textwidth]{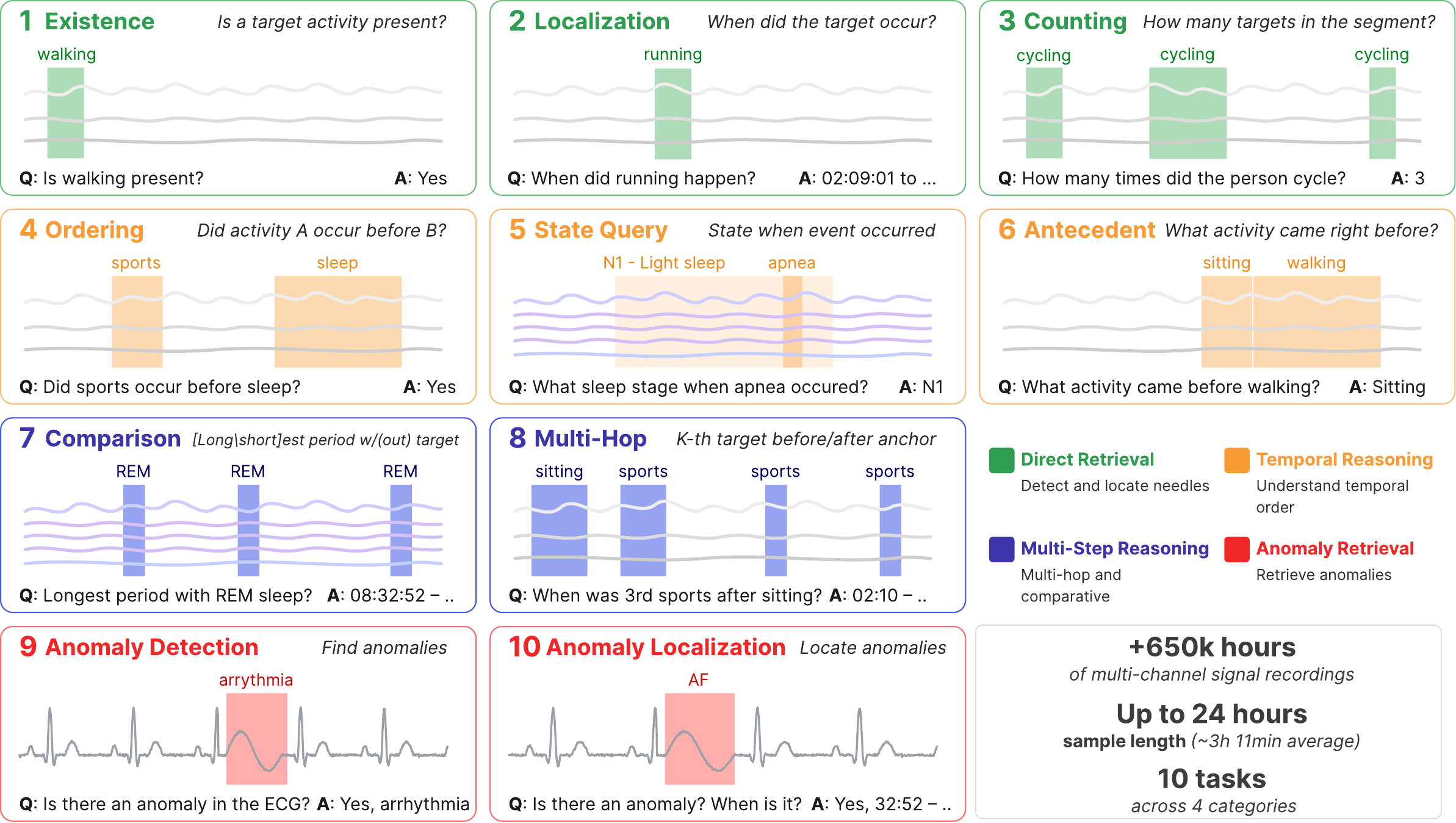}
\caption{\textbf{TS-Haystack Tasks Overview} Ten tasks spanning four domains for time-series reasoning and retrieval. 
%\emph{Direct Retrieval} probes whether models can maintain temporal granularity of target events (existence, localization, counting). \emph{Temporal Reasoning} tests understanding of temporal order and co-occurrence (ordering, state query, antecedent). \emph{Multi-Step Reasoning} requires comparative and multi-hop inference across multiple events (comparison, multi-hop). \emph{Anomaly Retrieval} evaluates detection and localization of relevant real-world and contextual anomalies. 
Tasks are instantiated with sample lengths ranging from 100~s up to 24~h (3~h 11~min average) across more than 650k hours of multi-channel recordings.}
\label{fig:all_tasks}
\vspace{-4mm}
\end{center}
\end{figure}

% Paragraph 04: PRESENT AND SELL THE DATASET
We present \textbf{TS-Haystack}, a novel benchmark for temporal reasoning over long-context time series that enables systematic diagnosis of long-context performance and offers a pathway toward more robust and scalable TSLMs. It comprises ten tasks across four domains: direct and temporal retrieval, multi-step reasoning, and anomaly detection as shown in Figure~\ref{fig:all_tasks}. Each task is instantiated at controlled context lengths from 100~s up to 24~h at different sampling rates, enabling the study of retrieval and reasoning ability against temporal scale. 
To create this benchmark, we extend the needle-in-a-haystack paradigm~\citep{wang2024mmniah} to time-series, enabling generation of retrieval-based question-answer pairs from existing long horizon labeled time-series data. We create samples under two complementary protocols. When local signals are uninformed by their surrounding context (e.g., local activity recognition, an appliance's power consumption at a given time), we synthesize samples by embedding existing annotated events (\emph{needles}) into other longer background recordings (\emph{haystacks}). For domains where local events are better informed by surrounding signals and are hard to disentangle from their local context (e.g., heartbeat anomalies, sleep stages that follow a cyclic macro-trend), we extract naturally occurring segments (e.g., a contiguous 1-hour window of an ECG recording with its native beat annotations, or a 2-hour slice of polysomnography with the original sleep-stage and arousal labels) from the source data and condition questions for that particular sample. 
We provide a framework and statistically validate both protocols in our dataset generation pipeline (Section~\ref{sec:construction}).

%%%% TS-Haystack figure in top of second page

% Paragraph 05: PRESENT AND SELL ARTS
We benchmark established TSLMs on \texttt{TS-Haystack}, finding a degradation of accuracy by context length, with performance on localization-specific tasks such as localization, comparison and multi-hop collapsing to near-zero beyond lengths of 1~h (see Appendix~\ref{app:results_breakdown}). To address this limitation, we demonstrate how the benchmark can be used to develop novel TSLM architectures for long context time-series retrieval. Specifically, while existing TSLMs ingest signals and answer questions in a single pass, we propose a novel agentic framework that equips an LLM orchestrator with specialized classifier tools for iterative temporal reasoning and retrieval, decoupling local perception from global reasoning. 
The proposed framework achieves state-of-the-art accuracy on \texttt{TS-Haystack}, providing an interesting direction for future research in agentic time-series retrieval. An additional oracle baseline using timestamped time-series labels as text input achieves ${>}90\%$ accuracy, pointing at fine-grained time-series feature representations as the origin of this long context performance degradation.

%%%%%%%%%%%%%%%%%%%%%%%%%%%%%%%%%%%%%%%%%%%%%%%%%%%%%%%%%%%%%%%%%%%%%%%%%%%%%%%
% RELATED WORK
%%%%%%%%%%%%%%%%%%%%%%%%%%%%%%%%%%%%%%%%%%%%%%%%%%%%%%%%%%%%%%%%%%%%%%%%%%%%%%%
\section{Related Work}
\label{sec:relatedwork}

\paragraph{Time-Series Language Models}
\label{p:tslms}
TSLMs are multi-modal LLMs that integrate time-series as a native modality~\citep{langer2025opentslm}. We categorize different architectures in how they map the time-series input of length $L$ into the language model's input space. \emph{Full-resolution adapters}, such as ChatTime~\citep{wang2024chattimeunifiedmultimodaltime} and ChatTS~\citep{Xie_2025}, pass time-series tokens or $N = L/P$ patch embeddings directly to the LLM, preserving signal fidelity at the cost of $\mathcal{O}(N^2)$ attention in the extended LLM context, which limits scaling to long contexts. ChatTime tokenizes the signal directly~\citep{wang2024chattimeunifiedmultimodaltime}, while ChatTS adopts a LLaVA-style~\citep{liu2023visualinstructiontuning} insertion of patch embeddings into the LLM input. \emph{Fixed-latent} methods instead integrate a module between the encoder and the LLM, which learns to compress the time series. ITFormer~\citep{wang2025itformer} aggregates patch embeddings into a fixed set of summary tokens, while OpenTSLM-Flamingo~\citep{langer2025opentslm} adopts a Perceiver Resampler~\citep{jaegle2021perceiver} in which $M$ learned queries $\mathbf{Q} \in \mathbb{R}^{M \times d}$ cross-attend~\citep{vaswani2017attention, alayrac2022flamingo} to the $N$ patch embeddings and produce an output $\mathbf{Z} \in \mathbb{R}^{M \times d}$ of constant size, enabling linear scaling in $L$.

\paragraph{Time-Series Reasoning Benchmarks}

Several foundational benchmarks for time-series classification~\citep{UCRArchive2018, bagnall2018ueamultivariatetimeseries} and forecasting~\citep{li2025tsfmbenchcomprehensiveunifiedbenchmark, aksu2024giftevalbenchmarkgeneraltime} evaluate fixed-length categorization or forecasting, yet do not test natural-language retrieval and reasoning over time-series in the ways that text and vision benchmarks have explored~\citep{yue2024mmmumassivemultidisciplinemultimodal, singh2019vqamodelsread}.
Recent TSLM-centric benchmarks include text and time-series data, but remain restricted in scope: TimeSeriesExam~\citep{cai2024timeseriesexam} probes statistical understanding via multiple-choice questions on short series. Due to this gap, TSLM implementations have focused their evaluations on short-context classification with optional chain-of-thought rationale generation~\citep{langer2025opentslm, wang2025itformer}, leaving an open question on their ability to retrieve and reason over target information at long horizons. We summarize how TS-Haystack relates to representative time-series benchmarks along five axes in Table~\ref{tab:benchmark_comparison}.

\begin{table}[h]
\centering
\caption{TS-Haystack compared to representative time-series benchmarks. \emph{Long context}: continuous samples exceeding 1~h. \emph{Language}: tasks formulated as natural-language question answering. \emph{Event-grounded}: questions reference real annotated events. \emph{Multi-step reasoning}: tasks require sequential retrieval and reasoning (e.g., find all target segments, compute their length and answer with the longest section without such activity). \emph{Multi-domain}: spans more than one signal modality.}
\label{tab:benchmark_comparison}
\resizebox{\textwidth}{!}{%
\begin{tabular}{lccccc}
\toprule
Benchmark & Long context & Language & Event-grounded & Multi-step reasoning & Multi-domain \\
\midrule
UCR/UEA~\citep{UCRArchive2018, bagnall2018ueamultivariatetimeseries} & \xmark & \xmark & \cmark & \xmark & \cmark \\
TSFM-Bench~\citep{li2025tsfmbenchcomprehensiveunifiedbenchmark}      & \cmark & \xmark & \xmark & \xmark & \cmark \\
GIFT-Eval~\citep{aksu2024giftevalbenchmarkgeneraltime}               & \cmark & \xmark & \xmark & \xmark & \cmark \\
TimeSeriesExam~\citep{cai2024timeseriesexam}                         & \xmark & \cmark & \xmark & \xmark & \cmark \\
Capture24~\citep{chan2024capture24}                                  & \cmark & \xmark & \cmark & \xmark & \xmark \\
PhysioNet'18~\citep{ghassemi2018you}                                 & \cmark & \xmark & \cmark & \xmark & \xmark \\
LTAF~\citep{petrutiu2007ltaf}                                        & \cmark & \xmark & \cmark & \xmark & \xmark \\
UK-DALE~\citep{kelly2015ukdale}                                      & \cmark & \xmark & \cmark & \xmark & \xmark \\
\midrule
\textbf{TS-Haystack (ours)}                                          & \cmark & \cmark & \cmark & \cmark & \cmark \\
\bottomrule
\end{tabular}%
}
\end{table}

\paragraph{Long-Context Retrieval Benchmarks}
The needle-in-a-haystack (NIAH) paradigm probes retrieval by embedding target information (the \emph{needle}) within a long background of unrelated content (the \emph{haystack})~\citep{kamradt2023needle, kuratov2024babilongtestinglimitsllms}, requiring the model to recover these needles to successfully answer direct or reasoning-heavy questions. NIAH-inspired benchmarks have revealed reasoning and retrieval degradation of LLMs on text~\citep{liu2023lostmiddle}, and analogous benchmarks have surfaced the same failure mode in vision~\citep{wang2024mmniah, song2024milebench} and audio~\citep{he2025audiomarathon}. Extending this paradigm to time-series is non-trivial: unlike text, where a needle sentence can be cleanly inserted into an unrelated passage, time-series signals lack discrete, semantically separable units.
Naive insertion for time-series risks two failure modes: the introduction of artifacts at the seam that may trivialize retrieval through pattern matching and the destruction of context-dependent semantics for labeled events (e.g., a few seconds of ECG recording is labeled as arrhythmia yet its diagnosis depends on adjacent beats excluded from the labeled event). To our knowledge, no equivalent long-context, multi-domain, diagnostic benchmark exists for time-series.

\paragraph{Agentic Long-Context Reasoning}
A parallel line of work studies \textit{agentic} frameworks in which an LLM iteratively issues tool calls and integrates intermediate results into its context~\citep{yao2023react, chen2023memwalker}. Such frameworks have been shown to recover much of the long-context performance lost by single-pass question answering, and have been applied to long-form multimodal retrieval including video understanding~\citep{wang2024videoagent}.

%%%%%%%%%%%%%%%%%%%%%%%%%%%%%%%%%%%%%%%%%%%%%%%%%%%%%%%%%%%%%%%%%%%%%%%%%%%%%%%
% METHODS (2-3 pages)
%%%%%%%%%%%%%%%%%%%%%%%%%%%%%%%%%%%%%%%%%%%%%%%%%%%%%%%%%%%%%%%%%%%%%%%%%%%%%%%

\section{Methods: TS-Haystack}
\label{sec:tshaystack}

We propose TS-Haystack, a benchmark that repurposes long, expert-annotated recordings from multiple domains into ten retrieval and reasoning tasks instantiated at controlled context lengths from 100~s up to 24~h. 
We organize TS-Haystack around relevant questions for long, continuous recordings in time-series rich domains, grouped into four categories: \emph{Direct Retrieval} (Existence, Localization, Counting) probes whether models maintain temporal granularity at scale; \emph{Temporal Reasoning} (Ordering, State Query, Antecedent) tests understanding of order and co-occurrence; \emph{Multi-Step Reasoning} (Comparison, Multi-Hop) requires composing multiple retrievals; and \emph{Anomaly Retrieval} (Anomaly Detection, Anomaly Localization) evaluates detection and localization of relevant anomalies. Figure~\ref{fig:all_tasks} presents a visualization of the ten tasks, answer types and example questions. Each task is implemented using a family of paraphrased natural-language templates and admits a deterministic ground-truth answer computed from the underlying annotations. Further details are reported in Appendix~\ref{app:task-construction-details}.

\subsection{Source Datasets}
\label{sec:sources}

TS-Haystack draws from four openly available, annotated long-form datasets selected for diversity in modality, sampling rate, and event sparsity (Table~\ref{tab:source_datasets}): Capture24~\citep{chan2024capture24}, Sleep Polysomnography~(PSG)~\citep{ghassemi2018you}, and LTAF~(Long-Term Atrial Fibrillation)~\citep{petrutiu2007ltaf} are three clinical signals (wrist accelerometer, polysomnography, and ECG). UK-DALE~\citep{kelly2015ukdale} is a household electricity consumption dataset spanning five homes over long horizons. 
Although these domain-specific datasets provide long, expert-annotated recordings, these datasets are released with per-window classification labels and have not been formulated as retrieval or reasoning tasks.
Complete per-dataset details and statistics are reported in Appendix~\ref{app:ts_haystack}. We note that despite UK-DALE's 24~h wall-clock horizon, its low sampling rate keeps the per-sample sequence length below that of the rest of datasets at longer contexts (e.g., a 24~h UK-DALE window contains $\sim$14k samples, while a 2~h Sleep PSG window contains $\sim$720k samples per channel), making target signals in Capture24, LTAF and Sleep PSG sparser despite the lower perceived context length.

\begin{table}[h]
\centering
\small
\setlength{\tabcolsep}{6pt}
\renewcommand{\arraystretch}{1.2}
\caption{TS-Haystack source datasets. Sleep PSG is downsampled from its 200\,Hz native rate to 100\,Hz. \emph{Classes} reports the size of each label vocabulary (aggregating 5 AASM stages + 5 arousal classes for Sleep and 4 beat classes + 9 rhythm codes for LTAF). Full per-dataset statistics are reported in Appendix~\ref{app:ts_haystack_overview}.}
\label{tab:source_datasets}
\begin{tabular}{@{}lcccll@{}}
\toprule
Dataset & Modality & Channels & Frequency (Hz) & Classes & Context lengths \\
\midrule
Capture24$^{a}$  & Wrist accelerometer   & 3  & 100               & 10   & 100\,s, 15\,min, 1\,h, 2\,h \\
Sleep PSG$^{b}$  & Polysomnography       & 13 & 100               & 10  & 100\,s, 15\,min, 1\,h, 2\,h, 9\,h \\
LTAF$^{c}$       & 2-lead ECG            & 2  & 128               & 13  & 100\,s, 15\,min, 1\,h, 2\,h \\
UK-DALE$^{d}$    & Household mains power & 1  & $\sim$0.17 (6\,s) & 10   & 15\,min, 1\,h, 2\,h, 9\,h, 24\,h \\
\bottomrule
\end{tabular}\\[3pt]
{\footnotesize
$^{a}$\citet{chan2024capture24};\quad
$^{b}$\citet{ghassemi2018you};\quad
$^{c}$\citet{petrutiu2007ltaf};\quad
$^{d}$\citet{kelly2015ukdale}.
}
\end{table}

\subsection{Dataset Construction}
\label{sec:construction}

We instantiate TS-Haystack by assigning recordings to a train/validation/test split. Sample generation follows one of two protocols depending on whether the source signal supports clean recombination: \emph{semi-synthetic needle insertion}, in which annotated bouts are embedded into sampled background recordings, and \emph{natural-segment sampling}, in which original windows are extracted from contiguous recordings with their original annotations as ground truth. The following section motivates the choice between the two and details the insertion protocol used when the conditions for it are met.

\paragraph{Semi-synthetic vs.\ natural sample generation}
Semi-synthetic sample generation is motivated by two limiting factors present in natural long-horizon recordings: class imbalance and predictable label transitions. It also unlocks the potential for two capabilities: controlled difficulty modulation (tuning needle length, count, and spacing independently of context length) and augmented sample generation (recombination of needles and backgrounds beyond the corpus's natural yield). 
Table~\ref{tab:construction_choice} summarizes the trade-offs and sample generation choices for each dataset.
The decision rests on whether a bout of a target class can be \emph{dissociated from its surrounding context} without altering its semantics. Formally, we consider a signal as \emph{locally context-independent} for class $c$ if the distribution of a class-$c$ bout conditional on its surrounding context is well-approximated by its marginal distribution given the label, $p(\mathbf{x}_{[s:e]} \mid \mathbf{X}_{\setminus[s:e]},\, c) \approx p(\mathbf{x}_{[s:e]} \mid c)$~(formal statement in Appendix~\ref{app:needle-insertion}). If this holds true, transplanting a bout into a different background of the same domain can be considered to preserve its semantics, and insertion yields composites statistically close to naturally observed recordings; when it fails, insertion produces out-of-distribution artifacts that no boundary smoothing can repair, and we fall back to natural-segment sampling.
Activity recognition is conventionally cast as classification over short, locally segmented windows~\citep{bulling2014tutorial}: a bout of running carries the same activity label whether it follows sitting or walking. A similar logic applies to additive appliance signatures on household mains, where use of an appliance at a given time is unconditional to its context. Sleep and ECG do not satisfy the condition, as labeled events are not naturally interpretable without their surrounding context: sleep stages are scored over 30~s epochs embedded in the cyclic NREM/REM macrostructure encoded by AASM rules~\citep{berry2017aasm}; ECG carries strong inter-beat dependencies (R-R autocorrelation, beat-to-beat morphology, and characteristic onset/offset patterns of arrhythmias).
To verify that insertions are not detectable from artifacts alone, we train an XGBoost~\citep{chen2016xgboost} binary classifier to distinguish background-only windows from windows with a same-activity needle inserted from a different participant; an $\text{AUC} \approx 0.50$ indicates that the insertion produces no systematic detectable artifact (Appendix~\ref{app:insertion_validation}).

\paragraph{Needle insertion protocol}
Semi-synthetic needle insertion provides a scalable method to generate complex retrieval and reasoning tasks at controlled lengths. Given a target context length $L$ and a queried class $c$, we sample a background recording from a participant, select one or more annotated bouts $\mathbf{x}_{\text{needle}}$ of class $c$ from a random participant in the same split, and insert them at controlled locations. The number, location, and labels of these needles determine the answer key for each task.
% Additionally, we use \emph{fixed-length} needles per task and context length pair to maintain a strictly decaying needle-to-background ratio mimicking real activity. Further ablation on this construction decision confirms a key factor of performance degradations with respect to context length.

\begin{table}[h]
\centering
\small
\caption{Trade-offs between TS-Haystack's two sample-construction protocols. When the precondition is satisfied, semi-synthetic insertion offers strictly more operational control; natural-segment sampling serves as the fallback when local segments are not approximately context-independent.}
\label{tab:construction_choice}
\begin{tabular}{@{}lcc@{}}
\toprule
                                   & \textbf{Semi-synthetic insertion}   & \textbf{Natural-segment sampling} \\
\midrule
Applicable when                    & Context-independent local segments  & Any signal                        \\
\midrule
Arbitrary context-length scaling   & \cmark                              & \xmark                            \\
Combinatorial sample diversity     & \cmark                              & \xmark                            \\
Explicit difficulty control        & \cmark                              & \xmark                            \\
Classifier-based validation        & \cmark                              & \cmark                            \\
\midrule
TS-Haystack use                    & Capture24, UK-DALE                           & Sleep PSG, LTAF                   \\
\bottomrule
\end{tabular}
\end{table}

\paragraph{TSLMs and NIAH Formalization for Time-Series}
\label{sec:formalization}
TS-Haystack adapts the needle-in-a-haystack paradigm to time-series, casting long-context retrieval and reasoning as a question-answering problem over multi-channel recordings. Each TS-Haystack sample consists of three components: a long recording (the \emph{haystack}), a set of sparse annotated \emph{events of interest} embedded within it (the \emph{needles}), and a natural-language question grounded in those needles. Task difficulty arises because needles occupy a vanishing fraction of the total recording length: a successful model must retrieve, localize, and reason about events that are sparse in time, preserving the fine-grained signal detail required to identify them. We formalize this as follows. 
Let $\mathbf{X} \in \mathbb{R}^{L \times C}$ denote a multi-channel recording of length $L$ with $C$ channels, and let $q$ denote a natural-language question over $\mathbf{X}$. Each TS-Haystack sample is a tuple $(\mathbf{X}, q, y)$, where $y$ is a structured answer (boolean, integer, label, or interval set). The annotation $\mathcal{A}(\mathbf{X}) = \{(s_i, e_i, c_i)\}_{i=1}^{n}$ comprises $n$ disjoint events of interest, each defined by start and end indices $0 \leq s_i < e_i \leq L$ and a class label $c_i \in \mathcal{C}$ drawn from the source dataset's annotations. By construction, the aggregate needle mass satisfies $\sum_i (e_i - s_i) \ll L$.

We define a TSLM as a function $f_\theta: (\mathbf{X}, q) \mapsto \hat{y}$ trained to maximize task-specific accuracy. We hypothesize that the constant-size compression employed by fixed-latent architectures introduces an information bottleneck under long contexts: as $L$ grows, the linearly increasing compression ratio $N/M$ forces each latent token to progressively marginalize away information relevant to $q$. In OpenTSLM-Flamingo, $M$ defaults to $64$, yielding a $\sim$2,800$\times$ compression ratio at 2\,h of context sampled at 100\,Hz with patch size $4$. TS-Haystack is designed to stress both encoding regimes by the formalized needle structure above, where the answer depends on intervals whose aggregate length is vanishing as $L$ grows.
% Fixed-latent architectures (e.g., OpenTSLM-Flamingo, ITFormer) may lose the resolution required to localize sparse needles under the bottleneck above, while full-resolution architectures (e.g., ChatTS, ChatTime) exhaust available compute under $\mathcal{O}(N^2)$ attention.

\section{ARTS: Agentic Retrieval for Time-Series}
\begin{figure}[t]
\begin{center}
\includegraphics[width=\textwidth]{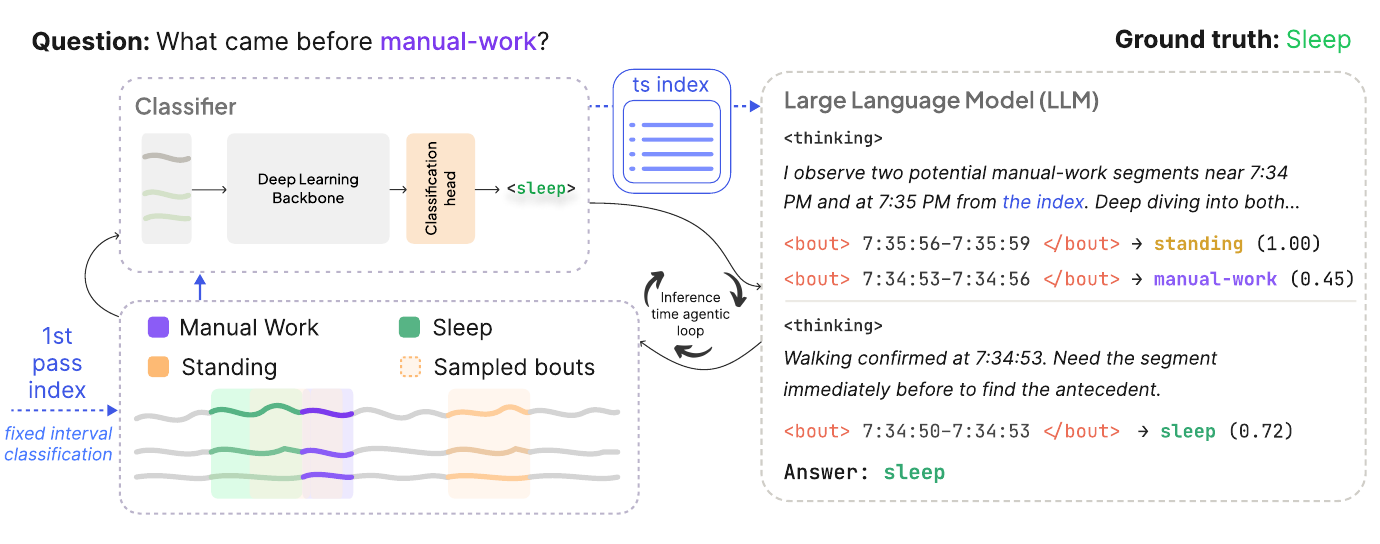}
\caption{\textbf{ARTS.} A classifier pre-pass sweeps the recording over a non-overlapping tiles, producing a textual timeline of predicted labels and confidences that is supplied to the orchestrator as initial context. The orchestrator then issues \texttt{<bout>} tool calls to re-classify specific segments, integrates intermediate observations into its running context, and iteratively reasons toward a final answer.}
\label{fig:arts_architecture}
\vspace{-4mm}
\end{center}
\end{figure}

Existing TSLM architectures couple local perception and global reasoning into a single forward pass, motivating an alternative that decouples \emph{local perception} (short segments at full resolution) from \emph{global reasoning} (a compact symbolic view of the whole recording). ARTS reframes long-context retrieval as a two-stage process using an \emph{orchestrator} $f_\theta$, responsible for reasoning and temporal navigation, and specialized \emph{classifier tools} $g_\phi$, responsible for fine-grained signal understanding over shorter segments of the full recording (Figure~\ref{fig:arts_architecture}). 
In the first stage, a \emph{classifier pre-pass} sweeps $g_\phi$ over a non-overlapping tiling of the full recording at a fixed window size, producing a textual timeline of predicted labels that summarizes $\mathbf{X}$ for the orchestrator. The orchestrator reads this \textit{Time-series index} and iteratively issues \texttt{<bout>} tool calls to ``zoom-in'', re-classify selected segments and resolve ambiguities left by the pre-pass, integrating intermediate observations until it emits a final answer. Crucially, the orchestrator never ingests $\mathbf{X}$ directly: it operates entirely over the symbolic timeline produced by $g_\phi$, decoupling local perception from global reasoning. The pre-pass amortizes the cost of covering the full recording into a single sweep, allowing the agentic loop to focus on targeted refinement rather than blind exploration.

We instantiate the orchestrator with GPT-5.4 (\texttt{gpt-5.4-2026-03-05})~\citep{singh2025openaigpt5card} used off-the-shelf with default temperature and high reasoning effort, demonstrating zero-shot agentic temporal retrieval capabilities. The classifier tool $g_\phi$ is instantiated per domain: for Capture24, a dual-encoder model concatenating frozen Chronos-2~\citep{ansari2025chronos2} and OxWearables~\citep{yuan2024oxwearables} embeddings, with a two-layer multi-layer perceptron (MLP) classification head trained on Capture24 for activity recognition. 
For Sleep, we use two parallel Chronos-2 heads for sleep staging and arousal detection. 
For ECG, we use two complementary classifiers trained from scratch on LTAF: a history-time-frequency CNN ensemble with a 3-class MLP head for per-beat classification, and a wide 1D-ResNet with test-time augmentation for per-rhythm classification~\citep{wang2021ecgcwtcnn, li2022ecghybrid, hannun2019cardiologist}.
We motivate the use of per-domain classifiers as one of the main advantages of ARTS: decoupling reasoning from time-series classification allows to leverage existing domain-specific models, extending the use of validated classifiers to a reasoning paradigm. 
A detailed explanation of their construction details can be found in Appendix~\cref{app:arts-classifier-capture24,app:arts-classifier-sleep,app:arts-classifier-ltaf,app:arts-classifier-ukdale}. 
At each step the orchestrator receives the question, together with the \textit{time-series index} produced by sweeping the classifier over non-overlapping tiles of the signal. It may then emit up to $T_{\max}=15$ \texttt{<bout>} tool calls to re-classify specific segments at finer resolution and verify the classifier predictions before committing to a final answer. Each iteration receives the history of prior tool calls and outputs as a reasoning trace.
% We emphasize that ARTS is offered as a \emph{reference baseline}: a simple approach that general TSLMs should aim to surpass without need of domain specific classifiers and an interesting future research direction, rather than a central methodological contribution of this work.

%%%%%%%%%%%%%%%%%%%%%%%%%%%%%%%%%%%%%%%%%%%%%%%%%%%%%%%%%%%%%%%%%%%%%%%%%%%%%%%
% EXPERIMENTAL SET UP
%%%%%%%%%%%%%%%%%%%%%%%%%%%%%%%%%%%%%%%%%%%%%%%%%%%%%%%%%%%%%%%%%%%%%%%%%%%%%%%

\section{Experimental Setup}
\label{sec:experimental_setup}

We generate train, validation, and test samples for every instantiated cell~(spanning task, context length, source dataset). Existence and Anomaly Detection are not instantiated at longer lengths for Sleep and ECG, since the held-out target classes are almost certainly present in the natural recordings at such lengths, collapsing the answer to a near-deterministic ``Yes''. Per-cell counts and the complete split tables are reported in Appendix~\ref{app:ts_haystack_overview}.

\paragraph{TSLM training and evaluation}
We evaluate four TSLM variants, i.e., ChatTS, ChatTime, OpenTSLM-Flamingo, ITFormer. All use Llama-3.2-1B~\citep{grattafiori2024llama3} as the LLM backbone. To isolate architectural differences from differences in time-series feature quality, we replace OpenTSLM-Flamingo and ITFormer's default encoders with frozen Chronos-2~\citep{ansari2025chronos2}, as this has also shown better results in classification tasks~\citep{langer2025opentslm}, leaving the remainder of each architecture unchanged. 
Models are trained jointly across all tasks and context lengths per signal domain. Optimizer, learning-rate schedule, and per-model hyperparameters are reported in Appendix~\ref{app:tslm_training}.
Discrete-answer tasks (existence, counting, ordering, state query, antecedent, anomaly detection) are scored by exact match after type-aware normalization. Time range tasks (localization, comparison, multi-hop, anomaly localization) are scored as correct if the predicted interval has $\text{IoU} \geq 0.25$ with the ground truth.

\paragraph{Reference Baselines} We complement the model evaluations with an \emph{Oracle} and a \emph{Random Baseline}: two reference baselines that bound accuracy with an upper and lower bound. The oracle bypasses perception entirely: it feeds the ground-truth annotation $\mathcal{A}(\mathbf{X})$ to GPT-5.4 as a textual list of timestamped events and asks it to answer $q$. 
% This information-theoretic oracle's accuracy serves as an upper bound on what any TSLM with perfect feature extraction could achieve; the gap between a model and this oracle quantifies its perception-induced loss.
%
The random baseline is the lower bound of a classifier that never actually receives the content. It is computed in closed form per task and context length pair. Appendix~\ref{app:random-baselines} presents further details on these baselines.
%
% For discrete tasks (existence, counting, ordering, state query, antecedent, anomaly detection), the answer space contains $K$ distinct normalized values observed in the cell and the baseline is $1/K$. For time range tasks (localization, comparison, multi-hop, anomaly localization), the scorer cuts-off $\text{IoU} \geq 0.25$ as a correct answer. For this, a uniformly placed prediction of the same width $w$ as the ground-truth interval, drawn within a recording of length $W$, matches the IoU threshold with probability $\min(1,\, 1.2\,w/W)$. Anomaly localization additionally inherits the scorer's ``no anomaly'' fallback, which we credit at the empirical negative-GT rate and avoids referencing a lower than expected random baseline in case model should learn to always answer with negation.

%%%%%%%%%%%%%%%%%%%%%%%%%%%%%%%%%%%%%%%%%%%%%%%%%%%%%%%%%%%%%%%%%%%%%%%%%%%%%%%
% RESULTS
%%%%%%%%%%%%%%%%%%%%%%%%%%%%%%%%%%%%%%%%%%%%%%%%%%%%%%%%%%%%%%%%%%%%%%%%%%%%%%%
\section{Results}
\label{sec:results}

%%%%%% MODEL ACCURACY OVERALLS - figure 3
\begin{figure}[!tbp]
\begin{center}
\includegraphics[width=\textwidth]{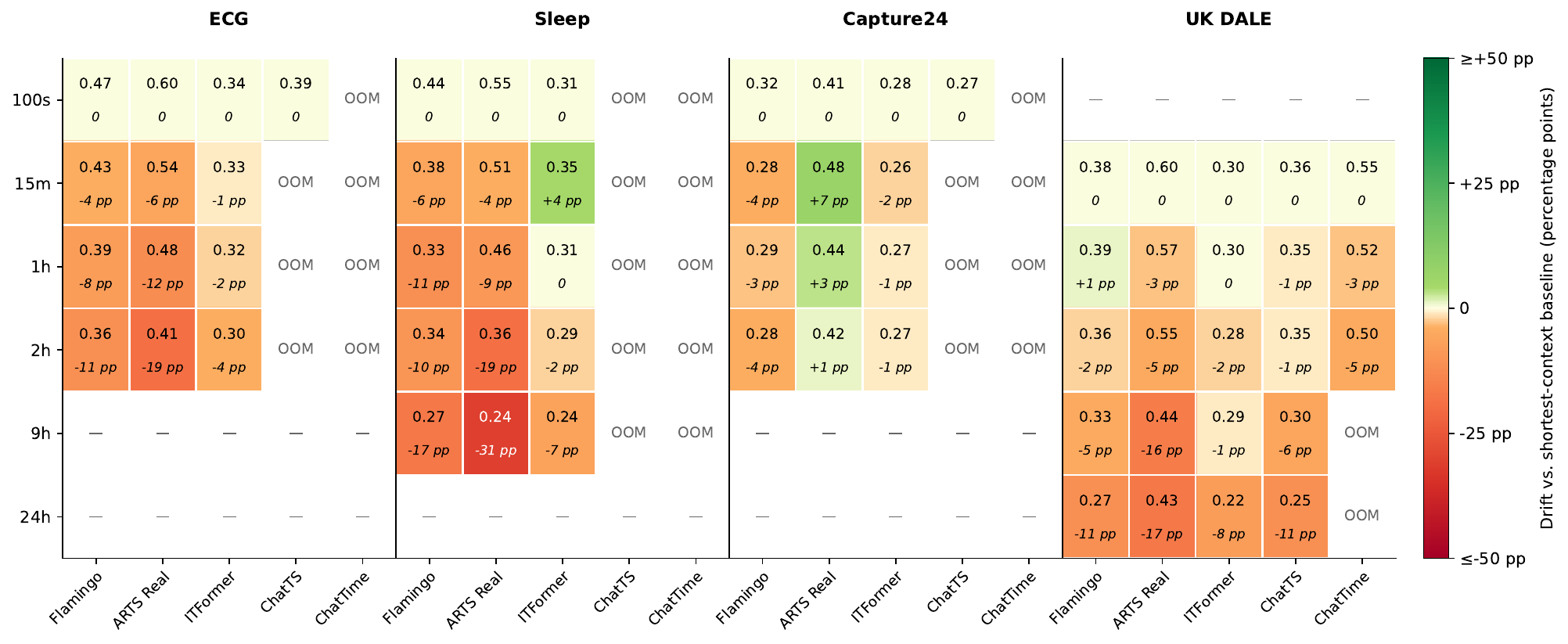}
\caption{\textbf{Per-dataset model performance vs.\ context length.} Mean accuracy across (dataset, model, context length). Cell color encodes relative drift against the shortest-context baseline within each row: green indicates improvement over the short-context baseline, red indicates degradation.}
\label{fig:accuracy_heatmap_per_dataset}
\vspace{-4mm}
\end{center}
\end{figure}

\paragraph{Per-dataset accuracy and drift}
We evaluate Open-TSLM Flamingo, ChatTS, ChatTime, ITFormer and ARTS across every (task, context length, dataset) cell of TS-Haystack; per-task results underlying these aggregates are reported in Appendix~\ref{app:per_domain_results}. Figure~\ref{fig:accuracy_heatmap_per_dataset} shows mean accuracy across tasks for each cell. ARTS attains the highest absolute accuracy in the short-context baseline of every dataset ($0.60$ on ECG and UK-DALE at the shortest length, $0.55$ on Sleep, $0.41$ on Capture24) and remains the strongest method at the longest context tested on three of four datasets. The four datasets exhibit qualitatively distinct decay profiles: ECG and Sleep degrade monotonically and steeply, with ARTS losing the highest relative performance, $19$ and $31$ percentage points (pp) at 2~h and 9~h respectively; UK-DALE degrades more gradually, with ARTS losing $17$~pp over a span that reaches 24~h. Capture24 degrades modestly, with ARTS improving on its 100~s baseline at 15~min. The two direct-tokenization baselines (ChatTS, ChatTime) exhaust GPU memory beyond 100~s on ECG, Sleep, and Capture24, and are evaluable across the full grid only on UK-DALE, where the ${\sim}0.17$~Hz sampling rate keeps token counts tractable.

\begin{figure}[H]
\begin{center}
\includegraphics[width=0.92\textwidth]{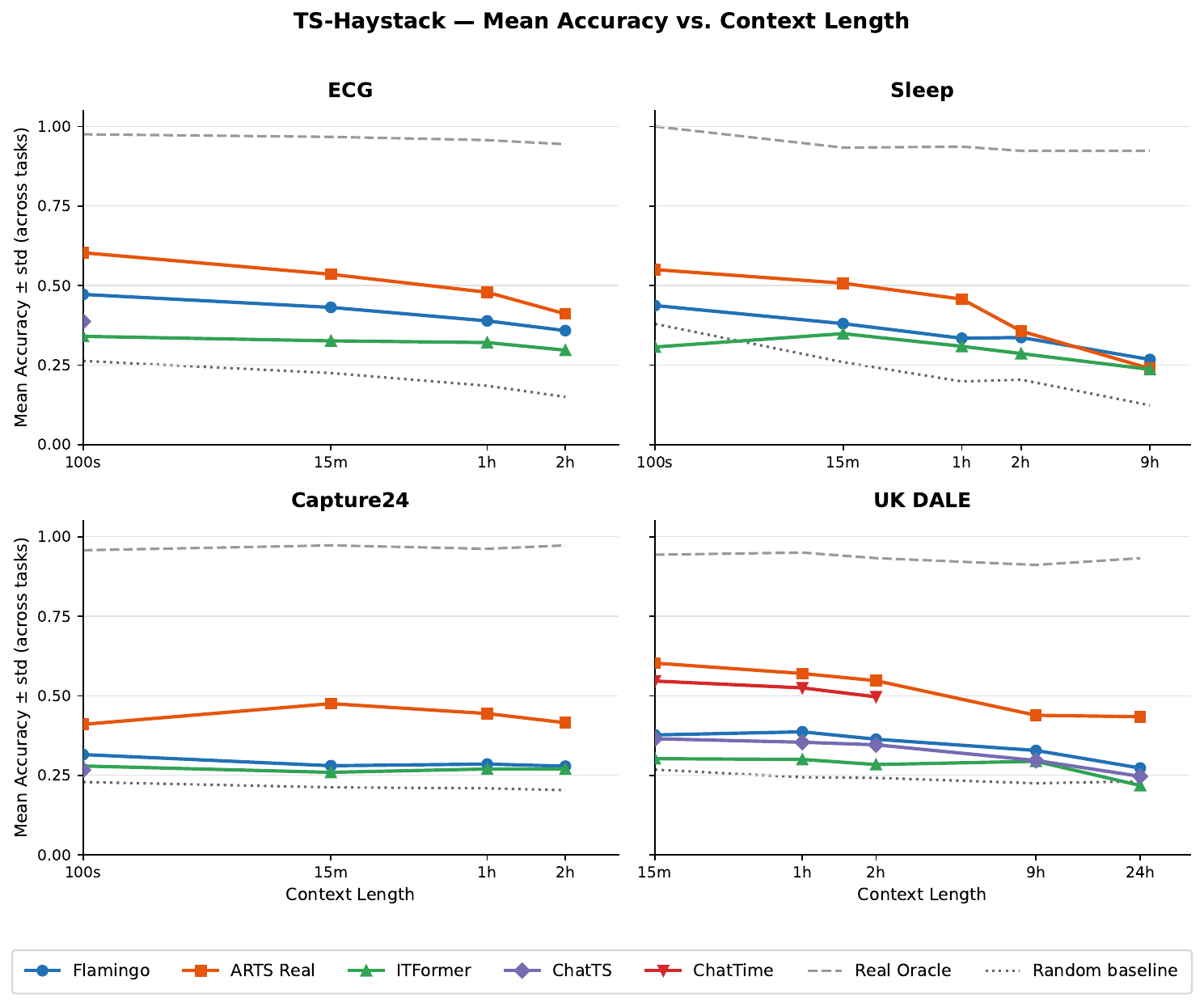}
\caption{\textbf{Per-dataset accuracy bounded between Real Oracle and Random baseline.} Mean accuracy across models and reference baselines at every context length, broken down by source dataset. The Real Oracle (dashed) marks the upper information-theoretic ceiling and the Random baseline (dotted) the content-blind lower bound.}
\label{fig:accuracy_vs_context_length}
\vspace{-4mm}
\end{center}
\end{figure}

\paragraph{Achievable band: oracle ceiling and random floor}
Figure~\ref{fig:accuracy_vs_context_length} situates the same numbers between the oracle and the random baseline. The oracle remains in the $0.93$--$0.97$ band across all four datasets and every evaluated context length, demonstrating that the answers are recoverable from the underlying annotations at every horizon tested. All TSLMs sit inside this band and drift downward within it as context grows. ARTS tracks closest to the oracle ceiling on ECG, UK-DALE, and Capture24 throughout, while on Sleep it converges with Flamingo and ITFormer near the random floor at 9~h. Per-task and per-dataset numbers underlying these aggregates are reported in Appendix~\ref{app:per_domain_results}.

%%%%%% MODEL ACCURACY BY DATASET - figure 4

%%%%%%%%%%%%%%%%%%%%%%%%%%%%%%%%%%%%%%%%%%%%%%%%%%%%%%%%%%%%%%%%%%%%%%%%%%%%%%%
% DISCUSSION
%%%%%%%%%%%%%%%%%%%%%%%%%%%%%%%%%%%%%%%%%%%%%%%%%%%%%%%%%%%%%%%%%%%%%%%%%%%%%%%
\section{Discussion}
\label{sec:discussion}

\paragraph{Context-length degradation in time-series mirrors other modalities}
Across all evaluated TSLMs, we observe an overall decline in accuracy with context length on TS-Haystack. The oracle's stability across the same range ($0.93$--$0.97$) demonstrates that the answer is recoverable from the underlying annotations at every length tested, ruling out an information-theoretic explanation: the degradation is a model failure, not a benchmark limitation. This pattern is qualitatively consistent with documented degradation in long-context LLMs for text~\citep{liu2023lostmiddle}, vision~\citep{wang2024mmniah, song2024milebench}, and audio~\citep{he2025audiomarathon}, and to our knowledge constitutes its first systematic characterization for time-series.

\paragraph{Architectural choices interact with task type, not just context length}
% ITFormer's strong retention (76\%) reflects its low absolute accuracy — close to the random baseline at 100~s — leaving it little to lose at long context, rather than genuine long-context robustness.
% Two confounds in the aggregate view are worth surfacing. First, ITFormer's strong relative retention ($76\%$) reflects its proximity to the random baseline at 100~s rather than genuine robustness; absolute and retained accuracy must be read together.
The per-task breakdowns (Appendix~\ref{app:per_domain_results}) reveal further insights on the performance differences of TSLMs and agentic reasoners: tasks with discrete answers retain performance across context lengths, while tasks with time-interval answers collapse toward zero by 1--2~h. The gap between models and the oracle is widest precisely on these localization-style tasks, suggesting the bottleneck at long context is fine-grained temporal grounding rather than high-level reasoning. ChatTS and ChatTime exhaust GPU memory beyond 100~s, pointing towards the precondition of a decoupling mechanism of context length with the quadratic cost of attention in LLMs for long context TSLMs.

\paragraph{Decoupling perception from reasoning helps but does not solve the problem}
ARTS achieves the highest accuracy in all time-series domains at their initial context lengths, leading the next-best baseline by 13--25 percentage points~(pp) and maintaining relative performance improvement at most context lengths. Yet ARTS also exhibits the largest absolute performance degradation ($-19$~pp on ECG at 2~h, $-31$~pp on Sleep at 9~h, $-17$~pp on UK-DALE at 24~h), and converges near random baseline at the longest horizons. The per-task breakdown in Appendix~\ref{app:per_domain_results} shows that this decline concentrates on time-interval tasks (Localization, Comparison, Multi-hop, Anomaly Localization), where ARTS starts from a higher performance than TSLMs but degrades steeply. Conversely, TSLMs outperform ARTS on tasks such as State Query and Antecedent especially in naturally-sampled datasets, which could point to their single-pass training internalizing marginal label distributions and transition priors that ARTS' inference-only agentic loop cannot exploit (Appendix~\ref{app:per_domain_results}). Plausible explanations for ARTS' residual decline include compounding classifier errors over longer traces and the growing pre-pass timeline diluting the orchestrator's attention over relevant entries. More broadly, results suggest that decoupling local perception from global reasoning is a promising direction for long-context time-series retrieval, but is not on its own sufficient. Future work should focus on addressing such limitations decoupling the main reasoning agent with exploratory sub-agents~\citep{zhang2026recursivelanguagemodels}.

\paragraph{Limitations and future work}
While our work presents a general framework for the generation of long-context retrieval benchmarks from any labeled time-series dataset, domains such as industrial sensor streams and climate signals remain future work. 
Our needle-insertion protocol is applied to accelerometer signals; extending it to additional domains where local-segment independence holds is a natural follow-up. A complementary direction is to ablate the protocol to understand the root cause of its performance deviation relative to the naturally-sampled datasets. 
Finally, while ARTS is conceived as a naive agentic approach for time-series retrieval, its dependence on a frontier API model (GPT-5.4) for the orchestrator confounds clean attribution of its gains to the agentic decoupling itself; an open-weights orchestrator at parity with TSLM backbones and further improvements in the architecture are left as future work.

%%%%%%%%%%%%%%%%%%%%%%%%%%%%%%%%%%%%%%%%%%%%%%%%%%%%%%%%%%%%%%%%%%%%%%%%%%%%%%%
% REFERENCES
%%%%%%%%%%%%%%%%%%%%%%%%%%%%%%%%%%%%%%%%%%%%%%%%%%%%%%%%%%%%%%%%%%%%%%%%%%%%%%%

\newpage
\bibliography{references}

@article{singh2025openaigpt5card,
  title={Openai gpt-5 system card},
  author={Singh, Aaditya and Fry, Adam and Perelman, Adam and Tart, Adam and Ganesh, Adi and El-Kishky, Ahmed and McLaughlin, Aidan and Low, Aiden and Ostrow, AJ and Ananthram, Akhila and others},
  journal={arXiv Preprints},
  year={2025},
  doi={10.48550/arXiv.2601.03267}
}

@article{grattafiori2024llama3,
  title={The Llama 3 Herd of Models},
  author={Grattafiori, Aaron and others},
  journal={arXiv Preprints},
  year={2024},
  doi={10.48550/arXiv.2407.21783}
}

@article{bulling2014tutorial,
  author    = {Bulling, Andreas and Blanke, Ulf and Schiele, Bernt},
  title     = {A Tutorial on Human Activity Recognition Using Body-Worn Inertial Sensors},
  journal   = {{ACM} Computing Surveys},
  volume    = {46},
  number    = {3},
  pages     = {1--33},
  year      = {2014},
  doi       = {10.1145/2499621}
}

@book{berry2017aasm,
  title     = {The {AASM} Manual for the Scoring of Sleep and Associated Events: Rules, Terminology and Technical Specifications, Version 2.4},
  author    = {Berry, Richard B. and Brooks, Rita and Gamaldo, Charlene and Harding, Susan M. and Lloyd, Robin M. and Quan, Stuart F. and Troester, Matthew T. and Vaughn, Bradley V.},
  publisher = {American Academy of Sleep Medicine},
  address   = {Darien, IL},
  year      = {2017},
  doi={10.5664/jcsm.6576}
}

@article{hindricks2020esc,
  author    = {Hindricks, Gerhard and Potpara, Tatjana and Dagres, Nikolaos and Arbelo, Elena and Bax, Jeroen J. and Blomstr{\"o}m-Lundqvist, Carina and Boriani, Giuseppe and Castella, Manuel and Dan, Gheorghe-Andrei and Dilaveris, Polychronis E. and Fauchier, Laurent and Filippatos, Gerasimos and Kalman, Jonathan M. and La Meir, Mark and Lane, Deirdre A. and Lebeau, Jean-Pierre and Lettino, Maddalena and Lip, Gregory Y. H. and Pinto, Fausto J. and Thomas, G. Neil and Valgimigli, Marco and Van Gelder, Isabelle C. and Van Putte, Bart P. and Watkins, Caroline L.},
  title     = {2020 {ESC} Guidelines for the diagnosis and management of atrial fibrillation developed in collaboration with the {European Association for Cardio-Thoracic Surgery {(EACTS)}}},
  journal   = {European Heart Journal},
  volume    = {42},
  number    = {5},
  pages     = {373--498},
  year      = {2021},
  doi       = {10.1093/eurheartj/ehaa612}
}

@inproceedings{vaswani2017attention,
  title={Attention Is All You Need},
  author={Vaswani, Ashish and Shazeer, Noam and Parmar, Niki and Uszkoreit, Jakob and Jones, Llion and Gomez, Aidan N. and Kaiser, {\L}ukasz and Polosukhin, Illia},
  booktitle={Advances in Neural Information Processing Systems {(NeurIPS)}},
  year={2017},
  url={https://proceedings.neurips.cc/paper/2017/hash/3f5ee243547dee91fbd053c1c4a845aa-Abstract.html}
}

@inproceedings{alayrac2022flamingo,
  title={Flamingo: a Visual Language Model for Few-Shot Learning},
  author={Alayrac, Jean-Baptiste and Donahue, Jeff and Luc, Pauline and Miech, Antoine and Barr, Iain and Hasson, Yana and Lenc, Karel and Mensch, Arthur and Millican, Katie and Reynolds, Malcolm and Ring, Roman and Rutherford, Eliza and Cabi, Serkan and Han, Tengda and Gong, Zhitao and Samangooei, Sina and Monteiro, Marianne and Menick, Jacob and Borgeaud, Sebastian and Brock, Andrew and Nematzadeh, Aida and Sharifzadeh, Sahand and Binkowski, Mikolaj and Barreira, Ricardo and Vinyals, Oriol and Zisserman, Andrew and Simonyan, Karen},
  booktitle={Advances in Neural Information Processing Systems {(NeurIPS)}},
  volume={35},
  pages={23716--23736},
  year={2022},
  url={https://proceedings.neurips.cc/paper/2022/hash/960a172bc7fbf0177ccccbb411a7d800-Abstract-Conference.html}
}

@inproceedings{liu2023visualinstructiontuning,
      title={Visual Instruction Tuning}, 
      author={Haotian Liu and Chunyuan Li and Qingyang Wu and Yong Jae Lee},
      year={2023},
      booktitle={Advances in Neural Information Processing Systems {(NeurIPS)}},
      volume={36},
      url={https://proceedings.neurips.cc/paper/2023/hash/6dcf277ea32ce3288914faf369fe6de0-Abstract-Conference.html}, 
}

@inproceedings{jaegle2021perceiver,
  title={Perceiver: General Perception with Iterative Attention},
  author={Jaegle, Andrew and Gimeno, Felix and Brock, Andrew and Zisserman, Andrew and Vinyals, Oriol and Carreira, Jo{\~a}o},
  booktitle={International Conference on Machine Learning {(ICML)}},
  pages={4651--4664},
  year={2021},
  organization={PMLR},
  url={https://proceedings.mlr.press/v139/jaegle21a.html}
}

@inproceedings{nie2023time,
  title={A Time Series is Worth 64 Words: Long-term Forecasting with Transformers},
  author={Nie, Yuqi and Nguyen, Nam H and Sinthong, Phanwadee and Kalagnanam, Jayant},
  booktitle={International Conference on Learning Representations {(ICLR)}},
  year={2023},
  url={https://openreview.net/forum?id=Jbdc0vTOcol}
}

@article{langer2025opentslm,
  title={{OpenTSLM}: Time-Series Language Models for Reasoning over Multivariate Medical Text- and Time-Series Data},
  author={Langer, Patrick and others},
  journal={arXiv Preprints},
  year={2025},
  doi={10.48550/arXiv.2510.02410}
}

@inproceedings{wang2025itformer,
  title={ITFormer: Bridging Time Series and Natural Language for Multi-Modal QA with Large-Scale Multitask Dataset},
  author={Wang, Yilin and Lei, Peixuan and Song, Jie and Hao, Yuzhe and Chen, Tao and Zhang, Yuxuan and Jia, Lei and Li, Yuanxiang and Wei, Zhongyu},
  booktitle={Proceedings of the 42nd International Conference on Machine Learning {(ICML)}},
  year={2025},
  url={https://icml.cc/virtual/2025/poster/45847}
}

@article{Xie_2025,
   title={ChatTS: Aligning Time Series with LLMs via Synthetic Data for Enhanced Understanding and Reasoning},
   volume={18},
   ISSN={2150-8097},
   DOI={10.14778/3742728.3742735},
   number={8},
   journal={Proceedings of the VLDB Endowment},
   publisher={Association for Computing Machinery {(ACM)}},
   author={Xie, Zhe and Li, Zeyan and He, Xiao and Xu, Longlong and Wen, Xidao and Zhang, Tieying and Chen, Jianjun and Shi, Rui and Pei, Dan},
   year={2025},
   month=Apr, 
   pages={2385–2398} 
}

@inproceedings{wang2024chattimeunifiedmultimodaltime,
  title={Chattime: A unified multimodal time series foundation model bridging numerical and textual data},
  author={Wang, Chengsen and Qi, Qi and Wang, Jingyu and Sun, Haifeng and Zhuang, Zirui and Wu, Jinming and Zhang, Lei and Liao, Jianxin},
  booktitle={Proceedings of the AAAI Conference on Artificial Intelligence},
  volume={39},
  issue={12},
  pages={12694--12702},
  year={2025},
  doi={10.1609/aaai.v39i12.33384 }
}

@article{bagnall2018ueamultivariatetimeseries,
  title={The UEA multivariate time series classification archive, 2018},
  author={Bagnall, Anthony and Dau, Hoang Anh and Lines, Jason and Flynn, Michael and Large, James and Bostrom, Aaron and Southam, Paul and Keogh, Eamonn},
  journal={arXiv Preprints},
  year={2018},
  doi={10.48550/arXiv.1811.00075}
}

@article{UCRArchive2018,
  title={The UCR time series archive},
  author={Dau, Hoang Anh and Bagnall, Anthony and Kamgar, Kaveh and Yeh, Chin-Chia Michael and Zhu, Yan and Gharghabi, Shaghayegh and Ratanamahatana, Chotirat Ann and Keogh, Eamonn},
  journal={IEEE/CAA Journal of Automatica Sinica},
  volume={6},
  number={6},
  pages={1293--1305},
  year={2019},
  publisher={IEEE},
  doi={10.1109/JAS.2019.1911747}
}

@inproceedings{yue2024mmmumassivemultidisciplinemultimodal,
  title={Mmmu: A massive multi-discipline multimodal understanding and reasoning benchmark for expert {AGI}},
  author={Yue, Xiang and Ni, Yuansheng and Zhang, Kai and Zheng, Tianyu and Liu, Ruoqi and Zhang, Ge and Stevens, Samuel and Jiang, Dongfu and Ren, Weiming and Sun, Yuxuan and others},
  booktitle={Proceedings of the IEEE/CVF conference on computer vision and pattern recognition},
  pages={9556--9567},
  year={2024},
  url={https://openaccess.thecvf.com/content/CVPR2024/html/Yue_MMMU_A_Massive_Multi-discipline_Multimodal_Understanding_and_Reasoning_Benchmark_for_CVPR_2024_paper.html}
}

@inproceedings{singh2019vqamodelsread,
  title={Towards vqa models that can read},
  author={Singh, Amanpreet and Natarajan, Vivek and Shah, Meet and Jiang, Yu and Chen, Xinlei and Batra, Dhruv and Parikh, Devi and Rohrbach, Marcus},
  booktitle={Proceedings of the IEEE/CVF conference on computer vision and pattern recognition},
  pages={8317--8326},
  year={2019},
  url={https://openaccess.thecvf.com/content_CVPR_2019/html/Singh_Towards_VQA_Models_That_Can_Read_CVPR_2019_paper.html}
}

@inproceedings{li2025tsfmbenchcomprehensiveunifiedbenchmark,
  title={Tsfm-bench: A comprehensive and unified benchmark of foundation models for time series forecasting},
  author={Li, Zhe and Qiu, Xiangfei and Chen, Peng and Wang, Yihang and Cheng, Hanyin and Shu, Yang and Hu, Jilin and Guo, Chenjuan and Zhou, Aoying and Jensen, Christian S and others},
  booktitle={Proceedings of the 31st ACM SIGKDD Conference on Knowledge Discovery and Data Mining V. 2},
  pages={5595--5606},
  year={2025},
  doi={10.1145/3711896.3737442}
}

@article{aksu2024giftevalbenchmarkgeneraltime,
  title={Gift-eval: A benchmark for general time series forecasting model evaluation},
  author={Aksu, Taha and Woo, Gerald and Liu, Juncheng and Liu, Xu and Liu, Chenghao and Savarese, Silvio and Xiong, Caiming and Sahoo, Doyen},
  journal={arXiv Preprints},
  year={2024},
  doi={10.48550/arXiv.2410.10393}
}

@article{kuratov2024babilongtestinglimitsllms,
  title={Babilong: Testing the limits of llms with long context reasoning-in-a-haystack},
  author={Kuratov, Yuri and Bulatov, Aydar and Anokhin, Petr and Rodkin, Ivan and Sorokin, Dmitry and Sorokin, Artyom and Burtsev, Mikhail},
  journal={Advances in Neural Information Processing Systems ({NeurIPS})},
  volume={37},
  pages={106519--106554},
  year={2024},
  doi={10.52202/079017-3381}
}

@article{liu2023lostmiddle,
  title={Lost in the middle: How language models use long contexts},
  author={Liu, Nelson F and Lin, Kevin and Hewitt, John and Paranjape, Ashwin and Bevilacqua, Michele and Petroni, Fabio and Liang, Percy},
  journal={Transactions of the association for computational linguistics},
  volume={12},
  pages={157--173},
  year={2024},
  doi={10.1162/tacl_a_00638}
}

@inproceedings{wang2024mmniah,
  title={Needle In A Multimodal Haystack},
  author={Wang, Weiyun and Zhang, Shuibo and Ren, Yiming and Duan, Yuchen and Li, Tiantong and Liu, Shuo and Hu, Mengkang and Chen, Zhe and Zhang, Kaipeng and Lu, Lewei and others},
  booktitle={Advances in Neural Information Processing Systems ({NeurIPS})},
  year={2024},
  doi={10.52202/079017-0649}
}

@article{he2025audiomarathon,
  title={Audiomarathon: A comprehensive benchmark for long-context audio understanding and efficiency in audio llms},
  author={He, Peize and Wen, Zichen and Wang, Yubo and Wang, Yuxuan and Liu, Xiaoqian and Huang, Jiajie and Lei, Zehui and Gu, Zhuangcheng and Jin, Xiangqi and Yang, Jiabing and others},
  journal={arXiv Preprints},
  year={2025},
  doi={10.48550/arXiv.2510.07293}
}

@article{gu2023mamba,
  title={Mamba: Linear-Time Sequence Modeling with Selective State Spaces},
  author={Gu, Albert and Dao, Tri},
  journal={arXiv Preprints},
  year={2023},
  doi={10.48550/arXiv.2312.00752}
}

@inproceedings{behrouz2025titans,
  title={Titans: Learning to Memorize at Test Time},
  author={Behrouz, Ali and Zhong, Peilin and Mirrokni, Vahab},
  booktitle={Advances in Neural Information Processing Systems {(NeurIPS)}},
  year={2025},
  url={https://neurips.cc/virtual/2025/loc/san-diego/poster/119639}
}

@article{chan2024capture24,
  title={{CAPTURE-24}: A large dataset of wrist-worn activity tracker data collected in the wild for human activity recognition},
  author={Chan, Shing and Yuan, Hang and Tong, Catherine and Acquah, Aidan and Schonfeldt, Abram and Gershuny, Jonathan and Doherty, Aiden},
  journal={Scientific Data},
  volume={11},
  number={1},
  pages={1135},
  year={2024},
  publisher={Nature Publishing Group},
  doi={10.1038/s41597-024-03960-3}
}

@inproceedings{ghassemi2018you,
  author    = {Ghassemi, MM and Moody, B and Lehman, L and Song, C and Li, Q and Sun, H and Westover, M and Clifford, GD},
  title     = {You Snooze, You Win: the {PhysioNet}/Computing in Cardiology Challenge 2018},
  booktitle = {2018 Computing in Cardiology Conference {(CinC)}},
  year      = {2018},
  pages     = {1--4},
  doi       = {10.22489/CinC.2018.049},
}

@article{petrutiu2007ltaf,
     author  = {Petrutiu, Simona and Sahakian, Alan V. and Swiryn, Steven},
     title   = {Abrupt Changes in Fibrillatory Wave Characteristics at the Termination of Paroxysmal Atrial Fibrillation in Humans},
     journal = {Europace},
     volume  = {9},
     number  = {7},
     pages   = {466--470},
     year    = {2007},
     doi     = {10.1093/europace/eum096},
}

@misc{kamradt2023needle,
  author = {Kamradt, Greg},
  title = {Needle in a Haystack - Pressure Testing {LLMs}},
  year = {2023},
  howpublished = {GitHub: \url{https://github.com/gkamradt/LLMTest_NeedleInAHaystack}},
  note = {Accessed: 2025}
}

@article{cai2024timeseriesexam,
  title={TimeSeriesExam: A Time Series Understanding Exam},
  author={Cai, Yifu and Choudhry, Arjun and Hu, Xing and Dubrawski, Artur},
  journal={arXiv Preprints},
  year={2024},
  doi={10.48550/arXiv.2410.14752}
}

@inproceedings{wang2025multimodal,
  title={Multimodal needle in a haystack: Benchmarking long-context capability of multimodal large language models},
  author={Wang, Hengyi and Shi, Haizhou and Tan, Shiwei and Qin, Weiyi and Wang, Wenyuan and Zhang, Tunyu and Nambi, Akshay and Ganu, Tanuja and Wang, Hao},
  booktitle={Proceedings of the 2025 Conference of the Nations of the Americas Chapter of the Association for Computational Linguistics: Human Language Technologies (Volume 1: Long Papers)},
  pages={3221--3241},
  year={2025},
  doi={10.18653/v1/2025.naacl-long.166}
}

@inproceedings{song2024milebench,
  title={Milebench: Benchmarking MLLMs in long context},
  author={Dingjie, Song and Chen, Shunian and Chen, Guiming Hardy and Yu, Fei and Wan, Xiang and Wang, Benyou},
  booktitle={First Conference on Language Modeling},
  year={2024},
  url={https://openreview.net/forum?id=Uhwze2LEwq}
}

@article{kelly2015ukdale,
  author  = {Kelly, Jack and Knottenbelt, William},
  title   = {The {UK-DALE} dataset, domestic appliance-level electricity demand and whole-house demand from five {UK} homes},
  journal = {Scientific Data},
  volume  = {2},
  number  = {1},
  pages   = {150007},
  year    = {2015},
  doi     = {10.1038/sdata.2015.7}
}

@inproceedings{chen2016xgboost,
  title={{XGBoost}: A scalable tree boosting system},
  author={Chen, Tianqi and Guestrin, Carlos},
  booktitle={Proceedings of the 22nd {ACM SIGKDD} International Conference on Knowledge Discovery and Data Mining},
  pages={785--794},
  year={2016},
  doi={10.1145/2939672.2939785}
}

@article{willetts2018statistical,
  title={Statistical machine learning of sleep and physical activity phenotypes from sensor data in 96,220 {UK Biobank} participants},
  author={Willetts, Matthew and Hollowell, Sven and Aslett, Louis and Holmes, Chris and Doherty, Aiden},
  journal={Scientific Reports},
  volume={8},
  number={1},
  pages={7961},
  year={2018},
  publisher={Nature Publishing Group},
  doi={10.1038/s41598-018-26174-1}
}

@inproceedings{yao2023react,
  title     = {{ReAct}: Synergizing Reasoning and Acting in Language Models},
  author    = {Yao, Shunyu and Zhao, Jeffrey and Yu, Dian and Du, Nan and Shafran, Izhak and Narasimhan, Karthik and Cao, Yuan},
  booktitle = {International Conference on Learning Representations (ICLR)},
  year      = {2023},
  url       = {https://openreview.net/forum?id=WE_vluYUL-X}
}

@article{zhang2026recursivelanguagemodels,
  title={Recursive language models},
  author={Zhang, Alex L and Kraska, Tim and Khattab, Omar},
  journal={arXiv Preprints},
  year={2025},
  doi={10.48550/arXiv.2512.24601}
}

@article{chen2023memwalker,
  title={Walking down the memory maze: Beyond context limit through interactive reading},
  author={Chen, Howard and Pasunuru, Ramakanth and Weston, Jason and Celikyilmaz, Asli},
  journal={arXiv Preprints},
  year={2023},
  doi={10.48550/arXiv.2310.05029}
}

@inproceedings{wang2024videoagent,
  title={Videoagent: Long-form video understanding with large language model as agent},
  author={Wang, Xiaohan and Zhang, Yuhui and Zohar, Orr and Yeung-Levy, Serena},
  booktitle={European Conference on Computer Vision},
  pages={58--76},
  year={2024},
  organization={Springer},
  doi={10.1007/978-3-031-72989-8_4}
}

@article{ansari2025chronos2,
  title={Chronos-2: From univariate to universal forecasting},
  author={Ansari, Abdul Fatir and Shchur, Oleksandr and K{\"u}ken, Jaris and Auer, Andreas and Han, Boran and Mercado, Pedro and Rangapuram, Syama Sundar and Shen, Huibin and Stella, Lorenzo and Zhang, Xiyuan and others},
  journal={arXiv Preprints},
  year={2025},
  doi={10.48550/arXiv.2510.15821}
}

@article{yuan2024oxwearables,
  title={Self-supervised learning for human activity recognition using 700,000 person-days of wearable data},
  author={Yuan, Hang* and Chan, Shing* and Creagh, Andrew P and Tong, Catherine and Acquah, Aidan and Clifton, David A and Doherty, Aiden},
  journal={NPJ digital medicine},
  volume={7},
  number={1},
  pages={91},
  year={2024},
  publisher={Nature Publishing Group UK London},
  doi={10.1038/s41746-024-01062-3}
}

@article{benidis2022deep,
  title={Deep learning for time series forecasting: Tutorial and literature survey},
  author={Benidis, Konstantinos and Rangapuram, Syama Sundar and Flunkert, Valentin and Wang, Yuyang and Maddix, Danielle and Turkmen, Caner and Gasthaus, Jan and Bohlke-Schneider, Michael and Salinas, David and Stella, Lorenzo and others},
  journal={ACM Computing Surveys},
  volume={55},
  number={6},
  pages={1--36},
  year={2022},
  publisher={ACM New York, NY},
  doi={10.1145/3533382}
}

@article{torres2021deep,
  title={Deep learning for time series forecasting: a survey},
  author={Torres, Jos{\'e} F and Hadjout, Dalil and Sebaa, Abderrazak and Mart{\'\i}nez-{\'A}lvarez, Francisco and Troncoso, Alicia},
  journal={Big data},
  volume={9},
  number={1},
  pages={3--21},
  year={2021},
  publisher={SAGE Publications Sage CA: Los Angeles, CA},
  doi={10.1109/JSEN.2019.2923982},
}

@article{zhang2024large,
  title={Large language models for time series: A survey},
  author={Zhang, Xiyuan and Chowdhury, Ranak Roy and Gupta, Rajesh K and Shang, Jingbo},
  journal={arXiv Preprints},
  year={2024},
  doi={10.48550/arXiv.2402.01801}
}

@article{abdullahi2025time,
  title={Time-series large language models: A systematic review of state-of-the-art},
  author={Abdullahi, Shamsu and Danyaro, Kamaluddeen Usman and Zakari, Abubakar and Aziz, Izzatdin Abdul and Zawawi, Noor Amila Wan Abdullah and Adamu, Shamsuddeen},
  journal={IEEE Access},
  volume={13},
  pages={30235--30261},
  year={2025},
  publisher={IEEE},
  doi={10.1109/ACCESS.2025.3535782}
}

@inproceedings{
loshchilov2018decoupled,
title={Decoupled Weight Decay Regularization},
author={Ilya Loshchilov and Frank Hutter},
booktitle={International Conference on Learning Representations},
year={2019},
url={https://openreview.net/forum?id=Bkg6RiCqY7},
}

@inproceedings{yeche2021hirid,
  title={HiRID-ICU-Benchmark -- A Comprehensive Machine Learning Benchmark on High-resolution ICU Data},
  author={Y{\`e}che, Hugo and Kuznetsova, Rita and Zimmermann, Marc and H{\"u}ser, Matthias and Lyu, Xinrui and Faltys, Martin and R{\"a}tsch, Gunnar},
  booktitle={Thirty-fifth Conference on Neural Information Processing Systems Datasets and Benchmarks Track (Round 1)},
  year={2021},
  url={https://arxiv.org/abs/2111.08536}
}

@inproceedings{aircraftfailues,
  author={Saxena, Abhinav and Goebel, Kai and Simon, Don and Eklund, Neil},
  booktitle={2008 International Conference on Prognostics and Health Management}, 
  title={Damage propagation modeling for aircraft engine run-to-failure simulation}, 
  year={2008},
  volume={},
  number={},
  pages={1-9},
  doi={10.1109/PHM.2008.4711414}
  }

@article{era5climate,
author = {Hersbach, Hans and Bell, Bill and Berrisford, Paul and Hirahara, Shoji and Horányi, András and Muñoz Sabater, Joaquín and Nicolas, Julien and Peubey, Carole and Radu, Raluca and Schepers, Dinand and Simmons, Adrian and Soci, Cornel and Abdalla, Saleh and Abellan, Xavier and Balsamo, Gianpaolo and Bechtold, Peter and Biavati, Gionata and Bidlot, Jean and Bonavita, Massimo and Thépaut, J.-N},
year = {2020},
month = {06},
pages = {},
title = {The ERA5 global reanalysis},
volume = {146},
journal = {Quarterly Journal of the Royal Meteorological Society},
doi = {10.1002/qj.3803}
}

@article{wang2021ecgcwtcnn,
  title   = {Automatic {ECG} Classification Using Continuous Wavelet Transform and Convolutional Neural Network},
  author  = {Wang, Tao and Lu, Changhua and Sun, Yining and Yang, Mei and Liu, Chun and Ou, Chunsheng},
  journal = {Entropy},
  volume  = {23},
  number  = {1},
  pages   = {119},
  year    = {2021},
  doi     = {10.3390/e23010119}
}

@article{li2022ecghybrid,
  title   = {Classification of {ECG} based on Hybrid Features using {CNNs} for Wearable Applications},
  author  = {Li, Xiaolin and Xiang, Fang and Panicker, Rajesh and Cardiff, Barry and John, Deepu},
  journal = {arXiv preprint arXiv:2206.07648},
  year    = {2022},
  doi     = {10.48550/arXiv.2206.07648}
}

@article{hannun2019cardiologist,
  title   = {Cardiologist-level arrhythmia detection and classification in ambulatory electrocardiograms using a deep neural network},
  author  = {Hannun, Awni Y. and Rajpurkar, Pranav and Haghpanahi, Masoumeh and Tison, Geoffrey H. and Bourn, Codie and Turakhia, Mintu P. and Ng, Andrew Y.},
  journal = {Nature Medicine},
  volume  = {25},
  number  = {1},
  pages   = {65--69},
  year    = {2019},
  publisher = {Nature Publishing Group},
  doi     = {10.1038/s41591-018-0268-3}
}
\bibliographystyle{plainnat}

%%%%%%%%%%%%%%%%%%%%%%%%%%%%%%%%%%%%%%%%%%%%%%%%%%%%%%%%%%%%%%%%%%%%%%%%%%%%%%%
% APPENDIX
%%%%%%%%%%%%%%%%%%%%%%%%%%%%%%%%%%%%%%%%%%%%%%%%%%%%%%%%%%%%%%%%%%%%%%%%%%%%%%%
\newpage
\appendix
% Small introduction about the appendix

%%%%%%%%%%%%%%%%%%%%%% RESULTS %%%%%%%%%%%%%%%%%%%%%%%%%%%%%%%%%%

\section{TS-Haystack detailed results}
\label{app:per_domain_results}

In the following, we report the per-task accuracy underlying the aggregate plots in Section~\ref{sec:results}, with one table per source dataset (Capture24, Sleep PSG, LTAF and Uk Dale). All tables follow the same conventions. Cells marked ``OOM'' indicate that the corresponding model exhausted GPU memory at training or inference time for that context length and was not evaluated. Per (task, length) row, the best result across the five evaluated methods is in \textbf{bold}; reference baselines are not eligible for bolding. Context length coverage varies by dataset, reflecting the natural durations of each source corpus and the task instantiation choices documented in Appendix~\ref{app:ts_haystack}.

The per-task tables present complementary findings to the aggregate results from Section~\ref{sec:results} worth surfacing.

\paragraph{ARTS leads on time-stamp retrieval; single-pass models lead on prior-dominated tasks.} ARTS performs best at long context lengths on Localization, Counting, Comparison, Multi-hop, and Anomaly Localization across datasets — the tasks where the answer must be specifically retrieved from the sample. On Antecedent and State Query, the picture inverts: Flamingo holds 0.90--0.97 across all four lengths on LTAF State Query, and 0.68--0.75 across Sleep Antecedent, while ARTS sits below in both.

We hypothesize that single-pass models internalize the marginal label distribution and Markovian priors over label transitions, answering correctly without per-segment perception. This is consistent with ARTS, whose inference-only agentic loop does not learn these priors and instead relies on bottom-up tool calls. Additionally, in Capture24, where needle insertion balances the weight of these prior coefficients, ARTS outperforms all TSLMs across context lengths. Nonetheless, further experiments that destabilize these priors and isolate the performance of TSLMs on these tasks remain future work to explain this performance gap between localization-heavy and prior-informed tasks.

\paragraph{Time-interval scoring localizes the bottleneck.} The collapse to near-zero on Localization and Multi-hop at longer samples is uniform across all single-pass models. ARTS preserves better accuracy on these tasks by decoupling timeseries classification and reasoning from the sample's context length, but also shows the steepest \emph{relative} degradation, leaving meaningful room for future work on long horizon time-series retrieval.

%%%%%%%%%%%%%%%%%%%%%% STATISTICAL VALIDATION %%%%%%%%%%%%%%%%%%%%%%%%%%%%%%%%%%

\subsection{Statistical significance of context-length degradation and model differences}
\label{app:significance}

To complement the descriptive trends in Section~\ref{sec:results} and the
per-task tables below, we report paired-difference tests on
the per-cell test accuracies. We use Wilcoxon signed-rank tests (two-sided)
because per-cell accuracies are bounded proportions over $150$ test samples
and are not jointly Gaussian across tasks. Within each family and dataset we
control the family-wise error rate with Holm-Bonferroni; we report the
adjusted $p$-values ($p_{\text{H}}$) alongside raw $p$-values.

\paragraph{Context-length degradation.}
For each (dataset, model) we pair accuracies across the $10$ tasks at the
shortest and longest context lengths at which the model is evaluable, and
ask whether accuracy at long context is lower. Length pairs are chosen to
maximize the number of paired tasks: when the extremes do not share enough
tasks (Sleep PSG, where existence and anomaly-detection are not instantiated
at $9$\,h), we fall back to the widest pair with $n\!\geq\!5$ paired tasks
(here $15$\,min vs.\ $9$\,h). Results are reported in
Table~\ref{tab:sig_context_length}. ARTS shows statistically significant
degradation on Sleep PSG ($p_{\text{H}}\!=\!0.023$, $\Delta\!=\!{+}0.205$),
LTAF ($p_{\text{H}}\!=\!0.023$, $\Delta\!=\!{+}0.149$), and UK-DALE
($p_{\text{H}}\!=\!0.039$, $\Delta\!=\!{+}0.169$); on Capture24 no model
degrades significantly, consistent with the qualitatively flat decay
profile reported in Section~\ref{sec:results}. Flamingo and ITFormer trend
downward on Sleep and UK-DALE but lose significance after correction,
reflecting their lower absolute accuracy and correspondingly smaller
absolute drops rather than greater long-context robustness.

\begin{table}[H]
\centering
\small
\setlength{\tabcolsep}{5pt}
\renewcommand{\arraystretch}{1.05}
\caption{\textbf{Context-length degradation per (dataset, model).} Paired
Wilcoxon signed-rank test across tasks, comparing accuracy at the shortest
($\bar{a}_{\text{s}}$) and longest ($\bar{a}_{\text{l}}$) context lengths
where the model has data, computed over the $n$ paired tasks.
$\Delta = \bar{a}_{\text{s}} - \bar{a}_{\text{l}}$; positive $\Delta$
indicates degradation with increasing context. $p_{\text{H}}$ is the
Holm-Bonferroni-adjusted $p$-value within each dataset; values $\le 0.05$
are in \textbf{bold}. ChatTS and ChatTime are evaluable only at a single
context length on Capture24, Sleep, and LTAF, so no degradation test is
possible there.}
\label{tab:sig_context_length}
\begin{tabular}{@{}llccccccc@{}}
\toprule
Dataset & Model & Short & Long & $n$ & $\bar{a}_{\text{s}}$ & $\bar{a}_{\text{l}}$ & $\Delta$ & $p_{\text{H}}$ \\
\midrule
\multirow{3}{*}{Capture24}
 & Flamingo & 100\,s & 2\,h & 10 & 0.315 & 0.279 & $+0.036$ & $0.492$ \\
 & ITFormer & 100\,s & 2\,h & 10 & 0.279 & 0.270 & $+0.009$ & $1.000$ \\
 & ARTS     & 100\,s & 2\,h & 10 & 0.410 & 0.415 & $-0.005$ & $1.000$ \\
\midrule
\multirow{3}{*}{Sleep PSG}
 & Flamingo & 15\,min & 9\,h & 8 & 0.361 & 0.267 & $+0.094$ & $0.062$ \\
 & ITFormer & 15\,min & 9\,h & 7 & 0.348 & 0.236 & $+0.111$ & $0.062$ \\
 & ARTS     & 15\,min & 9\,h & 8 & 0.444 & 0.239 & $+0.205$ & $\mathbf{0.023}$ \\
\midrule
\multirow{3}{*}{LTAF}
 & Flamingo & 100\,s & 2\,h & 8 & 0.412 & 0.358 & $+0.053$ & $0.500$ \\
 & ITFormer & 100\,s & 2\,h & 8 & 0.281 & 0.297 & $-0.016$ & $0.938$ \\
 & ARTS     & 100\,s & 2\,h & 8 & 0.560 & 0.411 & $+0.149$ & $\mathbf{0.023}$ \\
\midrule
\multirow{5}{*}{UK-DALE}
 & Flamingo & 15\,min & 24\,h & 10 & 0.377 & 0.273 & $+0.104$ & $0.156$ \\
 & ITFormer & 15\,min & 24\,h & 10 & 0.302 & 0.218 & $+0.084$ & $0.146$ \\
 & ChatTS   & 15\,min & 24\,h & 10 & 0.365 & 0.247 & $+0.118$ & $0.078$ \\
 & ChatTime & 15\,min & 2\,h  & 10 & 0.546 & 0.497 & $+0.050$ & $0.432$ \\
 & ARTS     & 15\,min & 24\,h & 10 & 0.603 & 0.434 & $+0.169$ & $\mathbf{0.039}$ \\
\bottomrule
\end{tabular}
\end{table}

\paragraph{Pairwise model differences.}
For each dataset we test, for every pair of models, whether their accuracies
differ across all (task, length) cells where both have evaluable data
(no OOM). Tests with fewer than five paired cells are omitted; this
excludes ChatTime everywhere except UK-DALE and excludes ChatTS on Sleep.
Results are reported in Table~\ref{tab:sig_model_pairwise}. ARTS
significantly outperforms Flamingo and ITFormer on every dataset
($p_{\text{H}} \le 0.024$ across all eight ARTS-vs-Flamingo and
ARTS-vs-ITFormer comparisons), and significantly outperforms ChatTS where
both are evaluable (Capture24 marginal at $p_{\text{H}}\!=\!0.107$; LTAF
$p_{\text{H}}\!=\!0.041$; UK-DALE $p_{\text{H}}\!\approx\!2 \cdot 10^{-5}$).
The one exception is ChatTime on UK-DALE, where ARTS and ChatTime are not
significantly different ($\Delta\!=\!{+}0.051$, $p_{\text{H}}\!=\!0.349$).
Among the single-pass TSLMs, Flamingo significantly outperforms ITFormer
on Sleep, LTAF, and UK-DALE ($p_{\text{H}}\!\le\!0.005$ in each), and
trends in the same direction on Capture24 ($p_{\text{H}}\!=\!0.107$), so
the ordering Flamingo $>$ ITFormer in the aggregate plots reflects a
genuine and consistent effect rather than a per-cell artifact.

\begin{table}[H]
\centering
\small
\setlength{\tabcolsep}{5pt}
\renewcommand{\arraystretch}{1.05}
\caption{\textbf{Pairwise model comparisons per dataset.} Paired Wilcoxon
signed-rank test across (task, context-length) cells where both models have
non-OOM accuracy. $\bar{a}_A$ and $\bar{a}_B$ are computed over the same
$n$ paired cells, so $\Delta = \bar{a}_A - \bar{a}_B$; positive favors
model $A$. $p_{\text{H}}$ is the Holm-Bonferroni-adjusted $p$-value within
each dataset; values $\le 0.05$ are in \textbf{bold}. Comparisons with
$n<5$ are omitted. ChatTime is evaluable beyond a single length only on
UK-DALE.}
\label{tab:sig_model_pairwise}
\begin{tabular}{@{}llccccc@{}}
\toprule
Dataset & Comparison ($A$ vs.\ $B$) & $n$ & $\bar{a}_A$ & $\bar{a}_B$ & $\Delta$ & $p_{\text{H}}$ \\
\midrule
\multirow{6}{*}{Capture24}
 & Flamingo vs.\ ITFormer & 40 & 0.290 & 0.270 & $+0.020$ & $0.107$ \\
 & Flamingo vs.\ ChatTS   & 10 & 0.315 & 0.267 & $+0.049$ & $0.107$ \\
 & ITFormer vs.\ ChatTS   & 10 & 0.279 & 0.267 & $+0.013$ & $0.492$ \\
 & ARTS vs.\ Flamingo     & 40 & 0.436 & 0.290 & $+0.146$ & $\mathbf{3.3\!\times\!10^{-4}}$ \\
 & ARTS vs.\ ITFormer     & 40 & 0.436 & 0.270 & $+0.167$ & $\mathbf{3.1\!\times\!10^{-5}}$ \\
 & ARTS vs.\ ChatTS       & 10 & 0.410 & 0.267 & $+0.144$ & $0.107$ \\
\midrule
\multirow{3}{*}{Sleep PSG}
 & Flamingo vs.\ ITFormer & 36 & 0.347 & 0.306 & $+0.041$ & $\mathbf{0.004}$ \\
 & ARTS vs.\ Flamingo     & 37 & 0.407 & 0.338 & $+0.069$ & $\mathbf{0.016}$ \\
 & ARTS vs.\ ITFormer     & 36 & 0.418 & 0.306 & $+0.112$ & $\mathbf{0.002}$ \\
\midrule
\multirow{6}{*}{LTAF}
 & Flamingo vs.\ ITFormer & 37 & 0.416 & 0.322 & $+0.094$ & $\mathbf{8.1\!\times\!10^{-6}}$ \\
 & Flamingo vs.\ ChatTS   & 10 & 0.472 & 0.387 & $+0.085$ & $0.055$ \\
 & ITFormer vs.\ ChatTS   & 10 & 0.340 & 0.387 & $-0.047$ & $0.074$ \\
 & ARTS vs.\ Flamingo     & 37 & 0.513 & 0.416 & $+0.097$ & $\mathbf{0.024}$ \\
 & ARTS vs.\ ITFormer     & 37 & 0.513 & 0.322 & $+0.191$ & $\mathbf{5.4\!\times\!10^{-6}}$ \\
 & ARTS vs.\ ChatTS       & 10 & 0.603 & 0.387 & $+0.216$ & $\mathbf{0.041}$ \\
\midrule
\multirow{10}{*}{UK-DALE}
 & Flamingo vs.\ ITFormer & 50 & 0.346 & 0.280 & $+0.066$ & $\mathbf{7.6\!\times\!10^{-7}}$ \\
 & Flamingo vs.\ ChatTS   & 50 & 0.346 & 0.322 & $+0.024$ & $\mathbf{0.046}$ \\
 & ChatTime vs.\ Flamingo & 30 & 0.523 & 0.376 & $+0.147$ & $\mathbf{2.2\!\times\!10^{-5}}$ \\
 & ITFormer vs.\ ChatTS   & 50 & 0.280 & 0.322 & $-0.042$ & $\mathbf{4.9\!\times\!10^{-4}}$ \\
 & ChatTime vs.\ ITFormer & 30 & 0.523 & 0.295 & $+0.227$ & $\mathbf{1.9\!\times\!10^{-5}}$ \\
 & ChatTime vs.\ ChatTS   & 30 & 0.523 & 0.355 & $+0.168$ & $\mathbf{2.0\!\times\!10^{-5}}$ \\
 & ARTS vs.\ Flamingo     & 50 & 0.519 & 0.346 & $+0.173$ & $\mathbf{2.7\!\times\!10^{-4}}$ \\
 & ARTS vs.\ ITFormer     & 50 & 0.519 & 0.280 & $+0.239$ & $\mathbf{1.1\!\times\!10^{-5}}$ \\
 & ARTS vs.\ ChatTS       & 50 & 0.519 & 0.322 & $+0.197$ & $\mathbf{2.0\!\times\!10^{-5}}$ \\
 & ARTS vs.\ ChatTime     & 30 & 0.573 & 0.523 & $+0.051$ & $0.349$ \\
\bottomrule
\end{tabular}
\end{table}

%%%%%%%%%%%%%%%%%%%%%%%%%%%%%%%%%%%%%%%%%%

\subsection{Per-dataset, per-task accuracy}
\label{app:results_breakdown}

We present per-dataset, per-task accuracy across context lengths in Tables~\ref{tab:capture24_per_task}--\ref{tab:ukdale_per_task} below.

%%%%%%%%%%%%%%%%%%%%%%%%%%%%%%% \subsection{Capture24 Haystack}
\begin{table}[!t]
\centering
\small
\setlength{\tabcolsep}{4pt}
\renewcommand{\arraystretch}{1.05}
\caption{\textbf{Capture24-Haystack: per-task accuracy across context lengths.}}
\label{tab:capture24_per_task}
\begin{tabular}{@{}ll ccccc cc@{}}
\toprule
& & \multicolumn{5}{c}{\textbf{Models}} & \multicolumn{2}{c}{\textbf{Reference}} \\
\cmidrule(lr){3-7} \cmidrule(l){8-9}
\textbf{Task} & \textbf{Length} & ChatTime & ChatTS & ITFormer & Flamingo & ARTS & \emph{Random} & \emph{Real Oracle} \\
\midrule
\multirow{4}{*}{Existence}
 & 100\,s  & OOM & 0.46          & 0.47          & 0.58          & \textbf{0.59} & 0.50 & 1.00 \\
 & 15\,min & OOM & OOM           & 0.48          & 0.54          & \textbf{0.65} & 0.50 & 1.00 \\
 & 1\,h    & OOM & OOM           & 0.50          & 0.44          & \textbf{0.66} & 0.50 & 1.00 \\
 & 2\,h    & OOM & OOM           & 0.48          & 0.47          & \textbf{0.67} & 0.50 & 1.00 \\
\midrule
\multirow{4}{*}{Localization}
 & 100\,s  & OOM & 0.00          & 0.11          & 0.12          & \textbf{0.41} & 0.08 & 1.00 \\
 & 15\,min & OOM & OOM           & 0.01          & 0.01          & \textbf{0.51} & 0.01 & 1.00 \\
 & 1\,h    & OOM & OOM           & 0.00          & 0.00          & \textbf{0.51} & 0.00 & 1.00 \\
 & 2\,h    & OOM & OOM           & 0.00          & 0.00          & \textbf{0.55} & 0.00 & 1.00 \\
\midrule
\multirow{4}{*}{Counting}
 & 100\,s  & OOM & \textbf{0.26} & 0.15          & 0.21          & 0.09          & 0.20 & 0.98 \\
 & 15\,min & OOM & OOM           & 0.19          & 0.19          & \textbf{0.22} & 0.20 & 0.97 \\
 & 1\,h    & OOM & OOM           & 0.18          & 0.19          & \textbf{0.31} & 0.20 & 0.95 \\
 & 2\,h    & OOM & OOM           & \textbf{0.25} & 0.19          & 0.24          & 0.20 & 0.97 \\
\midrule
\multirow{4}{*}{Ordering}
 & 100\,s  & OOM & 0.48          & 0.45          & 0.51          & \textbf{0.74} & 0.50 & 1.00 \\
 & 15\,min & OOM & OOM           & 0.53          & 0.46          & \textbf{0.83} & 0.50 & 1.00 \\
 & 1\,h    & OOM & OOM           & 0.53          & 0.47          & \textbf{0.79} & 0.50 & 1.00 \\
 & 2\,h    & OOM & OOM           & 0.55          & 0.48          & \textbf{0.74} & 0.50 & 1.00 \\
\midrule
\multirow{4}{*}{State Query}
 & 100\,s  & OOM & 0.24          & 0.27          & 0.28          & \textbf{0.39} & 0.13 & 0.74 \\
 & 15\,min & OOM & OOM           & 0.25          & 0.31          & \textbf{0.51} & 0.13 & 0.84 \\
 & 1\,h    & OOM & OOM           & 0.33          & 0.40          & \textbf{0.43} & 0.11 & 0.84 \\
 & 2\,h    & OOM & OOM           & 0.21          & 0.33          & \textbf{0.41} & 0.14 & 0.87 \\
\midrule
\multirow{4}{*}{Antecedent}
 & 100\,s  & OOM & 0.13          & 0.07          & \textbf{0.15} & \textbf{0.15} & 0.10 & 0.93 \\
 & 15\,min & OOM & OOM           & 0.07          & 0.14          & \textbf{0.27} & 0.10 & 0.97 \\
 & 1\,h    & OOM & OOM           & 0.08          & 0.12          & \textbf{0.27} & 0.10 & 0.90 \\
 & 2\,h    & OOM & OOM           & 0.17          & 0.10          & \textbf{0.21} & 0.10 & 0.97 \\
\midrule
\multirow{4}{*}{Comparison}
 & 100\,s  & OOM & 0.14          & 0.18          & 0.18          & \textbf{0.41} & 0.17 & 0.95 \\
 & 15\,min & OOM & OOM           & 0.13          & 0.19          & \textbf{0.43} & 0.18 & 0.97 \\
 & 1\,h    & OOM & OOM           & 0.18          & 0.23          & \textbf{0.39} & 0.18 & 0.95 \\
 & 2\,h    & OOM & OOM           & 0.13          & 0.18          & \textbf{0.35} & 0.09 & 0.95 \\
\midrule
\multirow{4}{*}{Multi-hop}
 & 100\,s  & OOM & 0.05          & 0.11          & 0.13          & \textbf{0.40} & 0.08 & 0.99 \\
 & 15\,min & OOM & OOM           & 0.02          & 0.02          & \textbf{0.41} & 0.01 & 0.99 \\
 & 1\,h    & OOM & OOM           & 0.01          & 0.01          & \textbf{0.31} & 0.00 & 0.99 \\
 & 2\,h    & OOM & OOM           & 0.00          & 0.00          & \textbf{0.24} & 0.00 & 0.97 \\
\midrule
\multirow{4}{*}{Anomaly Det.}
 & 100\,s  & OOM & 0.47          & 0.53          & 0.48          & \textbf{0.56} & 0.50 & 1.00 \\
 & 15\,min & OOM & OOM           & 0.51          & 0.53          & \textbf{0.63} & 0.50 & 1.00 \\
 & 1\,h    & OOM & OOM           & 0.47          & \textbf{0.53} & 0.51          & 0.50 & 1.00 \\
 & 2\,h    & OOM & OOM           & 0.46          & \textbf{0.56} & 0.52          & 0.50 & 1.00 \\
\midrule
\multirow{4}{*}{Anomaly Loc.}
 & 100\,s  & OOM & 0.43          & 0.45          & \textbf{0.50} & 0.38          & 0.04 & 0.99 \\
 & 15\,min & OOM & OOM           & 0.40          & \textbf{0.42} & 0.29          & 0.01 & 0.99 \\
 & 1\,h    & OOM & OOM           & 0.41          & \textbf{0.46} & 0.27          & 0.00 & 0.99 \\
 & 2\,h    & OOM & OOM           & 0.46          & \textbf{0.48} & 0.21          & 0.00 & 1.00 \\
\bottomrule
\end{tabular}
\end{table}

\clearpage

%%%%%%%%%%%%%%%%%%%%%%%%%%%%%%% \subsection{Sleep-Haystack}

\begin{table}[!t]
\centering
\small
\setlength{\tabcolsep}{4pt}
\renewcommand{\arraystretch}{1.05}
\caption{\textbf{Sleep-Haystack: per-task accuracy across context lengths.} Anomaly tasks span shorter horizons because arousal events are short and substantially more frequent than sleep-stage epochs (see Appendix~\ref{app:sleep-haystack-stats}). Cells marked ``---'' denote (task, length) pairs that are not instantiated for this dataset. Existence and Anomaly Detection are not instantiated at 9~h and (1~h, 9~h) respectively: the target class is near-deterministically present at such lengths.}
\label{tab:sleep_per_task}
\begin{tabular}{@{}ll ccccc cc@{}}
\toprule
& & \multicolumn{5}{c}{\textbf{Models}} & \multicolumn{2}{c}{\textbf{Reference}} \\
\cmidrule(lr){3-7} \cmidrule(l){8-9}
\textbf{Task} & \textbf{Length} & ChatTime & ChatTS & ITFormer & Flamingo & ARTS & \emph{Random} & \emph{Real Oracle} \\
\midrule
\multirow{4}{*}{Existence}
 & 15\,min & OOM & OOM & 0.49          & 0.46          & \textbf{0.89} & 0.50 & 0.99 \\
 & 1\,h    & OOM & OOM & 0.53          & 0.53          & \textbf{0.80} & 0.50 & 1.00 \\
 & 2\,h    & OOM & OOM & 0.44          & 0.47          & \textbf{0.63} & 0.50 & 1.00 \\
 & 9\,h    & ---  & ---  & ---           & ---           & ---           & ---   & ---   \\
\midrule
\multirow{4}{*}{Localization}
 & 15\,min & OOM & OOM & 0.06          & 0.05          & \textbf{0.34} & 0.17 & 1.00 \\
 & 1\,h    & OOM & OOM & 0.05          & 0.03          & \textbf{0.35} & 0.11 & 1.00 \\
 & 2\,h    & OOM & OOM & 0.01          & 0.01          & \textbf{0.13} & 0.05 & 1.00 \\
 & 9\,h    & OOM & OOM & 0.00          & 0.01          & \textbf{0.06} & 0.01 & 1.00 \\
\midrule
\multirow{4}{*}{Counting}
 & 15\,min & OOM & OOM & 0.51          & 0.49          & \textbf{0.51} & 0.14 & 0.97 \\
 & 1\,h    & OOM & OOM & \textbf{0.33} & \textbf{0.33} & \textbf{0.33} & 0.06 & 0.95 \\
 & 2\,h    & OOM & OOM & 0.15          & \textbf{0.18} & 0.10          & 0.04 & 0.95 \\
 & 9\,h    & OOM & OOM & 0.04          & \textbf{0.07} & 0.03          & 0.02 & 0.97 \\
\midrule
\multirow{4}{*}{Ordering}
 & 15\,min & OOM & OOM & 0.49          & 0.75          & \textbf{0.77} & 0.50 & 0.97 \\
 & 1\,h    & OOM & OOM & 0.58          & 0.71          & \textbf{0.81} & 0.50 & 1.00 \\
 & 2\,h    & OOM & OOM & 0.51          & \textbf{0.72} & 0.71          & 0.50 & 1.00 \\
 & 9\,h    & OOM & OOM & 0.47          & \textbf{0.76} & 0.67          & 0.50 & 0.99 \\
\midrule
\multirow{4}{*}{State Query}
 & 15\,min & OOM & OOM & 0.40          & 0.49          & \textbf{0.58} & 0.20 & 0.84 \\
 & 1\,h    & OOM & OOM & 0.42          & 0.43          & \textbf{0.62} & 0.20 & 0.95 \\
 & 2\,h    & OOM & OOM & 0.45          & \textbf{0.49} & 0.37          & 0.20 & 0.94 \\
 & 9\,h    & OOM & OOM & 0.42          & \textbf{0.50} & 0.36          & 0.20 & 0.92 \\
\midrule
\multirow{4}{*}{Antecedent}
 & 15\,min & OOM & OOM & \textbf{0.75} & \textbf{0.75} & 0.60          & 0.20 & 0.89 \\
 & 1\,h    & OOM & OOM & \textbf{0.75} & 0.73          & 0.53          & 0.20 & 0.91 \\
 & 2\,h    & OOM & OOM & 0.69          & \textbf{0.73} & 0.49          & 0.20 & 0.89 \\
 & 9\,h    & OOM & OOM & 0.68          & \textbf{0.73} & 0.45          & 0.20 & 0.89 \\
\midrule
\multirow{4}{*}{Comparison}
 & 15\,min & OOM & OOM & 0.18          & 0.26          & \textbf{0.39} & 0.21 & 0.85 \\
 & 1\,h    & OOM & OOM & 0.11          & 0.20          & \textbf{0.40} & 0.14 & 0.89 \\
 & 2\,h    & OOM & OOM & 0.04          & 0.07          & \textbf{0.36} & 0.10 & 0.91 \\
 & 9\,h    & OOM & OOM & 0.05          & 0.05          & \textbf{0.30} & 0.04 & 0.98 \\
\midrule
\multirow{4}{*}{Multi-hop}
 & 15\,min & OOM & OOM & 0.11          & 0.07          & \textbf{0.26} & 0.15 & 0.82 \\
 & 1\,h    & OOM & OOM & 0.03          & 0.05          & \textbf{0.17} & 0.07 & 0.73 \\
 & 2\,h    & OOM & OOM & 0.01          & 0.02          & \textbf{0.07} & 0.04 & 0.70 \\
 & 9\,h    & OOM & OOM & 0.01          & 0.01          & \textbf{0.02} & 0.01 & 0.68 \\
\midrule
\multirow{4}{*}{Anomaly Det.}
 & 100\,s  & OOM & OOM & 0.54          & 0.47          & \textbf{0.69} & 0.50 & 1.00 \\
 & 15\,min & OOM & OOM & 0.53          & 0.45          & \textbf{0.63} & 0.50 & 1.00 \\
 & 1\,h    & ---  & ---  & ---           & ---           & ---           & ---   & ---   \\
 & 9\,h    & ---  & ---  & ---           & ---           & ---           & ---   & ---   \\
\midrule
\multirow{4}{*}{Anomaly Loc.}
 & 100\,s  & OOM & OOM & 0.27          & 0.41          & \textbf{0.41} & 0.26 & 1.00 \\
 & 15\,min & OOM & OOM & 0.05          & 0.04          & \textbf{0.09} & 0.03 & 1.00 \\
 & 1\,h    & OOM & OOM & 0.00          & 0.00          & \textbf{0.09} & 0.01 & 1.00 \\
 & 9\,h    & OOM & OOM & 0.00          & 0.00          & \textbf{0.01} & 0.00 & 0.96 \\
\bottomrule
\end{tabular}
\end{table}

\clearpage

%%%%%%%%%%%%%%%%%%%%%%%%%%%%%%% \subsection{LTAF Haystack}

\begin{table}[!t]
\centering
\small
\setlength{\tabcolsep}{4pt}
\renewcommand{\arraystretch}{1.05}
\caption{\textbf{LTAF-Haystack: per-task accuracy across context lengths.} Cells marked ``---'' denote (task, length) pairs that are not instantiated for this dataset. Existence and Anomaly Detection are not instantiated at 2~h and (1~h, 2~h) respectively: the target classes are near-deterministically present at these horizons.}
\label{tab:ltaf_per_task}
\begin{tabular}{@{}ll ccccc cc@{}}
\toprule
& & \multicolumn{5}{c}{\textbf{Models}} & \multicolumn{2}{c}{\textbf{Reference}} \\
\cmidrule(lr){3-7} \cmidrule(l){8-9}
\textbf{Task} & \textbf{Length} & ChatTime & ChatTS & ITFormer & Flamingo & ARTS & \emph{Random} & \emph{Real Oracle} \\
\midrule
\multirow{4}{*}{Existence}
 & 100\,s  & OOM & 0.57          & 0.51          & 0.71          & \textbf{0.83} & 0.50 & 1.00 \\
 & 15\,min & OOM & OOM           & 0.53          & 0.75          & \textbf{0.87} & 0.50 & 1.00 \\
 & 1\,h    & OOM & OOM           & 0.55          & 0.69          & \textbf{0.84} & 0.50 & 1.00 \\
 & 2\,h    & ---  & ---           & ---           & ---           & ---           & ---   & ---   \\
\midrule
\multirow{4}{*}{Localization}
 & 100\,s  & OOM & 0.05          & 0.05          & 0.09          & \textbf{0.32} & 0.06 & 0.99 \\
 & 15\,min & OOM & OOM           & 0.01          & 0.01          & \textbf{0.29} & 0.02 & 0.99 \\
 & 1\,h    & OOM & OOM           & 0.00          & 0.01          & \textbf{0.28} & 0.02 & 0.99 \\
 & 2\,h    & OOM & OOM           & 0.00          & 0.01          & \textbf{0.26} & 0.01 & 0.99 \\
\midrule
\multirow{4}{*}{Counting}
 & 100\,s  & OOM & 0.23          & 0.24          & 0.40          & \textbf{0.49} & 0.10 & 0.99 \\
 & 15\,min & OOM & OOM           & 0.23          & 0.36          & \textbf{0.41} & 0.05 & 0.95 \\
 & 1\,h    & OOM & OOM           & 0.26          & 0.37          & \textbf{0.47} & 0.05 & 1.00 \\
 & 2\,h    & OOM & OOM           & 0.27          & 0.36          & \textbf{0.41} & 0.03 & 0.97 \\
\midrule
\multirow{4}{*}{Ordering}
 & 100\,s  & OOM & 0.71          & 0.67          & \textbf{0.77} & 0.65          & 0.50 & 1.00 \\
 & 15\,min & OOM & OOM           & 0.69          & \textbf{0.83} & 0.68          & 0.50 & 0.99 \\
 & 1\,h    & OOM & OOM           & 0.63          & \textbf{0.77} & 0.63          & 0.50 & 1.00 \\
 & 2\,h    & OOM & OOM           & 0.70          & \textbf{0.82} & 0.59          & 0.50 & 1.00 \\
\midrule
\multirow{4}{*}{State Query}
 & 100\,s  & OOM & 0.67          & 0.57          & \textbf{0.97} & 0.93          & 0.33 & 1.00 \\
 & 15\,min & OOM & OOM           & 0.65          & \textbf{0.90} & 0.84          & 0.25 & 1.00 \\
 & 1\,h    & OOM & OOM           & 0.73          & \textbf{0.91} & 0.85          & 0.14 & 1.00 \\
 & 2\,h    & OOM & OOM           & 0.76          & \textbf{0.91} & 0.85          & 0.25 & 0.98 \\
\midrule
\multirow{4}{*}{Antecedent}
 & 100\,s  & OOM & 0.47          & 0.57          & \textbf{0.61} & 0.59          & 0.14 & 1.00 \\
 & 15\,min & OOM & OOM           & 0.52          & \textbf{0.57} & \textbf{0.57} & 0.13 & 1.00 \\
 & 1\,h    & OOM & OOM           & \textbf{0.56} & 0.54          & 0.52          & 0.11 & 0.99 \\
 & 2\,h    & OOM & OOM           & 0.56          & \textbf{0.66} & 0.58          & 0.11 & 0.99 \\
\midrule
\multirow{4}{*}{Comparison}
 & 100\,s  & OOM & 0.19          & 0.06          & 0.14          & \textbf{0.47} & 0.25 & 0.90 \\
 & 15\,min & OOM & OOM           & 0.04          & 0.05          & \textbf{0.40} & 0.18 & 0.89 \\
 & 1\,h    & OOM & OOM           & 0.11          & 0.14          & \textbf{0.33} & 0.15 & 0.89 \\
 & 2\,h    & OOM & OOM           & 0.06          & 0.09          & \textbf{0.31} & 0.19 & 0.91 \\
\midrule
\multirow{4}{*}{Multi-hop}
 & 100\,s  & OOM & 0.23          & 0.10          & \textbf{0.30} & 0.28          & 0.23 & 0.87 \\
 & 15\,min & OOM & OOM           & 0.04          & 0.03          & \textbf{0.19} & 0.12 & 0.84 \\
 & 1\,h    & OOM & OOM           & 0.04          & 0.06          & \textbf{0.21} & 0.19 & 0.83 \\
 & 2\,h    & OOM & OOM           & 0.02          & 0.01          & \textbf{0.17} & 0.11 & 0.83 \\
\midrule
\multirow{4}{*}{Anomaly Det.}
 & 100\,s  & OOM & \textbf{0.75} & 0.65          & 0.72          & 0.72          & 0.50 & 1.00 \\
 & 15\,min & OOM & OOM           & 0.55          & \textbf{0.80} & 0.59          & 0.50 & 1.00 \\
 & 1\,h    & ---  & ---           & ---           & ---           & ---           & ---   & ---   \\
 & 2\,h    & ---  & ---           & ---           & ---           & ---           & ---   & ---   \\
\midrule
\multirow{4}{*}{Anomaly Loc.}
 & 100\,s  & OOM & 0.01          & 0.00          & 0.02          & \textbf{0.76} & 0.01 & 1.00 \\
 & 15\,min & OOM & OOM           & 0.00          & 0.00          & \textbf{0.51} & 0.00 & 1.00 \\
 & 1\,h    & OOM & OOM           & 0.00          & 0.00          & \textbf{0.19} & 0.00 & 0.91 \\
 & 2\,h    & OOM & OOM           & 0.00          & 0.00          & \textbf{0.12} & 0.00 & 0.87 \\
\bottomrule
\end{tabular}
\end{table}

\clearpage

%%%%%%%%%%%%%%%%%%%%%%%%%%%%%%% \subsection{UK-DALE Haystack}

\begin{table}[!t]
\centering
\small
\setlength{\tabcolsep}{4pt}
\renewcommand{\arraystretch}{1.05}
\caption{\textbf{UK-DALE Haystack: per-task accuracy across context lengths.} All ten tasks are instantiated at all five context lengths (15\,min, 1\,h, 2\,h, 9\,h, 24\,h) on UK-DALE, the longest context windows in the benchmark, supported by UK-DALE's low sampling rate (see Appendix~\ref{app:ukdale-haystack-stats}). ChatTime exhausts GPU memory at 9\,h and 24\,h (OOM).}
\label{tab:ukdale_per_task}
\begin{tabular}{@{}ll ccccc cc@{}}
\toprule
& & \multicolumn{5}{c}{\textbf{Models}} & \multicolumn{2}{c}{\textbf{Reference}} \\
\cmidrule(lr){3-7} \cmidrule(l){8-9}
\textbf{Task} & \textbf{Length} & ChatTime & ChatTS & ITFormer & Flamingo & ARTS & \emph{Random} & \emph{Real Oracle} \\
\midrule
\multirow{5}{*}{Existence}
 & 15\,min & 0.63          & 0.52          & 0.39          & 0.51          & \textbf{0.81} & 0.50 & 1.00 \\
 & 1\,h    & 0.66          & 0.47          & 0.51          & 0.60          & \textbf{0.80} & 0.50 & 1.00 \\
 & 2\,h    & 0.75          & 0.59          & 0.48          & 0.63          & \textbf{0.76} & 0.50 & 1.00 \\
 & 9\,h    & OOM           & 0.47          & 0.51          & 0.55          & \textbf{0.57} & 0.50 & 1.00 \\
 & 24\,h   & OOM           & 0.51          & 0.49          & 0.52          & \textbf{0.80} & 0.50 & 1.00 \\
\midrule
\multirow{5}{*}{Localization}
 & 15\,min & 0.51          & 0.31          & 0.16          & 0.33          & \textbf{0.73} & 0.18 & 0.91 \\
 & 1\,h    & 0.40          & 0.27          & 0.22          & 0.25          & \textbf{0.61} & 0.16 & 0.91 \\
 & 2\,h    & 0.40          & 0.21          & 0.15          & 0.19          & \textbf{0.63} & 0.12 & 0.91 \\
 & 9\,h    & OOM           & 0.11          & 0.09          & 0.13          & \textbf{0.43} & 0.07 & 0.93 \\
 & 24\,h   & OOM           & 0.07          & 0.03          & 0.07          & \textbf{0.44} & 0.03 & 0.92 \\
\midrule
\multirow{5}{*}{Counting}
 & 15\,min & 0.43          & 0.37          & 0.14          & 0.37          & \textbf{0.76} & 0.17 & 1.00 \\
 & 1\,h    & 0.41          & 0.36          & 0.18          & 0.38          & \textbf{0.69} & 0.17 & 1.00 \\
 & 2\,h    & 0.33          & 0.27          & 0.14          & 0.23          & \textbf{0.67} & 0.17 & 1.00 \\
 & 9\,h    & OOM           & 0.23          & 0.13          & 0.21          & \textbf{0.38} & 0.17 & 1.00 \\
 & 24\,h   & OOM           & 0.25          & 0.15          & 0.13          & \textbf{0.49} & 0.17 & 1.00 \\
\midrule
\multirow{5}{*}{Ordering}
 & 15\,min & 0.51          & 0.51          & 0.54          & 0.51          & \textbf{0.85} & 0.50 & 1.00 \\
 & 1\,h    & 0.55          & 0.50          & 0.53          & 0.51          & \textbf{0.73} & 0.50 & 1.00 \\
 & 2\,h    & 0.57          & 0.52          & 0.49          & 0.50          & \textbf{0.78} & 0.50 & 1.00 \\
 & 9\,h    & OOM           & 0.52          & 0.59          & 0.52          & \textbf{0.81} & 0.50 & 1.00 \\
 & 24\,h   & OOM           & 0.49          & 0.43          & 0.52          & \textbf{0.68} & 0.50 & 1.00 \\
\midrule
\multirow{5}{*}{State Query}
 & 15\,min & 0.66          & 0.69          & 0.65          & \textbf{0.71} & 0.58          & 0.14 & 0.96 \\
 & 1\,h    & \textbf{0.77} & 0.73          & 0.72          & 0.74          & 0.50          & 0.14 & 0.96 \\
 & 2\,h    & \textbf{0.79} & 0.74          & 0.74          & 0.75          & 0.39          & 0.20 & 0.90 \\
 & 9\,h    & OOM           & 0.77          & 0.84          & \textbf{0.85} & 0.27          & 0.25 & 0.69 \\
 & 24\,h   & OOM           & 0.32          & 0.35          & \textbf{0.58} & 0.17          & 0.33 & 0.99 \\
\midrule
\multirow{5}{*}{Antecedent}
 & 15\,min & \textbf{0.47} & 0.13          & 0.11          & 0.12          & 0.15          & 0.25 & 0.98 \\
 & 1\,h    & \textbf{0.61} & 0.22          & 0.15          & 0.36          & 0.33          & 0.17 & 0.97 \\
 & 2\,h    & \textbf{0.48} & 0.25          & 0.17          & 0.33          & 0.19          & 0.14 & 0.97 \\
 & 9\,h    & OOM           & 0.14          & 0.14          & \textbf{0.29} & 0.26          & 0.13 & 0.97 \\
 & 24\,h   & OOM           & \textbf{0.24} & 0.18          & \textbf{0.24} & 0.21          & 0.20 & 0.91 \\
\midrule
\multirow{5}{*}{Comparison}
 & 15\,min & 0.42          & 0.20          & 0.16          & 0.21          & \textbf{0.80} & 0.16 & 0.97 \\
 & 1\,h    & 0.31          & 0.17          & 0.08          & 0.11          & \textbf{0.73} & 0.10 & 0.98 \\
 & 2\,h    & 0.25          & 0.09          & 0.10          & 0.13          & \textbf{0.71} & 0.10 & 0.97 \\
 & 9\,h    & OOM           & 0.05          & 0.02          & 0.05          & \textbf{0.53} & 0.05 & 0.95 \\
 & 24\,h   & OOM           & 0.06          & 0.01          & 0.05          & \textbf{0.53} & 0.03 & 0.98 \\
\midrule
\multirow{5}{*}{Multi-hop}
 & 15\,min & 0.34          & 0.23          & 0.20          & 0.25          & \textbf{0.50} & 0.12 & 0.87 \\
 & 1\,h    & 0.35          & 0.17          & 0.09          & 0.15          & \textbf{0.45} & 0.09 & 0.86 \\
 & 2\,h    & 0.29          & 0.13          & 0.09          & 0.15          & \textbf{0.55} & 0.07 & 0.85 \\
 & 9\,h    & OOM           & 0.17          & 0.11          & 0.15          & \textbf{0.42} & 0.07 & 0.87 \\
 & 24\,h   & OOM           & 0.07          & 0.03          & 0.03          & \textbf{0.41} & 0.03 & 0.89 \\
\midrule
\multirow{5}{*}{Anomaly Det.}
 & 15\,min & \textbf{0.77} & 0.45          & 0.55          & 0.53          & 0.51          & 0.50 & 0.79 \\
 & 1\,h    & \textbf{0.75} & 0.51          & 0.49          & 0.59          & 0.49          & 0.50 & 0.83 \\
 & 2\,h    & \textbf{0.70} & 0.49          & 0.45          & 0.59          & 0.50          & 0.50 & 0.77 \\
 & 9\,h    & OOM           & 0.49          & 0.51          & \textbf{0.54} & 0.53          & 0.50 & 0.75 \\
 & 24\,h   & OOM           & 0.45          & 0.50          & \textbf{0.59} & 0.45          & 0.50 & 0.69 \\
\midrule
\multirow{5}{*}{Anomaly Loc.}
 & 15\,min & \textbf{0.73} & 0.24          & 0.13          & 0.22          & 0.33          & 0.16 & 0.97 \\
 & 1\,h    & \textbf{0.45} & 0.13          & 0.02          & 0.18          & 0.37          & 0.12 & 0.99 \\
 & 2\,h    & \textbf{0.41} & 0.16          & 0.01          & 0.14          & 0.30          & 0.12 & 0.95 \\
 & 9\,h    & OOM           & 0.01          & 0.00          & 0.00          & \textbf{0.19} & 0.02 & 0.95 \\
 & 24\,h   & OOM           & 0.01          & 0.01          & 0.01          & \textbf{0.17} & 0.01 & 0.93 \\
\bottomrule
\end{tabular}
\end{table}

\clearpage

%%%%%%%%%%%%%%%%%%%%%%%%%%%%%%%%%%%%%%%%%%%%%%%%%%%%%%%%%%%%%%%%%%%%%%%%%%%%%%%
%%%%%%%%%%%%%%%%%%%%%% TS-HAYSTACK BENCHMARK %%%%%%%%%%%%%%%%%%%%%%%%%%%%%%%%%%
%%%%%%%%%%%%%%%%%%%%%%%%%%%%%%%%%%%%%%%%%%%%%%%%%%%%%%%%%%%%%%%%%%%%%%%%%%%%%%%

%%

\section{Appendix: TS-Haystack}
\label{app:ts_haystack}
% Present the final TS-Haystack dataset statistics in depth. Present details of contetx lengths and tasks generated per dataset as an overview, (considering LTAF and Sleep don't have all tasks for all lengths -- thinking of existance and anomalies for example) and sleep spans different lengths for anomalies and the rest of tasks. This is a general overview since the details can be written further in depth in the dataset specific sections.

In the following, we present TS-Haystack's construction and report per-dataset statistics. We first give a high-level overview of the benchmark (Appendix~\ref{app:ts_haystack_overview}), then describe shared components of the generation pipeline: the needle-insertion protocol and its statistical validation (Appendix~\ref{app:needle-insertion}), task templates and ground-truth derivation (Appendix~\ref{app:task-construction-details}), evaluation metrics, and random-baseline calculation. Each of the four source datasets (Capture24, Sleep PSG, LTAF, UK-DALE) is then detailed in its own subsection, covering the source corpus, the haystack generation pipeline used to repurpose it, illustrative samples, and the corresponding ARTS classifier tool.

\subsection{TS-Haystack Overview}
\label{app:ts_haystack_overview}

TS-Haystack is built from four openly available, expert-annotated time-series corpora that span four distinct time-series modalities: wrist accelerometry, multi-channel polysomnography, ECG, and low-rate household power.
Each source dataset is repurposed into a subset of TS-Haystack's ten retrieval and reasoning tasks, instantiated at a discrete grid of context lengths. In aggregate, the benchmark is composed more than 60k hours of source recording hours augmented to controlled retrieval samples with contexts spanning multiple orders of magnitude (100~s to 24~h).

Table~\ref{tab:appendix_source_overview} summarizes the source corpora and the context-length grid instantiated per dataset.
Table~\ref{tab:appendix_task_grid} reports which of the ten tasks are instantiated at which context lengths per dataset, together with the total number of generated (train\,/\,validation\,/\,test) samples. Two systematic exclusions are worth noting up front:
\begin{itemize}
\setlength{\itemsep}{2pt}
  \item \emph{Existence} and \emph{Anomaly Detection} are not instantiated at
        the longest context length for some datasets, because the queried target
        class is near-deterministically present in any naturally drawn segment
        at that horizon, collapsing the answer to a trivial ``Yes''.
  \item On Sleep PSG, anomaly tasks (Anomaly Detection / Localization) operate
        on the \emph{arousal} annotation track rather than the sleep-stage
        track, and are therefore instantiated at shorter horizons (down to
        100~s) to account for the higher density and shorter duration of arousal events
        (Appendix~\ref{app:sleep-haystack-stats}).
\end{itemize}

\begin{table}[H]
\centering
\small
\setlength{\tabcolsep}{4pt}
\renewcommand{\arraystretch}{1.1}
\caption{\textbf{TS-Haystack source datasets.} Per-dataset modality, channel
count, sampling rate, number of source recordings, source corpus duration,
total generated benchmark duration (summed over tasks and context lengths),
and the context-length grid at which the source is instantiated in TS-Haystack.
\emph{Construction} indicates whether haystack samples are produced via
semi-synthetic needle insertion or natural-segment sampling
(Section~\ref{sec:construction}).}
\label{tab:appendix_source_overview}
\resizebox{\textwidth}{!}{%
\begin{tabular}{@{}l l c c r r r l l@{}}
\toprule
\textbf{Source dataset} & \textbf{Modality} & \textbf{Ch.} & \textbf{$f_s$ (Hz)}
& \textbf{\#\,Rec.} & \textbf{Source hrs} & \textbf{Benchmark hrs} & \textbf{Context lengths} & \textbf{Construction} \\
\midrule
Capture24~\citep{chan2024capture24}    & Wrist accelerometer & 3  & 100         & 151 & $\sim$3{,}883  & 42{,}611  & 100\,s,\,15\,min,\,1\,h,\,2\,h        & Semi-synthetic \\
Sleep PSG~\citep{ghassemi2018you}      & Polysomnography     & 13 & 100         & 994 & $\sim$6{,}958  & 103{,}963 & 100s, 15\,min,\,1\,h,\,2\,h,\,9\,h          & Natural \\
LTAF~\citep{petrutiu2007ltaf}          & ECG                 & 2  & 128         & 84  & $\sim$1{,}961  & 36{,}108  & 100\,s,\,15\,min,\,1\,h,\,2\,h        & Natural \\
UK-DALE~\citep{kelly2015ukdale}        & Household power     & 1  & $\sim$0.17  & 3   & $\sim$48{,}000 & 471{,}250 & 15\,min,\,1\,h,\,2\,h,\,9\,h,\,24\,h  & Semi-synthetic \\
\midrule
\textbf{Total}                         &                     &    &             & 1{,}232 & $\sim$60{,}802 & 653{,}932 &                                  &                  \\
\bottomrule
\end{tabular}%
}
\end{table}

\begin{table}[H]
\centering
\small
\setlength{\tabcolsep}{3.5pt}
\renewcommand{\arraystretch}{1.1}
\caption{\textbf{TS-Haystack task instantiation matrix.} For each (dataset, context length) cell we report which of the ten tasks are instantiated (\cmark) or excluded (---), together with the number of generated (train\,/\,validation\,/\,test) samples per (task, length) cell. Task abbreviations: \textbf{Ex} existence, \textbf{Loc} localization, \textbf{Cnt} counting, \textbf{Ord} ordering, \textbf{SQ} state query, \textbf{Ant} antecedent, \textbf{Cmp} comparison, \textbf{MH} multi-hop, \textbf{AD} anomaly detection, \textbf{AL} anomaly localization. Per-cell sample counts apply to each instantiated task in that row.}
\label{tab:appendix_task_grid}
\resizebox{\textwidth}{!}{%
\begin{tabular}{@{}l l cccc cc cc cc r@{}}
\toprule
& & \multicolumn{4}{c}{\textbf{Direct retrieval}} & \multicolumn{2}{c}{\textbf{Temporal}}
& \multicolumn{2}{c}{\textbf{Multi-step}} & \multicolumn{2}{c}{\textbf{Anomaly}} & \\
\cmidrule(lr){3-6} \cmidrule(lr){7-8} \cmidrule(lr){9-10} \cmidrule(lr){11-12}
\textbf{Dataset} & \textbf{Length}
& Ex & Loc & Cnt & Ord & SQ & Ant & Cmp & MH & AD & AL & \textbf{\#\,samples (tr/va/te)} \\
\midrule
\multirow{4}{*}{Capture24}
 & 100\,s  & \cmark & \cmark & \cmark & \cmark & \cmark & \cmark & \cmark & \cmark & \cmark & \cmark & 1000\,/\,150\,/\,150 \\
 & 15\,min & \cmark & \cmark & \cmark & \cmark & \cmark & \cmark & \cmark & \cmark & \cmark & \cmark & 1000\,/\,150\,/\,150 \\
 & 1\,h    & \cmark & \cmark & \cmark & \cmark & \cmark & \cmark & \cmark & \cmark & \cmark & \cmark & 1000\,/\,150\,/\,150 \\
 & 2\,h    & \cmark & \cmark & \cmark & \cmark & \cmark & \cmark & \cmark & \cmark & \cmark & \cmark & 1000\,/\,150\,/\,150 \\
\midrule
\multirow{5}{*}{Sleep PSG}
 & 100\,s  & ---    & ---    & ---    & ---    & ---    & ---    & ---    & ---    & \cmark & \cmark & 1000\,/\,150\,/\,150 \\
 & 15\,min & \cmark & \cmark & \cmark & \cmark & \cmark & \cmark & \cmark & \cmark & \cmark & \cmark & 1000\,/\,150\,/\,150 \\
 & 1\,h    & \cmark & \cmark & \cmark & \cmark & \cmark & \cmark & \cmark & \cmark & ---    & \cmark & 1000\,/\,150\,/\,150 \\
 & 2\,h    & \cmark & \cmark & \cmark & \cmark & \cmark & \cmark & \cmark & \cmark & ---    & ---    & 1000\,/\,150\,/\,150 \\
 & 9\,h    & ---    & \cmark & \cmark & \cmark & \cmark & \cmark & \cmark & \cmark & ---    & \cmark & 1000\,/\,150\,/\,150 \\
\midrule
\multirow{4}{*}{LTAF}
 & 100\,s  & \cmark & \cmark & \cmark & \cmark & \cmark & \cmark & \cmark & \cmark & \cmark & \cmark & 1000\,/\,150\,/\,$\sim$150$^{\dagger}$ \\
 & 15\,min & \cmark & \cmark & \cmark & \cmark & \cmark & \cmark & \cmark & \cmark & \cmark & \cmark & 1000\,/\,150\,/\,150 \\
 & 1\,h    & \cmark & \cmark & \cmark & \cmark & \cmark & \cmark & \cmark & \cmark & ---    & \cmark & 1000\,/\,150\,/\,150 \\
 & 2\,h    & ---    & \cmark & \cmark & \cmark & \cmark & \cmark & \cmark & \cmark & ---    & \cmark & 1000\,/\,150\,/\,150 \\
\midrule
\multirow{5}{*}{UK-DALE}
 & 15\,min & \cmark & \cmark & \cmark & \cmark & \cmark & \cmark & \cmark & \cmark & \cmark & \cmark & 1000\,/\,150\,/\,150 \\
 & 1\,h    & \cmark & \cmark & \cmark & \cmark & \cmark & \cmark & \cmark & \cmark & \cmark & \cmark & 1000\,/\,150\,/\,150 \\
 & 2\,h    & \cmark & \cmark & \cmark & \cmark & \cmark & \cmark & \cmark & \cmark & \cmark & \cmark & 1000\,/\,150\,/\,150 \\
 & 9\,h    & \cmark & \cmark & \cmark & \cmark & \cmark & \cmark & \cmark & \cmark & \cmark & \cmark & 1000\,/\,150\,/\,150 \\
 & 24\,h   & \cmark & \cmark & \cmark & \cmark & \cmark & \cmark & \cmark & \cmark & \cmark & \cmark & 1000\,/\,150\,/\,150 \\
\bottomrule
\end{tabular}%
}\\[3pt]
{\footnotesize
$^{\dagger}$ Per-task test counts at LTAF 100\,s vary slightly below the 150 target for five tasks: Localization (111), Multi-hop (115), Comparison (125), Antecedent (133), Ordering (147); the remaining five tasks reach 150.}
\end{table}

\subsection{Needle Insertion}
\label{app:needle-insertion}

Needle insertion is the construction protocol that lets TS-Haystack turn a single annotated corpus into a controlled, semi-synthetic, grid of (context length, target class, needle count) cells, by embedding segments of a target signal (\emph{needles}) into longer recordings of the same domain (\emph{haystacks}). The protocol decouples the \emph{what} happens (the needle) to then \emph{when} and \emph{within how much surrounding context} it happens (the haystack). This decoupling is what enables arbitrary context-length scaling, explicit difficulty control, and combinatorial sample diversity (Table~\ref{tab:construction_choice}). Two of the four source datasets in TS-Haystack admit this protocol (Capture24, UK-DALE); the other two (Sleep PSG, LTAF) do not, and instead rely on natural-segment sampling. Whether a domain admits the protocol turns on a single property of the source signal that we call \emph{local context-independence}.

\paragraph{Operationalization of context-independence.}
Let $\mathbf{x}_{[s:e]}$ denote a bout of class $c$ within a recording$\mathbf{X}$, and let $\mathbf{X}_{\setminus[s:e]}$ denote the rest of the recording. We say the source signal is \emph{locally context-independent} for $c$ if the bout's distribution conditional on its surrounding context is well-approximated by its marginal distribution given the class label,
\begin{equation}
    p\!\bigl(\mathbf{x}_{[s:e]} \mid \mathbf{X}_{\setminus[s:e]},\, c\bigr)
    \;\approx\;
    p\!\bigl(\mathbf{x}_{[s:e]} \mid c\bigr).
    \label{eq:context-independence}
\end{equation}
When~\eqref{eq:context-independence} holds, transplanting a bout into a different background of the same domain preserves its semantics and yields a composite whose joint distribution is statistically close to that of naturally observed class-$c$ bouts embedded in arbitrary contexts. Activity bouts on the wrist and appliance's additive signature on the household mains seem to satisfy this condition to a usable degree of approximation, with the residual gap quantifiable by a domain-specific validation check. This assumption may not hold for all time-series domains. In the case of sleep stages, these signals are embedded in the cyclic NREM/REM macrostructure scored by AASM rules~\citep{berry2017aasm}. In the case of ECG recordings, the interpretation of an arrhythmia depends on surrounding inter-beat morphology and clinical guidelines requiring multiple seconds of contiguous rhythm~\citep{hindricks2020esc} surrounding the labeled section. Thus, insertion would produce an out-of-distribution composite that no boundary-smoothing operation can repair, and we fall back to natural-segment sampling. Further details for needle insertion to validate this hypothesis can be found in the dataset specific sections below.

%%% EXPERIMENTAL SETUP
\section{Experimental setup}
\label{app:exp_setup}

\subsection{TSLM training details}
\label{app:tslm_training}

We aim to train all four single-pass TSLMs (Flamingo, ITFormer, ChatTS, and ChatTime) with a single, dataset-agnostic recipe so that any cross-dataset or cross-architecture difference reflects representational capacity rather than tuning. Our experiments are conducted on a single NVIDIA H100 GPU, with automatic retry on out-of-memory with the deterministic fallback ladder described below. Per-architecture hyperparameters are summarized in Table~\ref{tab:tslm_hparams}.

\paragraph{Optimizer, schedule, and trainable surface.}
AdamW~\citep{loshchilov2018decoupled} with $\beta_1\!=\!0.9$, $\beta_2\!=\!0.999$, weight decay $0.01$, learning rate $2\!\times\!10^{-4}$ for every trainable parameter group, and gradient clipping at L2-norm $1.0$. The schedule is a linear warmup over the first $3\%$ of total steps followed by linear decay to zero (Hugging Face \texttt{get\_linear\_schedule\_with\_warmup}). We train for up to 5 epochs with early stopping on validation loss (patience $=2$) and retain the best-validation checkpoint for test set evaluation. Evaluation generates up to 500 new tokens per sample and is run on $10\%$ of each validation/test split (\texttt{eval\_samples\_ratio: 0.1}) to keep generation cost bounded; the train split is unsubsampled.
The Llama-3.2-1B-Instruct~\citep{grattafiori2024llama3} backbone is frozen in
every configuration except ChatTime, as per the original implementation.

\paragraph{Per-architecture variants.}
\begin{itemize}
\setlength{\itemsep}{2pt}
\item \textbf{Flamingo} \citep{alayrac2022flamingo}: a Perceiver-Resampler
projector with 64 latents feeds gated cross-attention layers inserted into Llama
at every layer ($\mathrm{cross\_attn\_every\_n\_layers}\!=\!1$). Trainable: the
resampler, the cross-attention adapters, and the LM input embeddings.
\item \textbf{ITFormer} \citep{wang2025itformer}: a 2-layer instruction-aware
time-series transformer ($d_{\text{model}}\!=\!2048$, 16 heads, dropout $0.1$)
followed by three small projections (\texttt{ts\_project}, \texttt{query\_project},
\texttt{fusion\_project}) and a $25$-token learnable prefix prepended to the LM.
\item \textbf{ChatTS} \citep{Xie_2025}: a 3-layer MLP patch encoder (patch size
$=20$, hidden $=2048$, ``\texttt{sp}'' positional encoding) is co-trained with
the LM input embeddings; the rest of the LM is frozen. Patch size 20 caps the
effective sequence at 1024 patches per sample.
\item \textbf{ChatTime} \citep{wang2024chattimeunifiedmultimodaltime}: a pure-LLM
TSLM that discretises each time-series value into one of $10\,002$ bin tokens
appended to the LM tokenizer. Because the bin vocabulary is added to the
tokenizer, ChatTime requires a structurally different training mode: the new
token embeddings must be learnable, and the LM itself receives LoRA adapters
($r\!=\!16$, $\alpha\!=\!16$, dropout $=0$, applied to all attention and MLP
projections).
\end{itemize}

\paragraph{Time-series encoders.}
For accelerometer (Capture24), ECG (LTAF), polysomnography (Sleep), and household
power (UK-DALE), the default frozen encoder for Flamingo and ITFormer is
pretrained Chronos-2~\citep{ansari2025chronos2} in multivariate mode for
multi-channel inputs and univariate mode for single-channel inputs.
ChatTS retains its original MLP patch encoder.

\paragraph{Base prompt template.}
Each sample uses a shared three-part skeleton, populated by dataset-specific
question/answer-format fields. 
The \emph{pre-prompt} names the modality, sampling rate, recording duration,
and absolute time range so that all timestamps in the question and answer are
interpretable relative to a known clock. The \emph{question} is drawn from the
QA shard. The \emph{post-prompt} embeds an answer-type-specific format hint
(boolean / integer / category / time\_range) and instructs the model to terminate
its response with the literal string \texttt{"Answer: <your answer>"} so the
evaluator can extract the prediction by regex. The time-series payload is
inserted between pre- and post-prompt as one
\texttt{TextTimeSeriesPrompt(<channel description>, <numerical signal>)}
per channel (e.g.\ ``Triaxial accelerometer (g), x/y/z'' for Capture24).

\paragraph{Prompt example.}
To make the skeleton concrete, consider a UK-DALE-Haystack \texttt{existence}
task at the 1-hour context length. The assembled prompt is:

\begin{quote}\small\ttfamily
You are an expert energy analyst. You are given a UK domestic mains
active-power trace (watts), sampled on a regular 6-second grid. Your task
is to answer questions about which appliances were used and when, based on
a pre-computed appliance-bout analysis. Possible appliances: dishwasher,
fridge, fridge\_freezer, hair\_dryer, kettle, microwave, oven, toaster,
washer\_dryer, washing\_machine.
You are given a long-context UK domestic mains active-power trace
(watts, sampled every 6~s). \\
This window spans 1h00m00s of the recording, from 00:00:00 to 01:00:00.
Time references in the question and answer are window-relative offsets
from 00:00:00, not wall-clock times.\\[2pt]
Question: Was the hairdryer used at all?\\[4pt]
\textlangle TextTimeSeries: ``mains active power (W)'',
\,$\mathbf{x}\!\in\!\mathbb{R}^{600}$\textrangle\\[4pt]
Instructions: \\
- Analyze the mains active-power trace carefully. \\
- Think step-by-step about what the appliance signatures indicate. \\
- Answer with `Yes' or `No'. \\
- End your response with ``Answer: \textlangle your answer\textrangle''
\end{quote}

\paragraph{Context-length packing and OOM fallback.}
The benchmark spans context lengths varying by two to three orders of magnitude
per dataset. For each dataset, training launches with the full range of context lengths and \texttt{batch\_size}\,$=2$ (Flamingo, ITFormer, ChatTS) or $1$ (ChatTime). Each successful run therefore reports results on a single, fixed context mix, the largest the model could fit on the available hardware.
On a CUDA OOM event, we wipe the run's checkpoint directory and retry in this order:

\begin{enumerate}
\setlength{\itemsep}{2pt}
\item if $\mathrm{batch\_size} > 1$, drop to $\mathrm{batch\_size}\!=\!1$ and
retrain from scratch on the same context lengths.
\item otherwise, drop the largest remaining context-length bin and retrain from
scratch with the original batch size;
\item repeat until training succeeds or the ladder is empty (run marked
\textsc{failed}).
\end{enumerate}

\paragraph{Reproducibility.}
Every config sets \texttt{seed: 42} and dataset constructors derive deterministic
per-(task, context, split, sample-index) seeds, so the train/val/test parquet
shards are bit-identical across re-runs; the only run-to-run variance is the
training stochasticity of the trainable bridging modules. Deterministic RNG is
set in PyTorch, NumPy, and CUDA.

\begin{table}[h]
\centering
\small
\setlength{\tabcolsep}{4pt}
\renewcommand{\arraystretch}{1.05}
\caption{Per-architecture training hyperparameters. All runs share
$\mathrm{lr}\!=\!2\!\times\!10^{-4}$, AdamW $\mathrm{wd}\!=\!0.01$,
$\|g\|_2\!\le\!1$, linear warmup ratio $0.03$, max 5 epochs with early
stopping (patience 2) on validation loss, seed 42. Backbone is
Llama-3.2-1B-Instruct.}
\label{tab:tslm_hparams}
\begin{tabular}{@{}lllll@{}}
\toprule
                      & Flamingo                     & ITFormer                     & ChatTS                  & ChatTime                    \\
\midrule
TS encoder            & Chronos-2 (frozen) & Chronos-2 (frozen) & MLP patch (trained)     & --- (text)                  \\
Bridging module       & Perceiver-Resampler          & 25-token prefix              & encoder $\to$ LM        & LoRA on LM                  \\
LM backbone           & frozen                       & frozen                       & frozen                  & LoRA ($r{=}16,\alpha{=}16$) \\
LM input embeddings   & trained                      & trained                      & trained                 & trained                     \\
Default batch size    & 2                            & 2                            & 2                       & 1                           \\
\bottomrule
\end{tabular}
\end{table}

\subsection{ARTS details}
\label{app:arts_details}
% Explain here more in depth how the ARTS passes work, so first we do a pre-pass on the timeseries with the classifier sequentially and get a text prompt with some of the signals as a coarse representation then we let it rip through the agentic stuff.
% Present the maxes of tool calls per context length and sections.
% Present also an ablation of using and not using the pre-pass -- cost vs performance, number of tool calls in each thing.

\paragraph{Pre-pass rendering.}
The classifier pre-pass sweeps $g_\phi$ over a non-overlapping tiling of the recording at a fixed window size and returns per-tile predictions. Before reaching the orchestrator, this output is post-processed and rendered into text. Adjacent tiles sharing a label are compacted into bouts. The bout list is then rendered as a Markdown table (one line per bout, with per-bout peak-amplitude or mean-confidence aggregates depending on the domain) and inserted between the standard pre-prompt (modality, sampling rate, duration, time range) and the question. The orchestrator never receives a raw waveform or sample-level vector, only this rendered timeline plus the question. The total textual size of the pre-pass scales with the number of detected events rather than the recording length, which is what
makes the agentic loop tractable on multi-hour windows.

\paragraph{Prompt example.}
We showcase the prompt example at the same 1-hour context length used for the TSLM prompt example above. The full prompt assembled for the orchestrator is:

\begin{quote}\small\ttfamily
\textbf{System:} \\
You are an expert energy analyst. You are given a UK domestic mains
active-power trace (watts), sampled on a regular 6-second grid. Your task
is to answer questions about which appliances were used and when, based on
a pre-computed appliance-bout analysis. Possible appliances: dishwasher,
fridge, fridge\_freezer, hair\_dryer, kettle, microwave, oven, toaster,
washer\_dryer, washing\_machine.\\[4pt]

\textbf{User (pre):} \\
You are given a UK domestic mains active-power trace from house 1.\\
This window spans 1h00m00s of the recording, from 00:00:00 to 01:00:00.\\[6pt]
\# Pre-pass summary (window 00:00:00 - 01:00:00) \\
\#\# Appliance bouts (5 bouts) \\
\quad per-appliance counts: fridge\_freezer=2, hair\_dryer=1, kettle=1, microwave=1 \\
\quad 00:03:12 - 00:03:34 \  kettle          \  (22\,s) \  peak=2987\,W \  conf=0.98 \\
\quad 00:07:48 - 00:21:06 \  fridge\_freezer \  (13\,m\,18\,s) \  peak=121\,W  \  conf=0.94 \\
\quad 00:14:30 - 00:15:12 \  microwave       \  (42\,s) \  peak=1342\,W \  conf=0.89 \\
\quad 00:28:06 - 00:28:48 \  hair\_dryer     \  (42\,s) \  peak=1155\,W \  conf=0.91 \\
\quad 00:41:55 - 00:53:24 \  fridge\_freezer \  (11\,m\,29\,s) \  peak=118\,W  \  conf=0.96
Question: Was the hairdryer used at all?\\[4pt]

\textbf{User (post):} \\
Instructions: \\
- A pre-computed classifier analysis of the full recording is provided
  above. Consult it first. \\
- You may also call \texttt{classify\_appliance\_window} and
  \texttt{find\_appliance\_bouts} on any sub-range or appliance for a
  finer look; these tools return the same pre-computed predictions
  filtered to your query. \\
- Classifier predictions may be wrong --- cross-check against the
  question and prefer evidence from multiple bouts when possible. \\
- Recording duration: 3600\,s. Tool query ranges may span the full
  recording; results are per-appliance bouts, not raw samples. \\
- Answer with `Yes' or `No'. \\
- End your response with ``Answer: \textlangle your answer\textrangle''
\end{quote}

%%%%%%%%%%%% EVALUATION -- HOW DO WE EVALUATE AND RANDOM BASELINES %%%%%%%%%%%%

\subsection{Random baselines}
\label{app:random-baselines}

The random baseline reported in every results table is computed \emph{per task and context length} and is tied to the scoring rule this ``cell'' uses. For a discrete task we draw a class uniformly from the answer space the cell actually contains; for a time range task it places an interval of the ground-truth width uniformly inside the window. A model whose accuracy is at or below this value has not learned to use the signal in that cell. We compute the baseline on the held-out test split's normalised ground truth, so each cell's number reflects the answer space the cell actually contains rather than a global average.

\paragraph{Discrete tasks (boolean, integer, category).}
The scorer marks a sample correct if $\mathrm{normalize}(\hat{y}) = \mathrm{normalize}(y)$. Let $K$ be the number of distinct normalised ground-truth classes observed in the cell.
Then
\begin{equation}
    \text{Random}_{\text{cell}} \;=\; \frac{1}{K}.
\end{equation}
The number of observed classes can be smaller than the dataset's nominal vocabulary. Counting cells use $K = 5$ corresponding to the integers 1--5 emitted by the generator's \texttt{min\_bouts} -- \texttt{max\_bouts} parameters; ordering and antecedent cells use the per-cell observed class set; and boolean cells (existence, anomaly\_detection) use $K = 2$.

\paragraph{Time range tasks (localization, comparison, multi-hop, anomaly localization).}
The scorer marks a sample correct if $\mathrm{IoU}(\hat{y}, y) \ge \tau_{\text{IoU}}$, with
$\tau_{\text{IoU}} = 0.25$. For \texttt{anomaly\_localization} the scorer additionally has a ``negative-answer'' shortcut: if both prediction and ground truth begin with the literal string ``no'', the sample is marked correct without parsing intervals. The random baseline is derived from these two rules.

\textit{Geometric chance for the IoU rule.} For two intervals of equal
width $w$ centred at $c_1, c_2$ on a window of length $W$,
\begin{equation}
    \mathrm{IoU}
    = \frac{w - |c_1 - c_2|}{w + |c_1 - c_2|}
    \;\ge\; 0.25
    \quad\Longleftrightarrow\quad
    |c_1 - c_2| \;\le\; 0.6\, w.
\end{equation}
If both centers are independently uniform on $[0, W]$, the probability of meeting this bound is $\min(1,\ 1.2\, w / W)$. We apply this per-sample with $w$ set to the ground-truth interval width. Equivalently, this is the rate attained by a predictor that uses the correct interval width but places its center uniformly at random. The cell's random baseline is the mean over its samples,
\begin{equation}
    \text{Random}_{\text{cell}}
    \;=\; \frac{1}{N_{\text{cell}}}
        \sum_{j \in \text{cell}}
        \min\!\Bigl(1,\ \frac{1.2\, w_j}{W}\Bigr)
        \cdot \mathbf{1}\!\bigl[y_j \text{ is a range, not ``no''}\bigr].
\end{equation}

\paragraph{Behaviour vs.\ context length.}
The cell-by-cell construction makes the random baseline's dependence on
context length transparent and partially explains the trends visible in the
results tables.
\begin{itemize}
    \setlength{\itemsep}{2pt}
    \item \textbf{Discrete tasks.} The answer space does not depend on
    $W$, so the baseline is a function of the empirical class
    distribution only. State-conditioned tasks (\texttt{state\_query},
    \texttt{antecedent}) can shift modestly when longer windows expose a
    different sub-set of the vocabulary.
    \item \textbf{Fixed-needle time range tasks.} The
    ground-truth width $w$ is approximately context-independent, so the baseline scales
    as $\sim w/W$ and decays roughly linearly with $W$. While the random baseline decays linearly with context length, the higher rate of TSLM performance decay with context length motivates our result's discussion.
\end{itemize}

%%%%%%%%%%%%%%%%%%%%%% CAPTURE24 %%%%%%%%%%%%%%%%%%%%%%%%%%%%%%%%%%

\subsection{Capture24-Haystack}
\label{app:capture24-haystack}

\paragraph{Capture24: base dataset.}
\label{app:capture24-base}
We use Capture24~\citep{chan2024capture24} in its public release: 151
participants each wearing an Axivity AX3 wrist accelerometer at $100$\,Hz
for approximately 24 hours of free-living recording, totalling
${\sim}3{,}883$ source-hours. Each sample is a 3-channel acceleration
vector in units of $g$. Bouts are annotated against the
\citet{willetts2018statistical} ten-class scheme: \emph{sleep},
\emph{sitting}, \emph{standing}, \emph{walking}, \emph{bicycling},
\emph{vehicle}, \emph{household-chores}, \emph{manual-work}, \emph{sports},
and \emph{mixed-activity}. The class distribution is heavily skewed:
\emph{sleep} and \emph{sitting} together account for over $70\%$ of
naturally occurring time, while \emph{bicycling}, \emph{sports}, and
\emph{manual-work} each contribute under $2\%$. We retain the original
participant-level partition into train / validation / test splits and
draw all needles and backgrounds for each TS-Haystack split from the
corresponding Capture24 split, so that no participant contributes to
both a haystack background and any needle in a different split.

\paragraph{Capture24-Haystack: benchmark statistics.}
\label{app:capture24-haystack-stats}
Capture24-Haystack instantiates all ten TS-Haystack tasks at four context
lengths: $100$\,s, $15$\,min, $1$\,h, $2$\,h, with a fixed
$1{,}000\,/\,150\,/\,150$ train\,/\,validation\,/\,test sample budget per
(task, length) cell (Table~\ref{tab:appendix_task_grid}). The construction
draws on $151$ participants and spans $42{,}611$ aggregate sample-hours
across the benchmark. Aggregate per-task accuracy is reported in
Table~\ref{tab:capture24_per_task}, with a fuller per-domain discussion
in Appendix~\ref{app:per_domain_discussion}.

\paragraph{Construction pipeline.}
\label{app:capture24-haystack-construction}
We construct each Capture24-Haystack sample as a deterministic function of
the (task, context length, sample index, split) tuple via a
seeded random-number generator, ensuring bit-identical reproducibility
across re-runs. The shared three-step procedure is:

\begin{enumerate}
\setlength{\itemsep}{2pt}
\item \textbf{Background selection.} A target context length $L$ defines
the haystack window. We sample a participant from the appropriate split,
draw a contiguous $L$-second window from a recording in which the
queried target class $c$ is \emph{absent} for the entire window, and
adopt this window as the haystack background. 
\item \textbf{Needle sampling.} For each of $n$ needle slots required by
the task (Appendix~\ref{app:task-construction-details}), we sample an
annotated bout of class $c$ from a different participant. Bouts are
drawn uniformly at random from the pool of valid class-$c$ bouts that
exceed a minimum duration; the sampled bout is then trimmed to a
\emph{fixed} per-(task, length) needle width $w_{c,L}$ shared across all
samples in that cell.
\item \textbf{Insertion.} Each needle is inserted at a controlled
non-overlapping location within $[0, L]$ that determines the
ground-truth answer. This needle replaces the corresponding
background segment after transformations (see Appendix~\ref{app:needle-insertion}.
\end{enumerate}

\paragraph{Fixed needle width.}
Needle widths $w_{c,L}$ are either naturally bounded by their full length (Sleep, ECG and Uk Dale) or fixed in the generator (Capture24) rather than scaled with $L$. Equivalently, the needle-to-background ratio $w_{c,L} / L$ \emph{decreases} with $L$, which mirrors the natural sparsity of relevant events in long recordings.

\subsection{Capture24: Insertion Quality Validation}
\label{app:insertion_validation}

To verify that our style transfer and cosine blending protocol produces clean insertions without detectable artifacts, we design a classifier-based validation test. The key insight is that if a machine learning model cannot distinguish samples with insertions from samples without, the insertion process introduces no systematic artifacts.

\paragraph{Methodology.} We construct a binary classification task:
\begin{itemize}
    \item \textbf{Class 0 (negative):} Pure background windows with no insertion.
    \item \textbf{Class 1 (positive):} Pure background windows with a same-activity needle inserted from a \textit{different participant}.
\end{itemize}
Using same-activity needles isolates insertion artifacts from activity-class differences. If the classifier succeeds, it must be detecting blending imperfections rather than activity signatures.

We generate 5{,}000 training samples and 500 test samples at 2 and 3 seconds context length (200 and 300 samples at 100Hz), with needle lengths ranging from 2--8\% of context aligned with the true dataset configurations and balanced class distributions. An XGBoost classifier~\citep{chen2016xgboost} is trained on flattened triaxial accelerometer features (600 and 900 dimensions) using 100 estimators with max depth 6.

\paragraph{Results.} The classifier achieves an AUC of 0.499 and 0.490 respectively for different context lengths, close to random chance (0.50). This confirms that our mean-shift normalization combined with cosine blending produces insertions that are statistically undetectable, even to a gradient-boosted ensemble operating on raw signal features. The protocol successfully preserves activity dynamics while eliminating participant-specific biases and boundary discontinuities.

\paragraph{Limitations.}
The test as constructed has two known limitations. First, an XGBoost classifier on handcrafted features is a finite-capacity probe; a more expressive classifier might achieve $\mathrm{AUC} > 0.5$ on Capture24
where the present test does not, and the negative result should be read as an upper-bound argument given a reasonable feature space rather than a proof of full distributional equivalence. Second, the test verifies that \emph{inserted} samples match \emph{natural} samples of the same target class, the relevant condition for benchmarking. It does not verify that all marginal statistics of the haystack background remain unchanged
under insertion, because the insertion only modifies a $w$-wide segment of the $L$-wide background and the rest of the window is by construction unmodified.

\subsection{ARTS classifier construction (Capture24)}
\label{app:arts-classifier-capture24}

The ARTS classifier tool $g_\phi$ for Capture24 is a frozen-encoder
with a trainable classification head, finetuned once on the Capture24 train
split with the \citet{willetts2018statistical} ten-class label scheme
(Appendix~\ref{app:capture24-base}) and reused unchanged across all
Capture24-Haystack tasks. We fix the deployed inference window at 10\,s
at 30\,Hz (the ARTS pre-pass tiles the recording at this resolution). To make the head robust to short bouts that would otherwise be discarded at this tile size, training uses a mixed window sizes regime: each batch contains windows of $1$\,s, $3$\,s, $6$\,s, and $10$\,s, all zero-padded to the encoder's $300$-sample ($10$\,s @ $30$\,Hz) input grid, so the encoder also learns representations on short, padded inputs. Final test evaluation uses $10$\,s windows only, matching deployment, yet robust variable length classification remains important for the agentic tool use performance.

We compare three frozen-encoder choices for $g_\phi$ to motivate the
final architecture used in ARTS. \emph{Chronos-2}~\citep{ansari2025chronos2} applied here as a frozen encoder feeding a 2-layer
MLP head. \emph{OxWearables}~\citep{yuan2024oxwearables} is a
domain-specific wrist-accelerometer encoder pretrained on UK Biobank
self-supervised contrastive objectives, applied identically with a
2-layer MLP head on its output. \emph{Dual} concatenates the frozen
Chronos-2 and OxWearables embeddings and feeds the joint
representation to a 2-layer MLP head. All three configurations share
the same head architecture, optimizer (AdamW with weight decay $10^{-4}$),
learning rate ($10^{-3}$), batch size ($128$), class-balanced
cross-entropy loss with weights inversely proportional to training
class frequencies, cosine learning-rate schedule, and 50-epoch budget
with best-validation checkpoint selection on macro-F1. The encoders
remain frozen throughout; only the head is trained.

Table~\ref{tab:arts_capture24_classifier} reports the resulting test
accuracy at the 10\,s deployment window and at 30\,s for reference. The
dual encoder consistently improves classification. We adopt the
dual configuration for all Capture24-Haystack ARTS evaluations reported
in the main paper.

\begin{table}[h]
\centering
\small
\setlength{\tabcolsep}{6pt}
\renewcommand{\arraystretch}{1.05}
\caption{ARTS classifier ablation on Capture24. Three frozen-encoder
choices (Chronos-2, OxWearables, dual concatenation) with an identical
2-layer MLP head, evaluated at the 10\,s deployed inference window and
the 30\,s window for reference. Best per (window, metric) cell in
\textbf{bold}. Macro-F1 is reported alongside plain accuracy because
of Capture24's heavy class imbalance.}
\label{tab:arts_capture24_classifier}
\begin{tabular}{@{}llcc@{}}
\toprule
Window & Encoder      & Macro-F1       & Accuracy       \\
\midrule
\multirow{3}{*}{$10$\,s (deployed)}
       & Chronos-2    & 0.414          & 0.665          \\
       & OxWearables  & 0.481          & 0.690          \\
       & Dual         & \textbf{0.485} & \textbf{0.713} \\
\midrule
\multirow{3}{*}{$30$\,s}
       & Chronos-2    & 0.417          & 0.663          \\
       & OxWearables  & 0.481          & 0.690          \\
       & Dual         & \textbf{0.483} & \textbf{0.712} \\
\bottomrule
\end{tabular}
\end{table}

%%%%%%%%%%%%%%%%%%%%%% Sleep %%%%%%%%%%%%%%%%%%%%%%%%%%%%%%%%%%
\subsection{Sleep-Haystack}
\label{app:sleep-haystack}

\paragraph{Sleep PSG: base dataset.}
\label{app:sleep-base}
We use the PhysioNet/Computing in Cardiology Challenge 2018 ``You Snooze,
You Win'' release~\citep{ghassemi2018you}: $994$ subjects each recorded
with a $13$-channel polysomnogram at $200$\,Hz for one in-lab overnight
session. Recording lengths range from $5.1$\,h to $9.5$\,h (mean
$7.71$\,h, median $7.74$\,h), totalling ${\sim}6{,}958$ source-hours
once non-scoring tail segments are excluded. The signal stack consists
of six EEG derivations (F3-M2, F4-M1, C3-M2, C4-M1, O1-M2, O2-M1), one
EOG channel (E1-M2), one submental EMG (Chin1-Chin2), two inductive
respiratory belts (ABD, CHEST), an oronasal AIRFLOW signal, peripheral
SaO$_2$, and a single ECG lead. We decimate to $100$\,Hz once in the
loader to match the sampling regime of the rest of TS-Haystack.

The dataset ships with two parallel annotation regimes that we treat as
two distinct label families:
\begin{itemize}
\setlength{\itemsep}{2pt}
\item \textbf{Sleep stages}: AASM 5-class scoring~\citep{berry2017aasm}
in $30$\,s epochs: \emph{Wake}, \emph{N1}, \emph{N2}, \emph{N3},
\emph{REM}. Pooled time fractions are
$18.1\,/\,14.6\,/\,43.1\,/\,11.5\,/\,12.7\%$ respectively, with $N2$
dominating and $N3$/$REM$ sparser but architecturally structured.
$90.8\%$ of subjects exhibit all five stages.
\item \textbf{Arousals}: sample-level event annotations for
respiratory and EEG arousals: \emph{rera}, \emph{hypopnea},
\emph{obstructive\_apnea}, \emph{central\_apnea}, \emph{mixed\_apnea}. Median total arousals per subject is $150.5$, with median inter-event gap $25.6$\,s ($92.4\%$ of gaps under $5$\,min).
\end{itemize}

We split subjects deterministically (seed 42) at the participant level
into $695\,/\,149\,/\,150$ train\,/\,validation\,/\,test, shared across
both label families and all tasks. No subject contributes to more than
one split.

\paragraph{Sleep-Haystack: benchmark statistics.}
\label{app:sleep-haystack-stats}
Sleep-Haystack uses sleep-stage annotations as its primary label family
and arousal annotations only where the underlying clinical structure
demands it. All ten TS-Haystack tasks are instantiated on the
sleep-stage timeline at four context lengths: $15$\,min, $1$\,h,
$2$\,h, and $9$\,h (whole-night), with three exceptions that route
through arousals:
\begin{itemize}
\setlength{\itemsep}{2pt}
\item \texttt{anomaly\_detection} and \texttt{anomaly\_localization},
which target arousals because in PSG ``anomaly'' clinically refers to
respiratory and EEG arousals (\emph{rera}, \emph{hypopnea},
\emph{obstructive\_apnea}, \emph{central\_apnea}, \emph{mixed\_apnea})
rather than to a stage. These two tasks operate on shorter horizons where arousal density is sufficient for retrieval without the target class saturating the window.
\item \texttt{state\_query} (``what sleep stage at the time of the
$k$-th \emph{rera}?''), which is the one task that legitimately
crosses the two timelines: an arousal-indexed anchor and a
stage-valued answer. It is instantiated on the sleep-stage grid.
\end{itemize}

\textbf{Why different context-length grids.} The two regimes operate at
qualitatively different time-scales. Arousal events are short
(median $19.1$\,s; class-medians of $18.4\,/\,20.9\,/\,18.0\,/\,15.6$\,s
for the four common classes) and densely interleaved (median inter-event
gap $26$\,s), so a $100$\,s window already contains a useful retrieval
signal while remaining naturally balanced for existence ($75\%$ of
$100$\,s windows on the test split contain zero events of a given
class). Sleep stages are longer, with the canonical classification length being 30~s. For this reason, the shortest stage grid point
is $15$\,min, as shorter windows would not contain enough stage
transitions to populate ordering, antecedent, or multi-hop. We extend
stages to $9$\,h (whole-night) because at this horizon inter-cycle stage
ordering becomes the natural retrieval primitive; we cap arousal existence at the
shorter horizons because beyond those, \emph{rera} and \emph{hypopnea} saturate existence (${\sim}96\%$ of subjects positive).

\paragraph{Sleep-Haystack: construction pipeline.}
\label{app:sleep-haystack-construction}
Unlike Capture24-Haystack we do \emph{not} use needle insertion. Sleep
follows highly structured cycles:
Wake $\rightarrow$ N1 $\rightarrow$ N2 $\rightarrow$ N3 $\rightarrow$
REM, with REM episodes lengthening across the
night~\citep{berry2017aasm}, governed by the AASM scoring rules and the
ultradian organisation of human sleep. Splicing a synthetic REM bout
into sustained N3, or an obstructive apnea into a Wake epoch, would
create physiologically inconsistent segments.
Sleep is therefore a good example of the
\emph{natural-segment sampling} branch of Table~\ref{tab:construction_choice}:
we use unmodified slices of the original recordings and let natural
variation in window content provide the difficulty gradient.

Construction is again a deterministic function of the
(label family, task, context length, sample index, split) tuple via a
seeded RNG. The shared procedure is:

\begin{enumerate}
\setlength{\itemsep}{2pt}
\item \textbf{Window index.} For each (subject, context length $L$) we
build, once, the index of valid window starts: a sliding window of
length $L$ steps across the recording at stride
$\min(L/4, 30\,\text{min})$ for sleep stages and
$\min(L/4, 5\,\text{min})$ for arousals, and a start is admitted if
the window contains at least one bout from the relevant regime.
The $9$\,h length is treated as the whole recording (one window per
subject), which may vary in length.
\item \textbf{Window draw.} Sample generation draws (subject,
window\_start) pairs from the per-split slice of this index. The window
is presented to the model as starting at $00\!:\!00\!:\!00$, so
all timestamps in question and answer are window-relative.
\item \textbf{Window-scoped QA.} The regime timeline is clipped to the
window: bouts straddling a window boundary are truncated to the window
edge before being indexed.
\end{enumerate}

\subsection{ARTS classifier construction (Sleep)}
\label{app:arts-classifier-sleep}

The Sleep ARTS pre-pass exposes \emph{two} parallel classifier tools to
the orchestrator, one per label family. Both share the same backbone: a frozen Chronos-2
multivariate encoder~\citep{ansari2025chronos2} consuming all $13$ PSG
channels at $100$\,Hz, feeding a $2$-layer MLP head
($\text{Linear}(d, 512) \!\to\! \text{ReLU} \!\to\! \text{Dropout}(0.3)
\!\to\! \text{Linear}(512, K)$) with $K$ depending on the label family.
Training uses AdamW (lr $10^{-3}$, weight decay $10^{-4}$), batch size
$64$, class-weighted cross-entropy with weights inversely proportional
to per-split class frequencies, cosine LR schedule, and a $30$-epoch
budget with best-validation-macro-F1 checkpoint selection. The encoder
remains frozen throughout; only the head is trained. Both classifiers
are trained on the same participant split as the benchmark, so a
subject used in any benchmark window is never used to train either
classifier.

The two configurations differ in their per-window contract and in how
training samples are drawn from each subject:

\begin{itemize}
\setlength{\itemsep}{2pt}
\item \textbf{Sleep-stages classifier.} $K = 5$ (\emph{Wake}, \emph{N1},
\emph{N2}, \emph{N3}, \emph{REM}). The deployed inference window is
$30$\,s ($3{,}000$ samples), matching the AASM scoring grain of the
source annotations. Training samples are the natural deterministic
tiling: every contiguous $30$\,s epoch inside an annotated bout
becomes one (subject, start, end, label) example.
\item \textbf{Arousals classifier.} $K = 6$ (\emph{rera},
\emph{hypopnea}, \emph{obstructive\_apnea}, \emph{central\_apnea},
\emph{mixed\_apnea}, plus a synthetic \emph{none} class). The deployed
inference window is $20$\,s ($2{,}000$ samples). Positive examples
are one window per kept event with a random offset that guarantees
overlap of $\min(10\,\text{s}, \text{event length})$ between window
and event. Negative (\emph{none}) examples are drawn from gaps that
lie $\geq\!10$\,s away from any annotation, with one negative per
positive event per subject; offsets and negatives are re-sampled each
epoch via a per-epoch RNG seed so the head sees a fresh slice of the
(continuous) window space across training.
\end{itemize}

Test-set metrics for the deployed checkpoints are reported in
Table~\ref{tab:arts_sleep_classifier}. We note that these classifier performances are not conceptualized to become a state-of-the-art, but rather a simple classifier implementation close to the TSLM baselines for the instantiation of ARTS. Future work could explore better performance on agentic retrieval using existing frontier architectures for sleep-stage and arousal classification.

\begin{table}[h]
\centering
\small
\setlength{\tabcolsep}{6pt}
\renewcommand{\arraystretch}{1.05}
\caption{ARTS classifier metrics on Sleep. Both classifiers use a
frozen Chronos-2 multivariate encoder over the full $13$-channel PSG
input at $100$\,Hz with an identical $2$-layer MLP head and identical
optimizer/schedule.}
\label{tab:arts_sleep_classifier}
\begin{tabular}{@{}lccc@{}}
\toprule
Classifier                & Window  & Macro-F1          & Accuracy          \\
\midrule
sleep stages     & $30$\,s & $0.731 $  & $0.746$  \\
arousals                  & $20$\,s & $0.464$            & $0.626$            \\
\bottomrule
\end{tabular}
\end{table}

\begin{table}[h]
\centering
\small
\setlength{\tabcolsep}{6pt}
\renewcommand{\arraystretch}{1.05}
\caption{Per-class F1 for the deployed arousals classifier on the
held-out test split.}
\label{tab:arts_sleep_arousals_perclass}
\begin{tabular}{@{}lcccccc@{}}
\toprule
Class & rera & hypopnea & obstructive\_apnea & central\_apnea & mixed\_apnea & none \\
\midrule
F1    & 0.39 & 0.47     & 0.41               & 0.57           & 0.12         & 0.82 \\
\bottomrule
\end{tabular}
\end{table}

%%%%%%%%%%%%%%%%%%%%%% LTAF %%%%%%%%%%%%%%%%%%%%%%%%%%%%%%%%%%
\subsection{LTAF-Haystack}
\label{app:ltaf-haystack}

\paragraph{LTAF: base dataset.}
\label{app:ltaf-base}
We use the PhysioNet Long-Term Atrial Fibrillation
database~\citep{petrutiu2007ltaf} in its public release: $84$ two-lead ECG recordings sampled at $128$\,Hz, each approximately
$24$ hours long, totalling ${\sim}1{,}961$ source-hours. Recordings are
annotated against two parallel timelines that we treat as the two label
families of LTAF-Haystack.

\textbf{Rhythm timeline.} Bouts are annotated against the AHA rhythm
vocabulary, which we restrict to the $9$ rhythm codes that occur in the
$84$-record subset: \emph{NSR} (normal sinus rhythm), \emph{AFIB}
(atrial fibrillation), \emph{SBR} (sinus bradycardia), \emph{AB} (atrial
bigeminy), \emph{B} (ventricular bigeminy), \emph{T} (ventricular
trigeminy), \emph{SVTA} (supraventricular tachyarrhythmia), \emph{VT}
(ventricular tachycardia), and \emph{IVR} (idioventricular rhythm). The bout
distribution is heavily skewed: \emph{NSR} and \emph{SBR} together
account for the majority of annotated time, while \emph{T}, \emph{VT},
and \emph{IVR} are rare (Table~\ref{tab:ltaf_bout_counts}). This skew is
preserved during sample construction, no resampling or class
re-weighting is applied at generation time.

We partition the $84$ records into train / validation / test at the
\emph{record} level with a deterministic seeded shuffle (seed $42$,
ratios $0.8\,/\,0.1\,/\,0.1$), yielding $67\,/\,8\,/\,9$ recordings.
All needles, backgrounds, and questions for a given LTAF-Haystack split
are drawn exclusively from the corresponding LTAF split, so no recording
contributes to more than one split.

\begin{table}[h]
\centering
\small
\setlength{\tabcolsep}{8pt}
\renewcommand{\arraystretch}{1.05}
\caption{Annotated rhythm bout counts in the $84$-record LTAF subset
(across all splits, after the LTAF timeline builder). Rare rhythms
(\emph{T}, \emph{VT}, \emph{IVR}) constrain how many samples can be
drawn at the shorter context lengths.}
\label{tab:ltaf_bout_counts}
\begin{tabular}{@{}lr@{}}
\toprule
Rhythm & Bouts \\
\midrule
NSR  & 22{,}879 \\
SBR  & 11{,}326 \\
AFIB &  7{,}358 \\
AB   &  4{,}472 \\
SVTA &  3{,}268 \\
B    &  2{,}696 \\
VT   &      828 \\
T    &      785 \\
IVR  &      137 \\
\bottomrule
\end{tabular}
\end{table}

\paragraph{LTAF-Haystack: benchmark statistics.}
\label{app:ltaf-haystack-stats}
LTAF-Haystack instantiates ten TS-Haystack tasks at four context lengths
--- $100$\,s, $15$\,min, $1$\,h, and $2$\,h, with a fixed
$1{,}000\,/\,150\,/\,150$ train\,/\,validation\,/\,test sample budget
per (task, length) cell. Three cells are gated off because the target
class saturates: \emph{existence} at $2$\,h and \emph{anomaly\_detection} at $1$\,h and $2$\,h, as most LTAF windows contain most of the class labels at these lengths and the yes/no question becomes near-uniformly
``yes''). Five $100$\,s cells under-fill the $1{,}300$ target by a small margin ($1{,}261$ to $1{,}297$ samples) because their joint annotation constraints (e.g., a localizable rhythm transition inside a $100$\,s window) admit fewer feasible windows; thin shards are accepted rather
than topped up by synthesis.

\paragraph{LTAF-Haystack: construction pipeline.}
\label{app:ltaf-haystack-construction}
We use natural segment sampling for LTAF-Haystack. The decision not to use
needle insertion is motivated by the idea that ECG carries strong inter-beat
dependencies that no boundary-smoothing operation can repair. Splicing
an AFIB bout into a sinus background would produce a distributional
artifact that is not closable by mean alignment or cosine tapering at
the seam. Every LTAF-Haystack
sample is consequently an \emph{unmodified} two-lead window taken
directly from one source recording. The labels driving each task come from
the rhythm and beat timelines of the same recording, both of which are
intrinsic annotations of that recording's signal.

Sample construction proceeds in three steps and is a deterministic
function of the (task, context length, sample index, split) tuple via a
seeded random-number generator:

\begin{enumerate}
\setlength{\itemsep}{2pt}
\item \textbf{Window indexing.} For each record and each context length
$L$, we enumerate candidate window starts on a stride of
$\max(1\,\text{s},\, \min(L/4,\, 30\,\text{min}))$. For every candidate,
we precompute (i) a $9$-bit rhythm-presence mask (one bit per rhythm in
the canonical order) by intersecting the window with the rhythm
timeline, and (ii) a $4$-tuple of $(N, A, V, Q)$ beat counts from the
beat timeline. The result is persisted per (label family, $L$) as a JSON
window index and shared across all tasks at that context length.
\item \textbf{Window draw.} For each (task, context length, split)
triple, the task generator queries the window index over the split's
records using whatever predicate it requires (e.g., ``contains
$\geq 1$ \emph{V} beat'' for \texttt{anomaly\_detection}, or ``rhythm
$r$ present / absent'' for the positive / negative draws of
\texttt{existence}). The task draws its sample budget by sampling
uniformly from the candidate (record, window\_start) pairs that satisfy
its predicate.
\item \textbf{Window-scoped QA.} The task generator clips both the
rhythm timeline and the beat timeline to the sampled window, shifts
timestamps to be window-relative, and instantiates question and answer
from the clipped annotations using the LTAF prompt template bank. The signal is
materialized lazily from the per-record \texttt{.npy} cache via
memory-mapped slicing for efficient training and inference.
\end{enumerate}

\subsection{ARTS classifier construction (LTAF)}
\label{app:arts-classifier-ltaf}

The ARTS pre-pass on LTAF resembles the Sleep-Haystack setup in that we
expose \emph{two} specialized classifiers. A rhythm classifier produces the symbolic timeline used for
rhythm-level retrieval, and a beat classifier produces a per-beat label
sequence used for beat-level retrieval and anomaly tasks. Both
classifiers are trained from scratch on the LTAF train$+$validation
records with the same record-level split as the benchmark and reused
unchanged across all LTAF-Haystack tasks.

\paragraph{Rhythm classifier ($g_\phi^{\text{rhythm}}$).}
A wide $1$D-ResNet (\texttt{RhythmResNet1D}: stem plus four stages with
two basic residual blocks each, \texttt{base\_channels}\,$=64$,
$8.79$\,M parameters) takes $z$-scored two-lead $10$\,s windows at
$128$\,Hz and emits a softmax over the $6$-class subset
$\{$NSR, AFIB, SBR, AB, SVTA, B$\}$. At pre-pass time the classifier
is tiled across the QA window in non-overlapping $10$\,s chunks and
consecutive same-label predictions are merged into the derived
rhythm timeline. Test-time augmentation with $7$ random window-start
offsets and softmax averaging ($+4.2$ macro-F$1$ points over
single-view) is the deployed inference setting.

\paragraph{Beat classifier ($g_\phi^{\text{beat}}$).}
A History-Time-Frequency (HTF) ensemble (\texttt{EcgBeatHTFClassifier}:
$1.14$\,M parameters) operates on $2$\,s two-lead windows centred on
each annotated R-peak. Three parallel streams capture complementary
information: a $1$D-CNN over the raw two-lead time window (QRS
morphology), a $1$D-CNN over the log-magnitude rFFT of the same window
(spectral signature of \emph{V} vs.\ \emph{A} vs.\ \emph{N}), and an
MLP over the $K{=}5$-step history of preceding RR intervals and
preceding predicted labels (rhythm context such as bigeminy / trigeminy
coupling, teacher-forced at training, autoregressive at inference).
Stream features are concatenated before a $3$-class MLP head over
$\{$N, A, V$\}$. The unclassifiable \emph{Q} class is dropped from the
head because its support is negligible in this subset
(${\sim}0.06\%$ peak per-record ratio; see
Appendix~\ref{app:ltaf-base}).

\paragraph{Test results.}
The two classifiers are evaluated on the $9$ held-out test records
(no overlap with training records). Headline figures are summarised in
Table~\ref{tab:arts_ltaf_classifiers}. We report the rhythm classifier
at both single-view and TTA-$7$ inference, and the beat classifier at
its native $2$\,s window.

\begin{table}[h]
\centering
\small
\setlength{\tabcolsep}{6pt}
\renewcommand{\arraystretch}{1.05}
\caption{ARTS pre-pass classifiers on LTAF-Haystack. Both models are
trained from scratch on the $67$ train and $8$ validation records and
evaluated on the $9$ held-out test records. Macro-F$1$ is reported
alongside accuracy because of LTAF's class imbalance.}
\label{tab:arts_ltaf_classifiers}
\begin{tabular}{@{}llcc@{}}
\toprule
Classifier & Setting & Accuracy & Macro-F$1$ \\
\midrule
\multirow{2}{*}{\shortstack[l]{Rhythm \\ ($6$-class, $10$\,s)}}
 & single-view  & 0.636 & 0.614 \\
 & TTA-$7$      & 0.684 & \textbf{0.656} \\
\midrule
Beat ($3$-class, $2$\,s)
 & single-view  & 0.954 & \textbf{0.936} \\
\bottomrule
\end{tabular}
\end{table}

%%%%%%%%%%%%%%%%%%%%%% UK-DALE %%%%%%%%%%%%%%%%%%%%%%%%%%%%%%%%%%
\subsection{UK-DALE-Haystack}
\label{app:ukdale-haystack}

\paragraph{UK-DALE: base dataset.}
\label{app:ukdale-base}
We use UK-DALE~\citep{kelly2015ukdale}: appliance-level
sub-meter and whole-house mains active-power demand for five UK domestic
homes at a $6$\,s nominal sampling period. The mains channel for each
kept house is its \texttt{meter1} active-power site meter; sub-meter
channels are the per-appliance individual meters. We retain houses
\textbf{1, 2, and 5} and drop houses 3 and 4: house 3 carries only four
appliance-bearing meters over a 6-week recording, and house 4's metering
is circuit-bundled (e.g.\ \texttt{tv\_dvd\_digibox\_lamp},
\texttt{washing\_machine\_microwave\_breadmaker}) so per-appliance
ground truth is unavailable. The kept recording windows are
$2012\text{-W}45 \to 2017\text{-W}17$ (house 1, $233$ ISO-weeks),
$2013\text{-W}08 \to 2013\text{-W}41$ (house 2, $34$ weeks), and
$2014\text{-W}27 \to 2014\text{-W}46$ (house 5, $20$ weeks), totalling
$287$ source-weeks ($\sim$$48{,}000$ source-hours). The base dataset for
UK-DALE-Haystack is split at the (house, ISO-week) level into
$80\,/\,10\,/\,10$ train\,/\,validation\,/\,test with seed $42$.

The activity vocabulary covers $10$ appliances grouped into $4$ regimes
that drive distractor selection in needle-based tasks
(Table~\ref{tab:appendix_uk_dale_vocab}). Per-appliance availability is
uneven by construction: \emph{washing\_machine} and \emph{fridge} are
present in only one kept house each (h2), \emph{washer\_dryer} in two
(h1, h5), and \emph{oven} in one (h5).

\begin{table}[h]
\centering
\small
\setlength{\tabcolsep}{6pt}
\renewcommand{\arraystretch}{1.05}
\caption{UK-DALE-Haystack activity vocabulary. Per-appliance bout counts
are reported across all $287$ kept ISO-weeks, totalling $57{,}672$ bouts.}
\label{tab:appendix_uk_dale_vocab}
\begin{tabular}{@{}llrrr@{}}
\toprule
Regime & Appliance & \# bouts & Median dur.\ (s) & Mean peak (W) \\
\midrule
\multirow{4}{*}{impulse}
  & kettle           &     985 &  163 & 3128 \\
  & microwave        &   8{,}089 &   52 & 1646 \\
  & toaster          &      33 &   98 & 1117 \\
  & hair\_dryer      &   1{,}981 &   48 & 1294 \\
\midrule
\multirow{3}{*}{long\_cycle}
  & washing\_machine &      57 & 2414 & 2417 \\
  & dishwasher       &     779 & 5107 & 2655 \\
  & washer\_dryer    &   1{,}289 & 5397 & 2546 \\
\midrule
cooking
  & oven             &      58 &  950 & 2387 \\
\midrule
\multirow{2}{*}{refrig}
  & fridge           &   3{,}829 &  918 &  328 \\
  & fridge\_freezer  &  40{,}572 & 1396 &  280 \\
\bottomrule
\end{tabular}
\end{table}

\paragraph{UK-DALE-Haystack: benchmark statistics.}
\label{app:ukdale-haystack-stats}
UK-DALE-Haystack instantiates all ten TS-Haystack tasks at five context
lengths: $15$\,min, $1$\,h, $2$\,h, $9$\,h, and $24$\,h, with a
fixed $1{,}000\,/\,150\,/\,150$ train\,/\,validation\,/\,test sample
budget per (task, length) cell. The $24$\,h context is the longest
horizon in TS-Haystack, supported by UK-DALE's low $6$\,s sampling
period: a full day at $6$\,s is $14{,}400$ samples, well within the
context windows TSLMs can actually ingest, whereas at $100$\,Hz the
same duration would require $8.6$\,M samples.

\paragraph{UK-DALE-Haystack: construction pipeline.}
\label{app:ukdale-haystack-construction}
UK-DALE-Haystack is constructed by semi-synthetic insertion under the
\emph{additive} composition operator described in
Appendix~\ref{app:needle-insertion}. The operator is determined by the
sensor model: a house's whole-mains active-power demand is the sum of its appliance sub-meters, so given a target-OFF mains window
and a real ON-event sub-meter trace of the queried appliance, the
``what if this appliance were also running'' mains signal is exactly
their sum. We therefore do not apply the necessary statistical transformations used for Capture24 recordings: the sum is the physically correct mains that would have
been recorded if the appliance had been running.

We construct each sample as a deterministic function of the (task,
context length, sample index, split) tuple via a seeded random-number
generator, ensuring bit-identical reproducibility across re-runs. The
shared three-step procedure is:

\begin{enumerate}
\setlength{\itemsep}{2pt}
\item \textbf{Background selection.} A target context length $L$
defines the haystack window. We sample a house from those in which the
queried target appliance $c$ exists, take that house's recording within
the relevant split, and compute the OFF-intervals as the complement of
the precomputed bout list for $c$ on that house, and uniformly sample a window start inside one of
the surviving intervals.
\item \textbf{Needle sampling.} For each needle slot required by the
task, we sample an extracted bout of class $c$ from the bout index,
restricted to the same split as the background and (by default) to the
same house as the background. \textbf{Bouts are kept at their native
duration and unscaled}: the needle is the real sub-meter trace of one
ON-event, so the duration encodes appliance identity (a kettle bout is
a $\sim$$2$\,min impulse, a wash cycle is a $\sim$$90$\,min envelope).
Same-house pairing is the default as we assume each physical appliance has
its own electrical signature.
\item \textbf{Additive insertion.} Each needle is placed at a
controlled non-overlapping position in $[m, L - m]$ that determines the
ground-truth answer, and inserted by addition rather than replacement:
\begin{equation}
\mathbf{x}'_{[p:p+w]}
\;=\; \mathbf{x}_{\text{bg},[p:p+w]} \;+\; \mathbf{x}_{\text{needle}},
\end{equation}
where $\mathbf{x}_{\text{bg}}$ is the target-OFF mains window, $w$ is
the needle's native length, and $p$ is the insertion position. Outside
$[p, p+w]$ the synthetic mains is exactly the target-OFF background.
\end{enumerate}

\paragraph{Native needle duration.}
A design choice worth flagging: the needle width $w$ is the native
duration of the sampled bout rather than a fixed per-(task, length)
constant. This contrasts with Capture24-Haystack, where bouts are
trimmed to a shared per-cell width. The motivation is that appliance
identity in mains-only NILM is partly carried by duration, so
trimming would erase part of the ground-truth signal.

\paragraph{Anomaly synthesis.}
The two anomaly tasks (\texttt{anomaly\_detection},
\texttt{anomaly\_localization}) draw their positive needles from
synthetic modifications of real bouts rather than statistical-tail
outliers. Two classes are implemented:
(i) \textbf{truncated\_cycle} (long\_cycle and cooking regimes only):
trim a real bout from its end to a fraction $t \in [0.20, 0.50]$ of its
duration, modelling a wash cycle interrupted mid-run or an oven turned
off mid-cook;
(ii) \textbf{abnormal\_peak} (impulse and cooking regimes only): scale
the bout's sub-meter trace by $s \in [1.5, 2.5]$ capped at $3.5$\,kW. Negative
\texttt{anomaly\_detection} samples insert a randomly drawn nominal
bout so the model cannot shortcut on ``needle present $\Rightarrow$ anomaly''.

\paragraph{Insertion validation.}
\label{app:ukdale-insertion-validation}
The Capture24-style discrimination test of
Appendix~\ref{app:insertion_validation} is not the right validation
criterion for UK-DALE: the mains is an aggregate of the household's
loads, and a natural ``appliance-on'' window is by definition drawn
from moments when the household was running that appliance.

\subsection{ARTS classifier construction (UK-DALE)}
\label{app:arts-classifier-ukdale}

The ARTS classifier tool $g_\phi$ for UK-DALE is a 1-D dilated
temporal convolutional network (TCN) trained end-to-end on the
single-channel $6$\,s active-power mains, producing per-sample
multi-label predictions over the $10$-appliance vocabulary
(Table~\ref{tab:appendix_uk_dale_vocab}). The input is featurised as
two channels: $\log(1 + \text{power}_W)$ for the absolute level, and
$\text{sign}(\Delta) \cdot \log(1 + |\Delta|)$ on the first-difference
for edge sign and magnitude. The latter is what distinguishes
appliances whose level distributions overlap (\emph{kettle},
\emph{microwave}, \emph{hair\_dryer} all peak around $1$--$3$\,kW) but
whose turn-on and turn-off transients differ markedly. The architecture
is a $7$-stage stack of residual dilated blocks, each consisting of two
padded $3{\times}3$ Conv1d layers, BatchNorm, ReLU, $0.1$ dropout, and
a residual connection, with dilations $(1, 2, 4, 8, 16, 32, 64)$ and
channel widths $(96, 96, 128, 128, 192, 192, 192)$. A $7$-tap stem
fronts the stack and a $1\!\times\!1$ classification head maps to $10$
logits per input sample. The receptive field is $\sim$$133$ samples
($\sim$$13$\,min at $6$\,s), wide enough to span impulse-class bouts
and the boundaries of long-cycle bouts.

The training curriculum for ARTS' UK-Dale classifier happens in two steps. \textbf{Stage 1: pretraining on raw mains.} The TCN is trained from
scratch on raw mains windows sampled from the $232$ training ISO-weeks,
with rare-class oversampling (anchored crops biased toward
\texttt{dishwasher}, \texttt{kettle}, \texttt{toaster}, \texttt{oven},
etc.), per-class \texttt{pos\_weight} capped at $50$, AdamW
($\text{lr} = 10^{-3}$, weight decay $10^{-4}$), batch size $32$,
$768$-sample windows, $15$ epochs of $1{,}500$ steps. The Stage-1
checkpoint attains best-validation macro-F$1$ $= 0.868$ and held-out
test macro-F$1$ $= 0.796$ on real-mains test weeks. \textbf{Stage 2: fine-tuning on the QA distribution.} The Stage-1 checkpoint is fine-tuned on the UK-DALE-Haystack training parquets
directly, using the additively reconstructed mains as input and the
union of inserted-needle and natural-bout intervals as ground truth. Fine-tuning uses $\text{lr} = 5 \times 10^{-5}$, $12$ epochs of $1{,}500$ steps, anchored crops with probability $0.7$, and best-validation checkpoint selection. The Stage-2 checkpoint is the deployed $g_\phi$ used by all UK-DALE-Haystack ARTS evaluations reported in the main paper.

%%%%%%%%%%%%%%%%%%%%%% TASKS CONSTRUCTION DETAILS %%%%%%%%%%%%%%%%%%%%%%%
\subsection{Task construction details}
\label{app:task-construction-details}

In the following, we provide details of each of the 10 proposed tasks for TS-Haystack, exemplified with a random sample visualization from the Capture24 Haystack generation pipeline. 
Task variations such as \emph{non} Existence, \emph{non} Anomaly Detection or all 8 Comparison variants are excluded for brevity.

%% ============================================================================
%% TASK 1: Existence
%% ============================================================================
\paragraph{Task 1: Existence.} \textit{``Is there \{activity\} in this recording?''}

This task tests whether the model can detect the presence of a specific activity pattern within a longer recording. To prevent variance-based detection shortcuts (where models detect \textit{any} insertion rather than the specific activity), we insert multiple same-regime distractors. The model must learn activity-specific patterns rather than relying on statistical anomalies.

\textbf{Construction:} (1) Sample background window; (2) Select activity regime (sedentary/active); (3) Insert $N$ needles from regime activities not in background; (4) For positive samples, ask about an inserted activity; for negative, ask about a non-inserted activity from the same regime.

\textbf{Example:} \textit{Q: ``Is there manual-work in this recording?''} \textit{A: ``No.''}

\begin{figure}[h]
\centering
\includegraphics[width=0.8\textwidth]{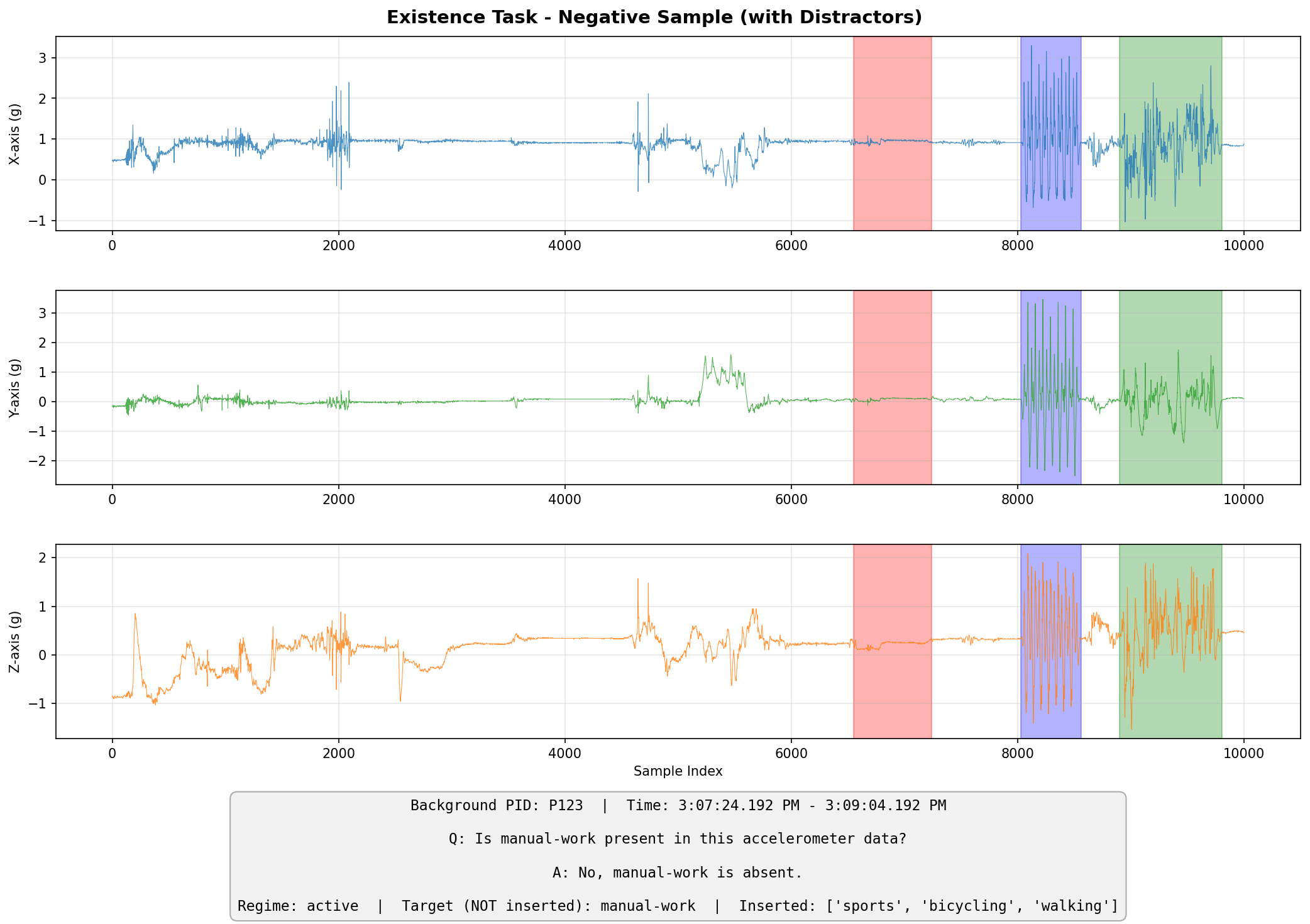}
\caption{Task 1: Existence. The model must detect whether a specific activity is present among same-regime distractors.}
\label{fig:task_existence}
\end{figure}

%% ============================================================================
%% TASK 2: Localization
%% ============================================================================
\paragraph{Task 2: Localization.} \textit{``When did the \{activity\} bout occur?''}

Extends existence by requiring temporal grounding. The model must not only detect the activity but report its precise time range. Distractors with similar statistical properties force the model to identify activity-specific temporal signatures.

\textbf{Construction:} Same as Existence, but answer requires the timestamp range of the target needle.

\textbf{Example:} \textit{Q: ``When did the sports bout occur?''} \textit{A: ``From 02:09:52:635 to 02:10:08:967.''}

\begin{figure}[h]
\centering
\includegraphics[width=0.8\textwidth]{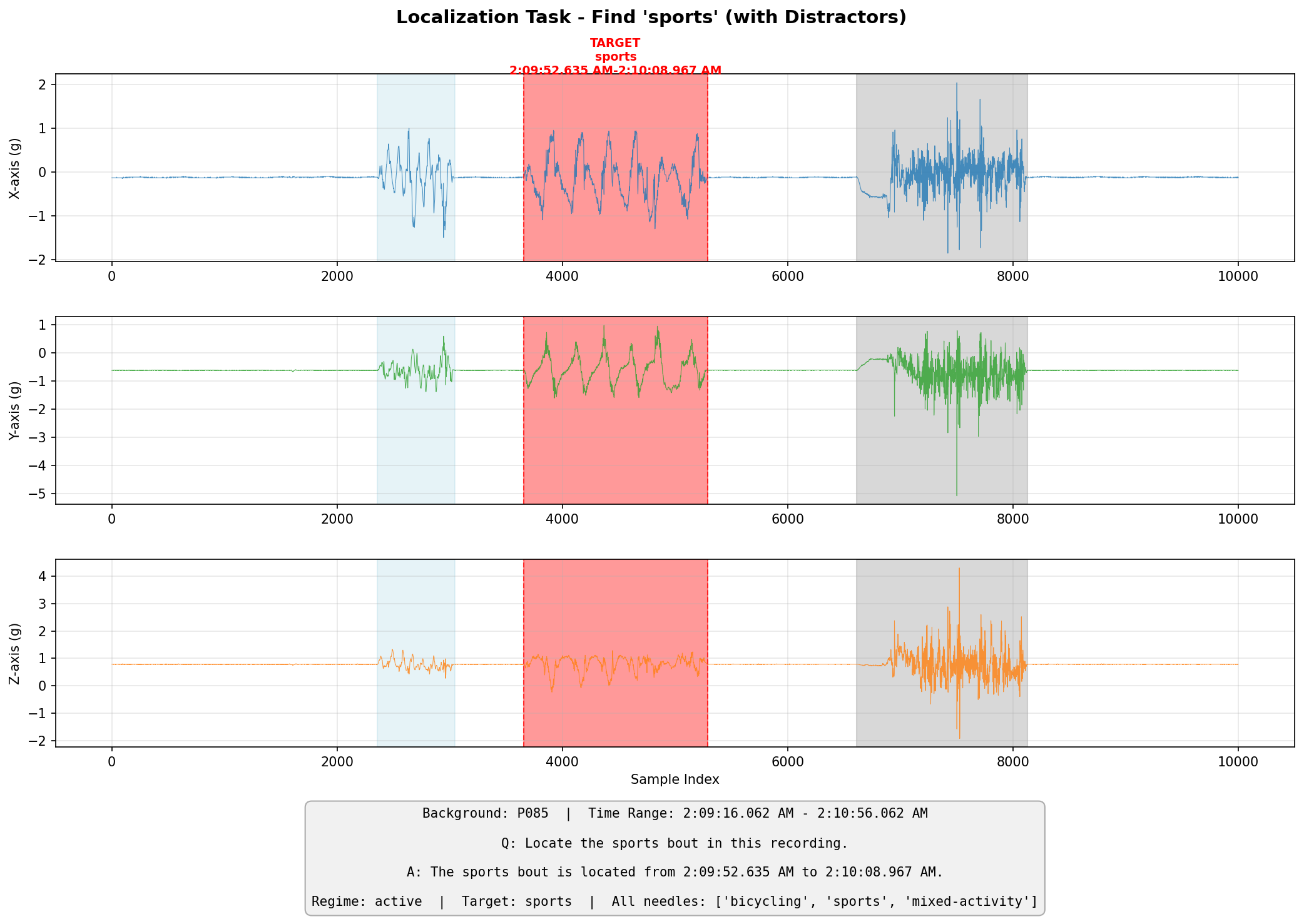}
\caption{Task 2 (Localization). The model must identify and temporally ground a specific activity}
\label{fig:task_localization}
\end{figure}

%% ============================================================================
%% TASK 3: Counting
%% ============================================================================
\paragraph{Task 3: Counting.} \textit{``How many \{activity\} bouts occurred?''}

Tests enumeration capabilities: the model must detect all instances of an activity and count them correctly. Minimum gaps between bouts ensure distinguishability while making the task non-trivial.

\textbf{Construction:} (1) Sample background; (2) Sample $N \in [1, 5]$ bouts of target activity; (3) Insert at non-overlapping positions with minimum gap constraint.

\textbf{Example:} \textit{Q: ``How many standing bouts occurred?''} \textit{A: ``3.''}

\begin{figure}[h]
\centering
\includegraphics[width=0.8\textwidth]{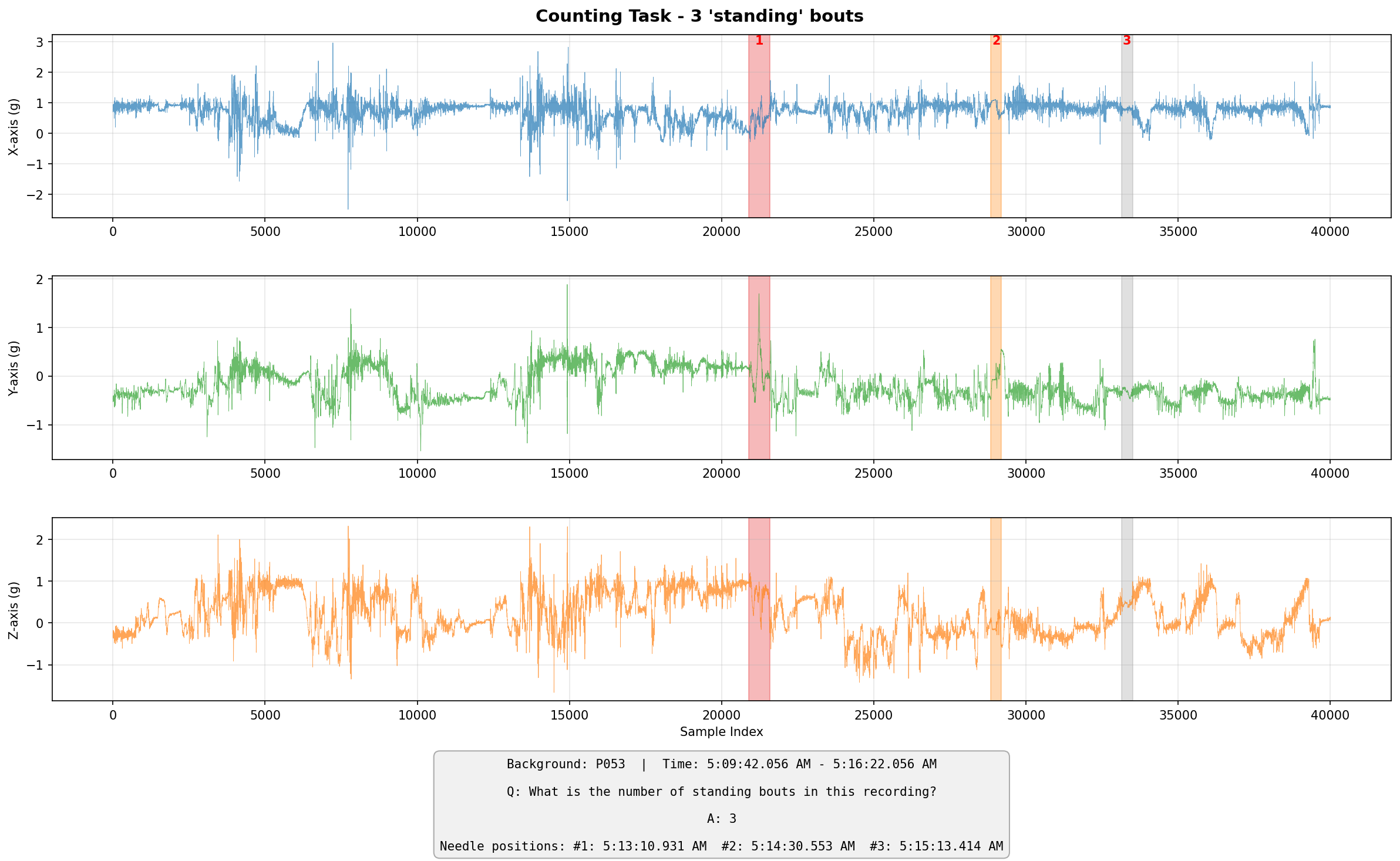}
\caption{Task 3 (Counting). The model must enumerate all occurrences of a specific activity}
\label{fig:task_counting}
\end{figure}

%% ============================================================================
%% TASK 4: Ordering
%% ============================================================================
\paragraph{Task 4: Ordering.} \textit{``Did \{activity\_a\} occur before \{activity\_b\}?''}

Tests temporal comparison between two distinct activities. We use needle insertion rather than natural bout selection to prevent models from exploiting learned activity transition statistics. Using random needle sampling pairings instead of existing activity transitions in the data ensure models cannot game the benchmark through learned Markovian priors, requiring genuine temporal retrieval for correct answers \citep{chan2024capture24}.

\textbf{Construction:} (1) Sample two distinct activities $A$ and $B$; (2) Sample background excluding both; (3) Randomly assign temporal order (50/50 balance); (4) Insert both needles sequentially.

\textbf{Example:} \textit{Q: ``Did sports occur before sleep?''} \textit{A: ``Yes.''}

\begin{figure}[h]
\centering
\includegraphics[width=0.8\textwidth]{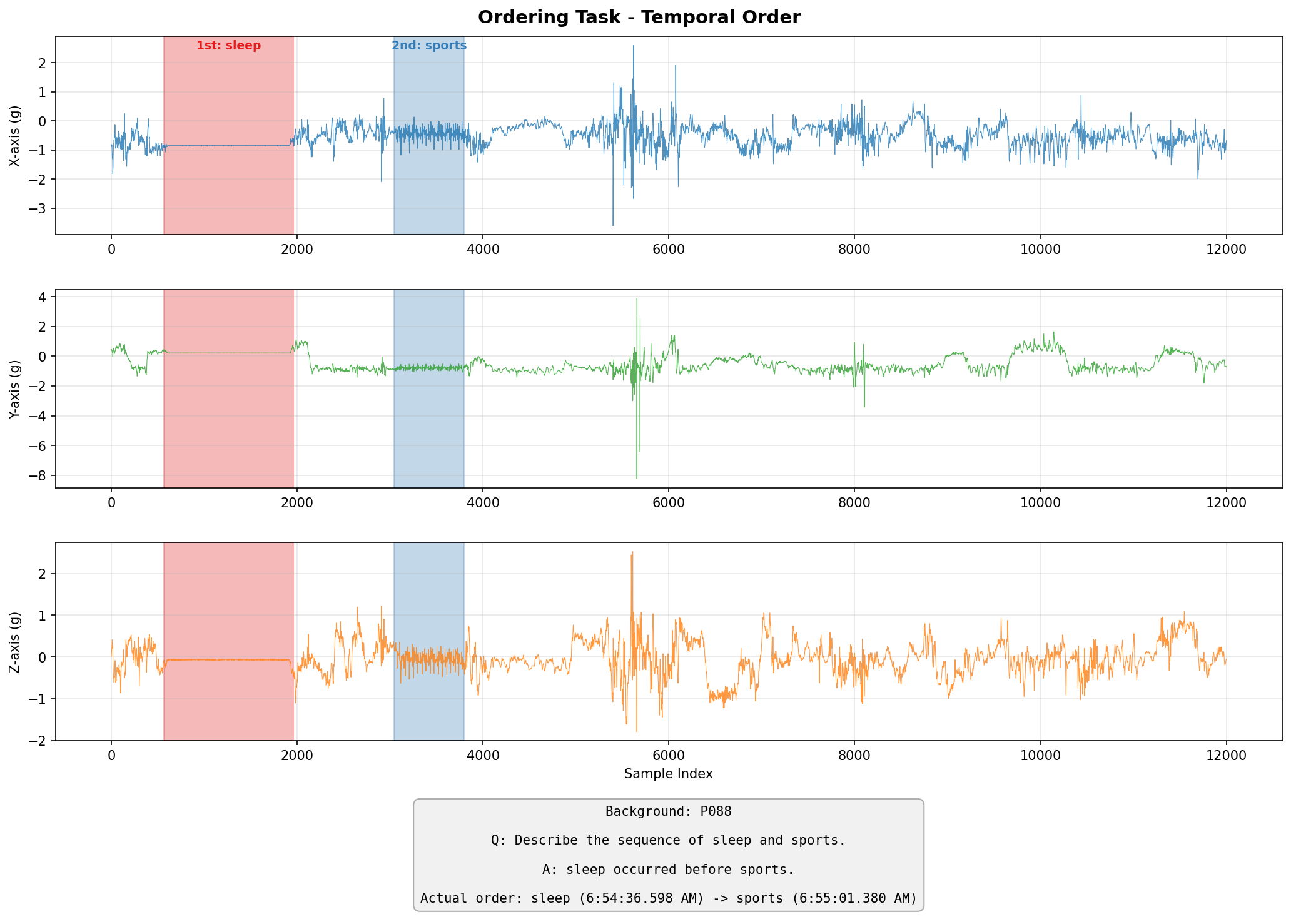}
\caption{Task 4 (Ordering). The model must determine the temporal relationship between two activities}
\label{fig:task_ordering}
\end{figure}

%% ============================================================================
%% TASK 5: State Query
%% ============================================================================
\paragraph{Task 5: State Query.} \textit{``What was the activity level when \{event\} occurred?''}

Tests cross-scale integration: the model must detect a local event (needle, seconds-scale) and identify the surrounding global activity regime (minutes-scale). This requires understanding hierarchical temporal context.

\textbf{Construction:} (1) Sample mixed background with 2--5 activity states; (2) Select a global state region; (3) Insert needle within that region; (4) Ask about the global state, not the needle activity.

\textbf{Example:} \textit{Q: ``What was the overall activity when the vehicle activity occurred?''} \textit{A: ``Standing.''}

\begin{figure}[h]
\centering
\includegraphics[width=0.8\textwidth]{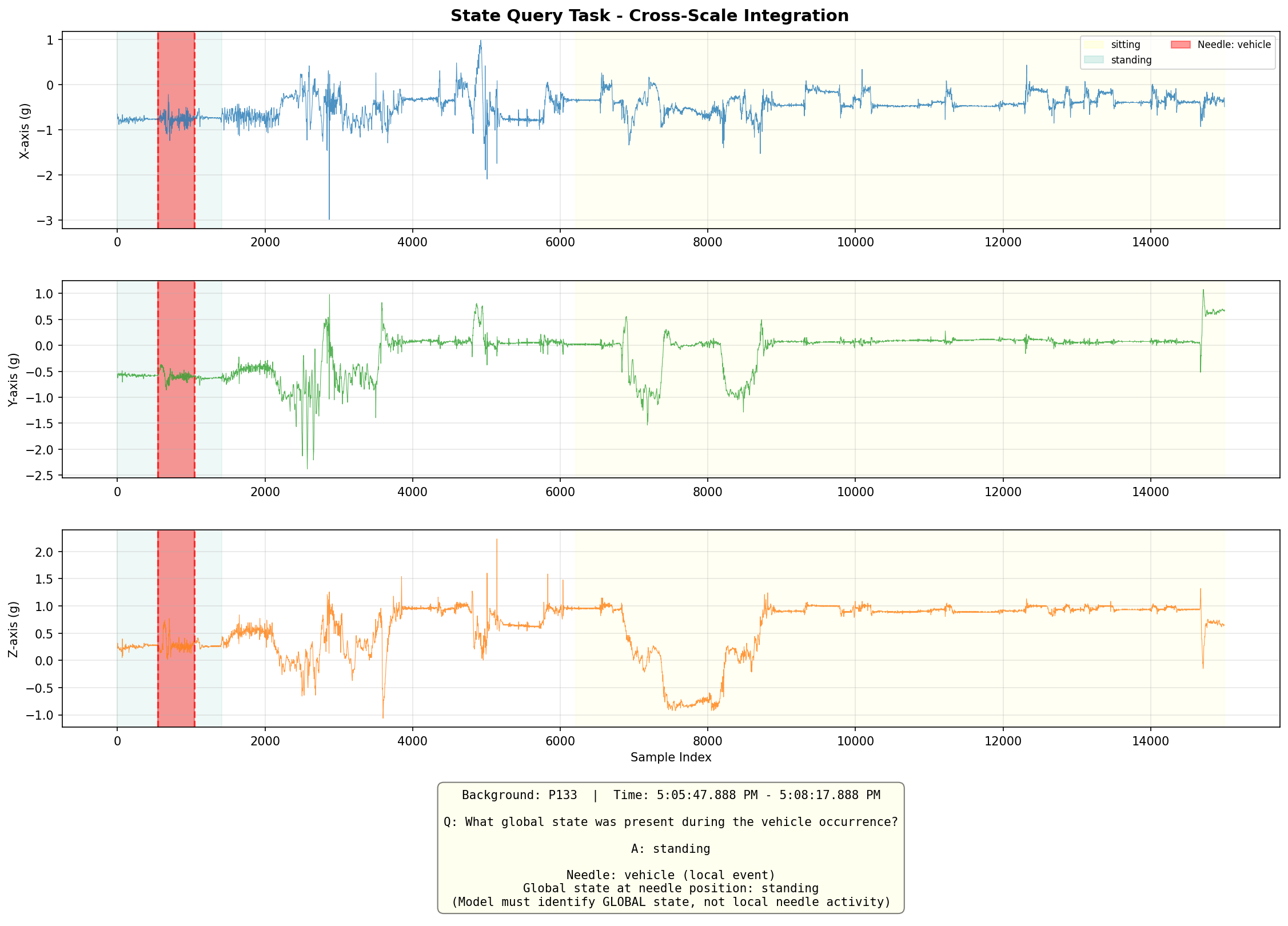}
\caption{Task 5 (State Query). The model must identify the global activity regime surrounding a local event}
\label{fig:task_state_query}
\end{figure}

%% ============================================================================
%% TASK 6: Antecedent
%% ============================================================================
\paragraph{Task 6: Antecedent.} \textit{``What activity occurred immediately before \{target\}?''}

Tests sequential reasoning: the model must locate the target activity and identify what directly preceded it. Two-needle insertion gives control over the antecedent-target pairing, preventing reliance on natural transition statistics.

\textbf{Construction:} (1) Sample background (prefer low-activity); (2) Sample antecedent $A$ and target $T$; (3) Insert $A$ followed immediately by $T$ with small adjacency gap.

\textbf{Example:} \textit{Q: ``Looking at the sequence, what came before bicycling?''} \textit{A: ``Sports.''}

\begin{figure}[h]
\centering
\includegraphics[width=0.8\textwidth]{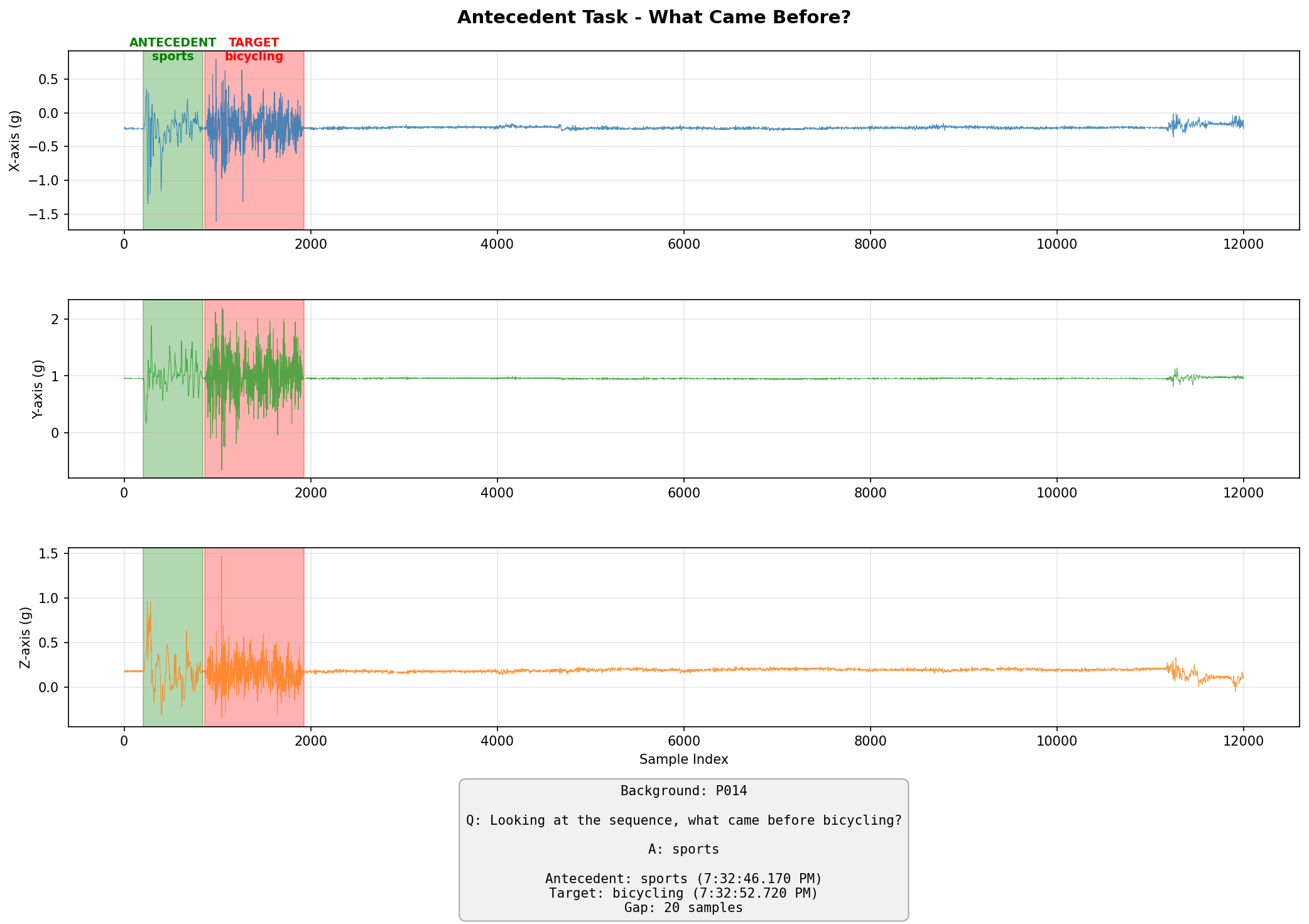}
\caption{Task 6 (Antecedent). The model must identify the activity immediately preceding a target}
\label{fig:task_antecedent}
\end{figure}

%% ============================================================================
%% TASK 7: Comparison
%% ============================================================================
\paragraph{Task 7: Comparison.} \textit{``What was the \{longest/shortest\} period \{with/without\} \{activity\}?''}

Tests comparison reasoning with four question variants: longest/shortest $\times$ with/without. The ``without'' polarity requires understanding negation and computing gaps between activity bouts. All inserted bouts have distinct durations to ensure unambiguous answers.

\textbf{Construction:} (1) Sample question type (extremum $\times$ polarity); (2) Insert $N \geq 2$ bouts with distinct durations; (3) For ``with'': answer is extremum bout; for ``without'': answer is extremum gap.

\textbf{Example:} \textit{Q: ``What was the longest period without household-chores?''} \textit{A: ``From 8:32:52:619 to 08:35:30:682.''}

\begin{figure}[h]
\centering
\includegraphics[width=0.8\textwidth]{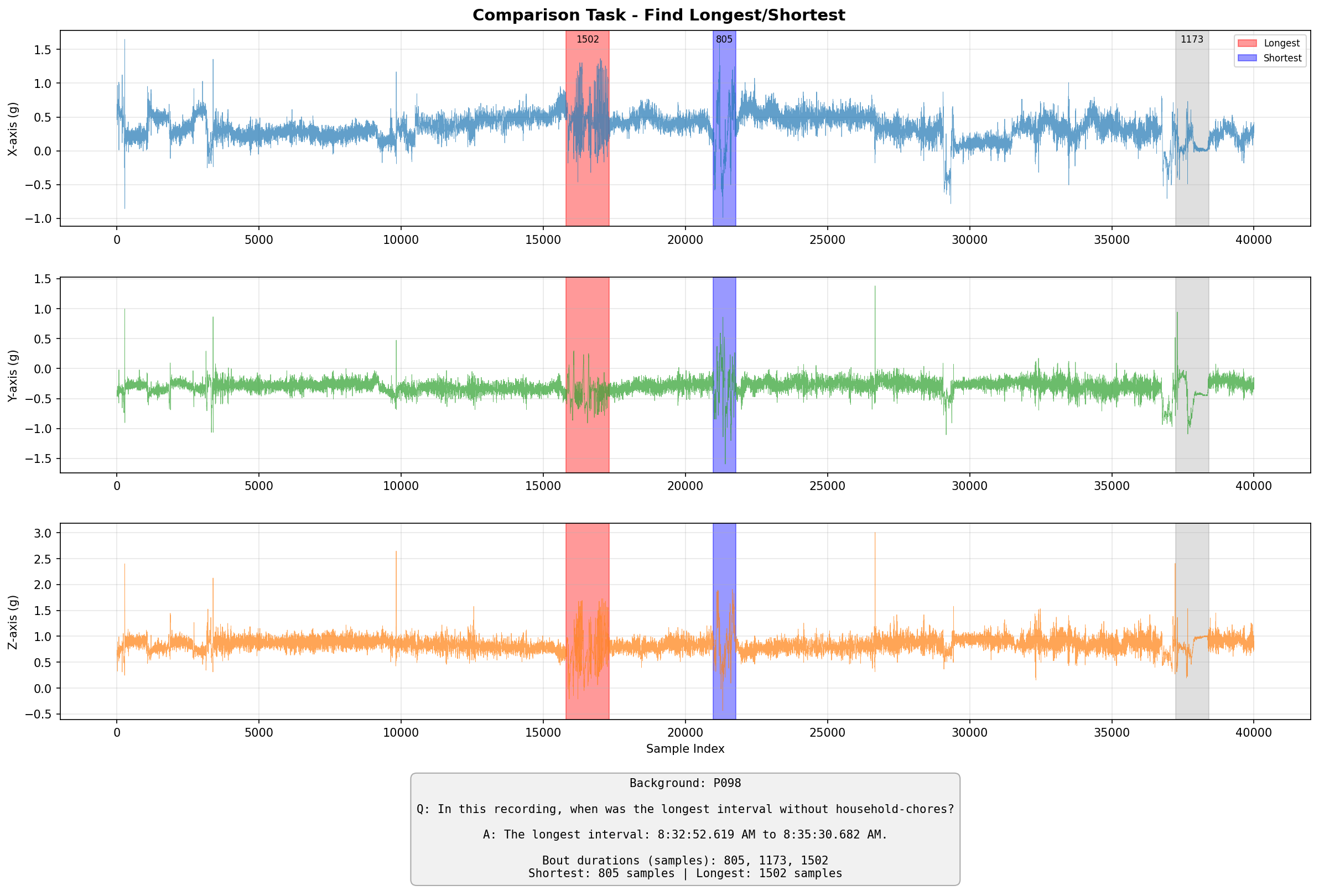}
\caption{Task 7 (Comparison). The model must compare durations across multiple activity bouts or gaps}
\label{fig:task_comparison}
\end{figure}

%% ============================================================================
%% TASK 8: Multi-Hop
%% ============================================================================
\paragraph{Task 8: Multi-Hop.} \textit{``When did the K-th \{target\} occur \{before/after\} \{anchor\}?''}

Tests multi-step reasoning: (1) locate the anchor activity, (2) identify target bouts relative to anchor, (3) count to the $K$-th occurrence in the specified direction. Optional distractors on the opposite side test directional understanding.

\textbf{Construction:} (1) Sample anchor $A$ and target $T$; (2) Sample $K \in \{1, 2, 3\}$ and direction; (3) Position anchor with room for $K$ targets; (4) Insert $K$ target needles in order.

\textbf{Example:} \textit{Q: ``When did the third sports bout occur after mixed-activity?''} \textit{A: ``From 09:38:32:874 to 09:38:40:924.''}

\begin{figure}[h]
\centering
\includegraphics[width=0.8\textwidth]{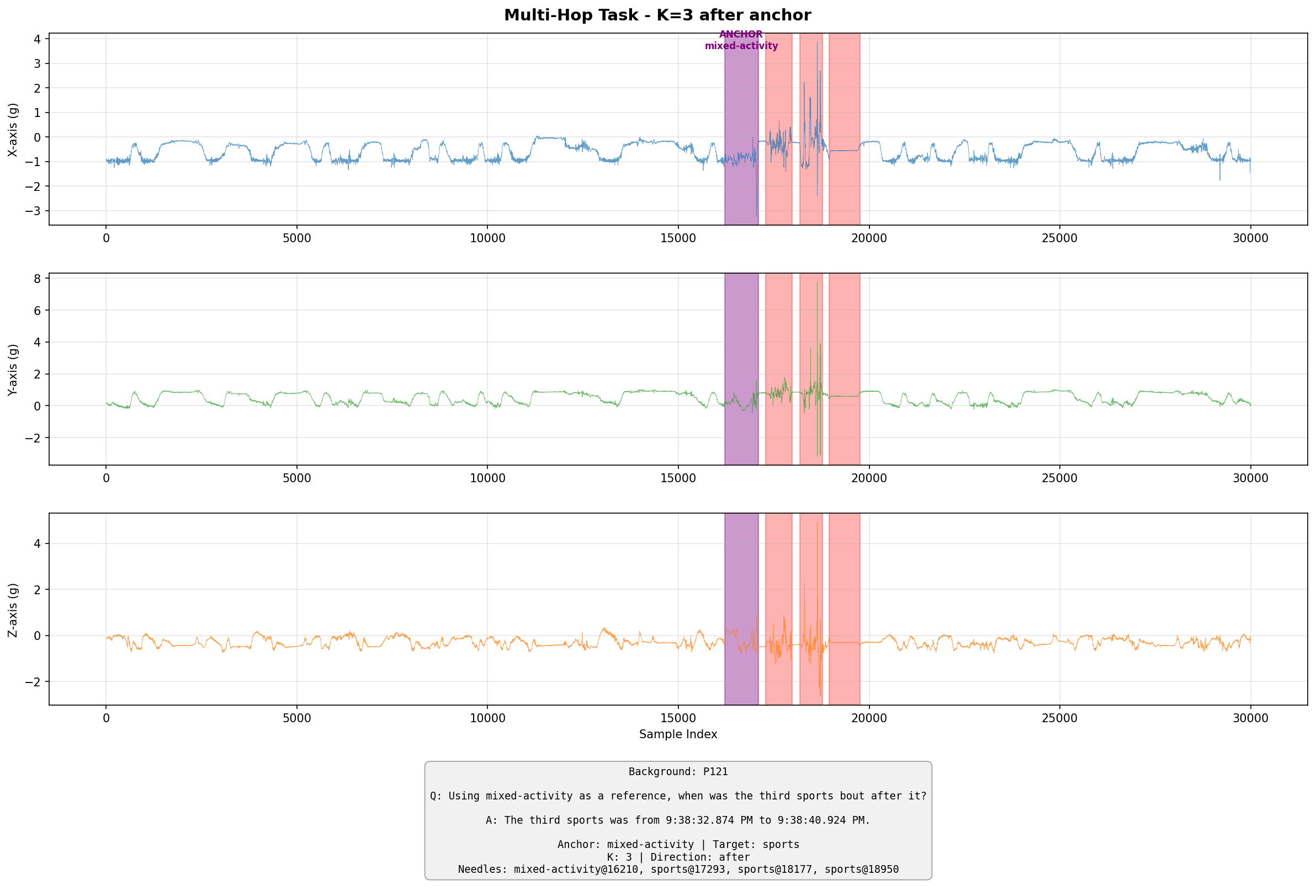}
\caption{Task 8 (Multi-Hop). The model must locate an anchor, then count to the $K$-th target in a specified direction}
\label{fig:task_multi_hop}
\end{figure}

%% ============================================================================
%% TASK 9: Anomaly Detection
%% ============================================================================
\paragraph{Task 9: Anomaly Detection.} \textit{``Is there an anomaly in this recording?''}

Tests contextual reasoning: the model must detect activity that is anomalous \textit{relative to the background regime}. Unlike Existence (where the target is given), the model must identify what constitutes an anomaly. We define anomalies as cross-regime insertions (e.g., vigorous activity in a sedentary background). We partition activities into two regimes: \textit{sedentary} (sleep, sitting, standing, vehicle) and \textit{active} (walking, mixed-activity, bicycling, manual-work, sports, household-chores). Mandatory same-regime distractors prevent ``insertion detected'' shortcuts: both positive and negative samples contain insertions, so the model must identify regime mismatch.

\textbf{Construction:} (1) Sample pure background from one regime; (2) Positive: insert 1 cross-regime needle (anomaly) + $N$ same-regime distractors; Negative: insert only same-regime distractors.

\textbf{Example:} \textit{Q: ``Is there an anomaly in this recording?''} \textit{A: ``Yes, there is anomalous sitting activity in the active background.''}

\begin{figure}[h]
\centering
\includegraphics[width=0.8\textwidth]{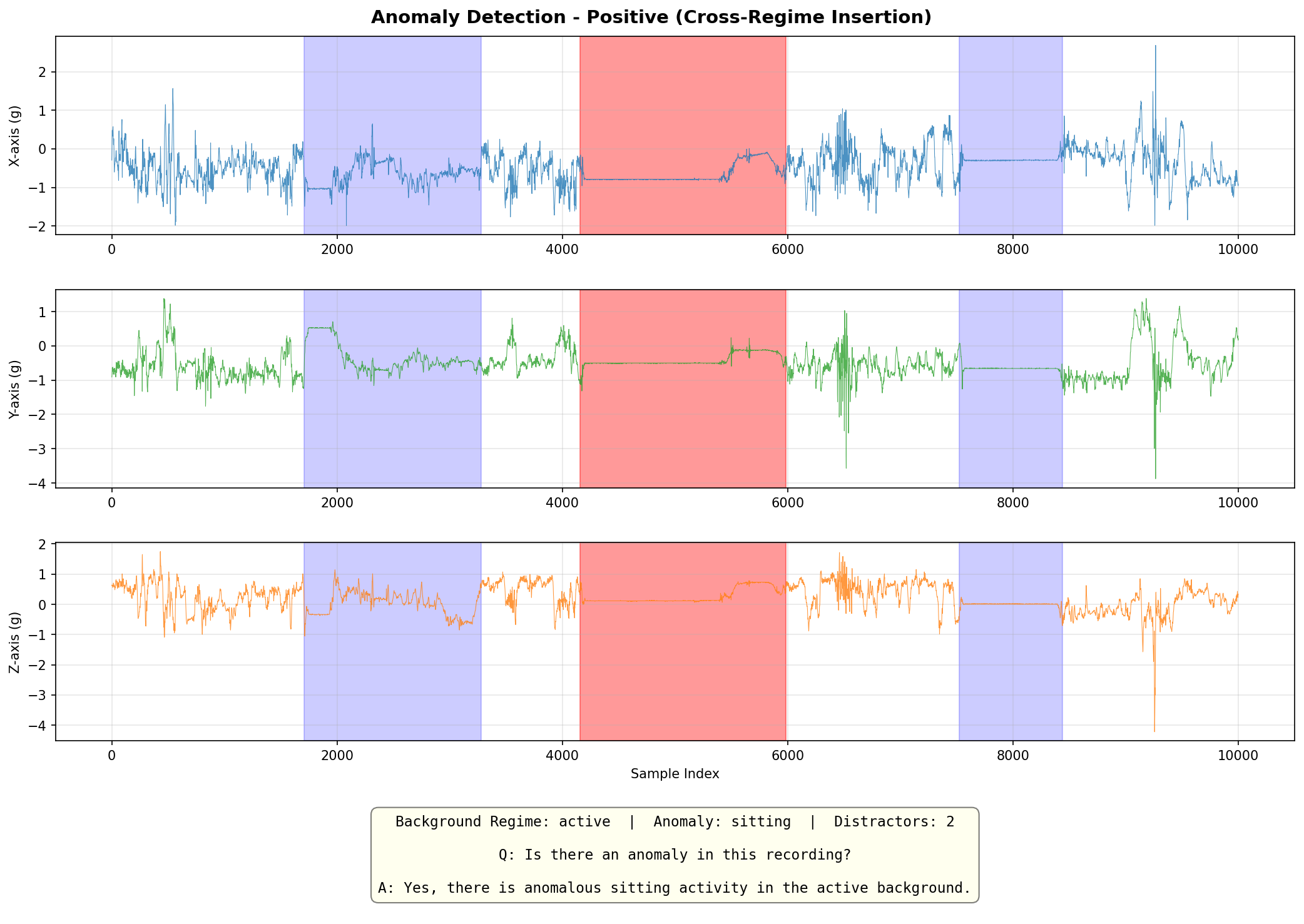}
\caption{Task 9 (Anomaly Detection). The model must identify cross-regime activity without being told what to look for}
\label{fig:task_anomaly_detection}
\end{figure}

%% ============================================================================
%% TASK 10: Anomaly Localization
%% ============================================================================
\paragraph{Task 10: Anomaly Localization.} \textit{``Is there an anomaly, and if so, when did it occur?''}

Extends anomaly detection by requiring temporal localization. The model must detect the cross-regime violation and specify its time range. This combines contextual reasoning with precise temporal grounding.

\textbf{Construction:} Same as Anomaly Detection, but positive answers must include the time range of the anomalous activity.

\textbf{Example:} \textit{Q: ``Is there an anomaly, and if so, when?''} \textit{A: ``Yes, there is anomalous sitting activity from 04:21:20:792 to 04:21:34:333.''}

\begin{figure}[h]
\centering
\includegraphics[width=0.8\textwidth]{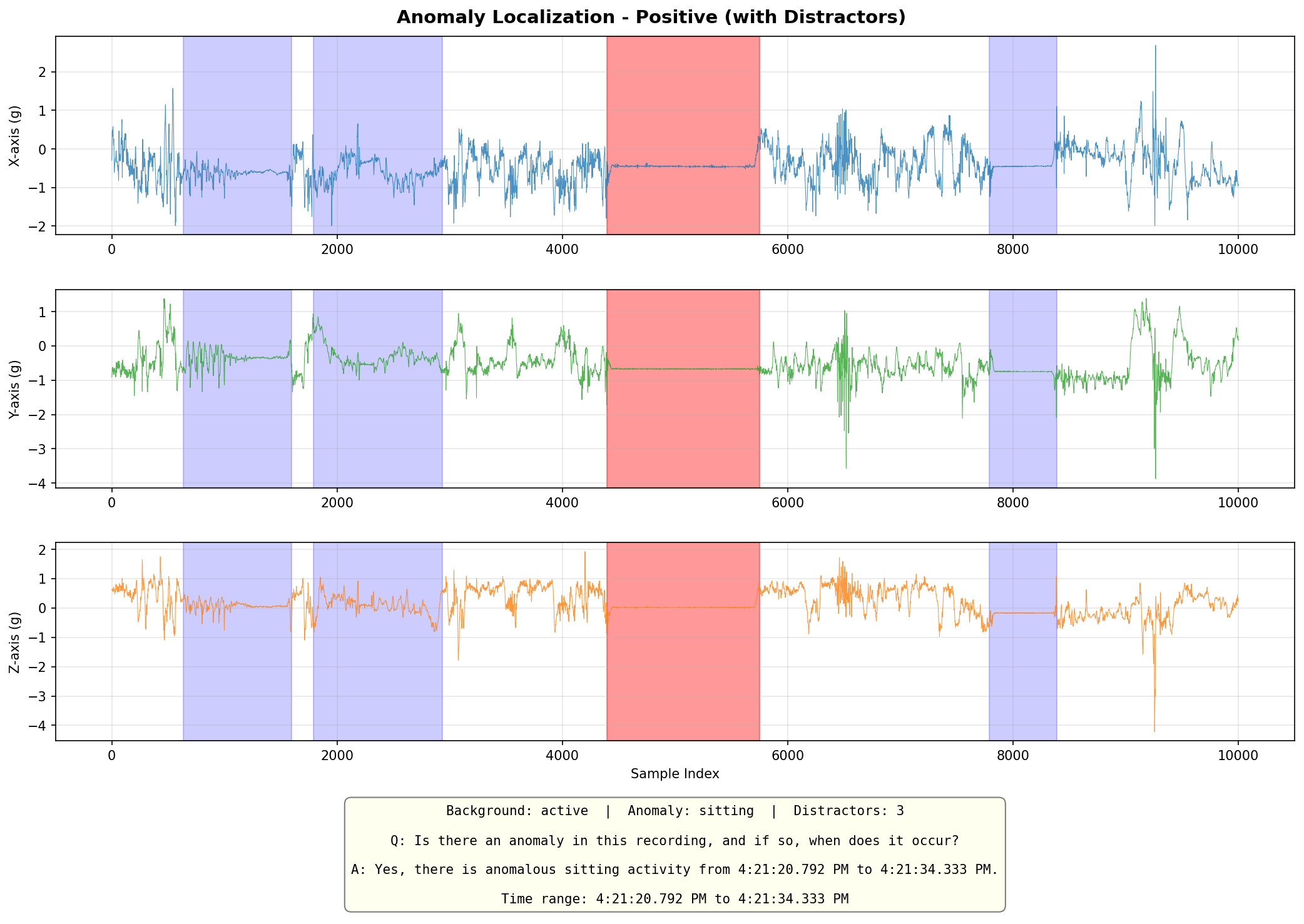}
\caption{Task 10 (Anomaly Localization). The model must detect and temporally ground cross-regime anomalies}
\label{fig:task_anomaly_localization}
\end{figure}

\end{document}